%% file: main.tex
\documentclass[runningheads]{llncs}
\usepackage{graphicx}

\usepackage{tikz}
\usepackage{comment}
\usepackage{amsmath,amssymb} 
\usepackage{color}

\usepackage{graphicx}
\usepackage{amsmath}
\usepackage{amssymb}
\usepackage{booktabs}

\usepackage[utf8]{inputenc} 
\usepackage[T1]{fontenc}    
\usepackage{url}            
\usepackage{booktabs}       
\usepackage{amsfonts}       
\usepackage{nicefrac}       
\usepackage{microtype}      

\usepackage{multirow}
\usepackage{xspace}
\usepackage{mathtools}
\usepackage{wasysym}
\usepackage{marvosym}
\usepackage{xcolor}
\usepackage{color}
\usepackage{tabularx}
\usepackage{float}
\usepackage{diagbox,tabu,stackengine}
\usepackage{bbm}
\usepackage{bm}
\usepackage{mwe}
\usepackage{graphbox}
\usepackage{sidecap}

\usepackage{algorithm}
\usepackage[noend]{algpseudocode}
\usepackage{comment}
\usepackage{cancel}
\usepackage{ulem}
\usepackage{array}
\newcommand{\PreserveBackslash}[1]{\let\temp=\\#1\let\\=\temp}
\newcolumntype{C}[1]{>{\PreserveBackslash\centering}p{#1}}
\newcolumntype{R}[1]{>{\PreserveBackslash\raggedleft}p{#1}}
\newcolumntype{L}[1]{>{\PreserveBackslash\raggedright}p{#1}}

\usepackage{xspace}
\makeatletter
\DeclareRobustCommand\onedot{\futurelet\@let@token\@onedot}
\def\@onedot{\ifx\@let@token.\else.\null\fi\xspace}
\def\eg{\textit{e.g}\onedot} 
\def\ie{\textit{i.e}\onedot}

\def\etal{\textit{et al}\onedot}
\makeatother

\newcommand{\tabincell}[2]{\begin{tabular}{@{}#1@{}}#2\end{tabular}}

\newcommand{\mysubsubsection}[1]{\noindent\textbf{#1 }}
\newcommand{\myparagraph}[1]{\noindent\emph{#1 }}

\usepackage[pagebackref,breaklinks,colorlinks]{hyperref}

\usepackage[capitalize]{cleveref}
\crefname{section}{Sec.}{Secs.}
\Crefname{section}{Section}{Sections}
\Crefname{table}{Table}{Tables}
\crefname{table}{Tab.}{Tabs.}

\begin{document}
\pagestyle{headings}
\mainmatter
\def\ECCVSubNumber{6516}  

\title{Neural Light Field Estimation for Street Scenes with Differentiable Virtual Object Insertion} 

\titlerunning{Neural Light Field Estimation for Street Scenes}
%
\author{Zian Wang\inst{1,2,3} \and
Wenzheng Chen\inst{1,2,3} \and
David Acuna\inst{1,2,3} \and \\
Jan Kautz\inst{1} \and
Sanja Fidler\inst{1,2,3}}
\authorrunning{Z. Wang et al.}
%
\institute{
$^1$NVIDIA \quad $^2$University of Toronto \quad $^3$Vector Institute \\
\email{\{zianw,wenzchen,dacunamarrer,jkautz,sfidler\}@nvidia.com}}
\maketitle

\begin{abstract}
We consider the challenging problem of outdoor lighting estimation for the goal of photorealistic virtual object insertion into photographs. Existing works on outdoor lighting estimation typically simplify the scene lighting into an environment map which cannot capture the spatially-varying lighting effects in outdoor scenes. In this work, we propose a neural approach that estimates the 5D HDR light field from a single image, and a differentiable object insertion formulation that enables end-to-end training with image-based losses that encourage realism. Specifically, we design a hybrid lighting representation tailored to outdoor scenes, which contains an HDR sky dome that handles the extreme intensity of the sun, and a volumetric lighting representation that models the spatially-varying appearance of the surrounding scene. With the estimated lighting, our shadow-aware object insertion is fully differentiable, which enables adversarial training over the composited image to provide additional supervisory signal to the lighting prediction. We experimentally demonstrate that our hybrid lighting representation is more performant than existing outdoor lighting estimation methods. We further show the benefits of our AR object insertion in an autonomous driving application, where we obtain performance gains for a 3D object detector when trained on our augmented data. 
\end{abstract}

\input{intro}

\input{related}

\input{method}

\input{results}

\input{conc}

%
%
\bibliographystyle{splncs04}
\bibliography{egbib}
\end{document}


\pagestyle{headings}
\mainmatter
\def\ECCVSubNumber{6516}  

\title{Supplementary Material:\\Neural Light Field Estimation for Street Scenes with Differentiable Virtual Object Insertion} 

\titlerunning{Supplementary Material: Neural Light Field Estimation for Street Scenes}
%
\author{Zian Wang\inst{1,2,3} \and
Wenzheng Chen\inst{1,2,3} \and
David Acuna\inst{1,2,3} \and \\
Jan Kautz\inst{1} \and
Sanja Fidler\inst{1,2,3}}
\authorrunning{Z. Wang et al.}
%
\institute{
$^1$NVIDIA \quad $^2$University of Toronto \quad $^3$Vector Institute \\
\email{\{zianw,wenzchen,dacunamarrer,jkautz,sfidler\}@nvidia.com}}
\maketitle

In the supplementary material, we include additional details and results of our approach. 
We provide technical details in Section~\ref{sec:details}. 
We show additional auxiliary results which further ablate and explain our method in Section~\ref{sec:results}. 
We discuss limitations and broader impact in Section~\ref{sec:discussion}.

\section{Implementation Details} 
\label{sec:details} 

\subsection{Sky Modeling Architecture} 
\begin{figure*}[t!]
\centering
\begingroup
\includegraphics[width=0.99\linewidth]{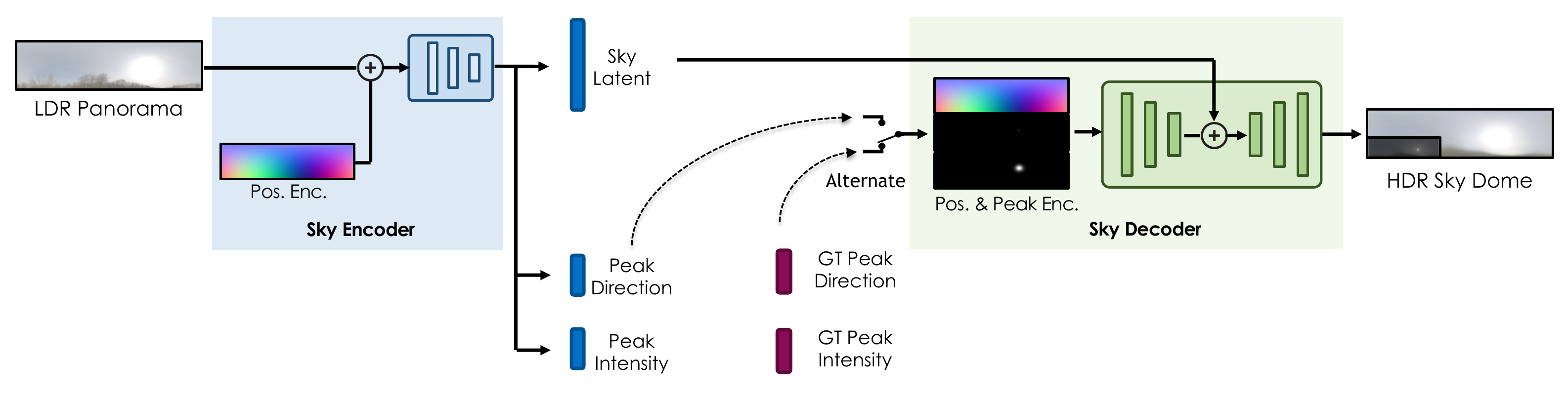} 
\endgroup
\caption{\textbf{Architecture of our sky modeling network.} The encoder takes as input an additional positional encoding, concatenated with the input LDR panorama. We compute a panorama image encoding peak and positional information, and feed it into a 2D UNet~\cite{unet} decoder. 
The sky latent code is fused in the latent space of the decoder. 
During training, we alternate between end-to-end training and teacher forcing. 
}
\label{fig:skymodel_arch} 
\end{figure*}

The primary goal of the sky modeling neural network is to learn a low-dimensional feature space for the sky dome. The resulting network can also predict HDR information given input LDR panorama. The model architecture is illustrated in Figure~\ref{fig:skymodel_arch}. 

A 2D CNN encoder takes as input an LDR panorama with a positional encoding, and predicts the sky vector $\hat{\mathbf{f}}$ in the feature space. 
As each pixel in the panorama corresponds to a direction through equi-rectangular projection, the positional encoding $(\mathbb{R}^{3 \times H \times W})$ encodes the directional information, where each pixel contains a unit vector indicating the direction of that pixel location. 
The decoder is a 2D UNet~\cite{unet} decoding the sky vector into an HDR sky panorama. We carefully design its architecture to facilitate the reconstruction of the HDR sun peak. 
The input to the 2D UNet is a 7-channel panorama, including a 4-channel peak encoding and a 3-channel positional encoding, as shown in Figure~\ref{fig:skymodel_arch}. 
Specifically, we embed the peak direction $\hat{\mathbf{f}}_\text{dir}$ information into a 1-channel peak direction encoding $\mathbb{R}^{1 \times H \times W}$ with a spherical Gaussian lobe. 
For each pixel location corresponding to the direction $\mathbf{u}$, we compute the peak direction encoding as:
\begin{equation}
  \text{PeakDirEncoding}(\mathbf{u}) = e^{100(\mathbf{u} \cdot \hat{\mathbf{f}}_\text{dir} - 1)}. 
\end{equation}
We encode the peak intensity into a 3-channel panorama by assigning the peak pixels to the predicted peak intensity $\hat{\mathbf{f}}_\text{intensity}$, 
\begin{equation}
    \text{PeakIntensityEncoding} (\mathbf{u}) = \left\{
    \begin{aligned}
        &\hat{\mathbf{f}}_\text{intensity},      & \ \ \text{if} \ \text{PeakDirEncoding}(\mathbf{u}) \ge 0.98 \\
        &0,     & \text{Otherwise}
    \end{aligned}
    \right. 
    \label{eq:PeakIntensityEncoding}
\end{equation}
This results in 7-channel input to the 2D UNet by concatenating the 1-channel peak direction encoding, 3-channel peak intensity encoding, and the 3-channel positional encoding used in the sky encoder. 
The sky latent code $\hat{\mathbf{f}}_\text{latent}$ is concatenated with the latent vector output by the 2D UNet to jointly decode the HDR sky dome. 

The sky encoder contains two separate 2D CNNs with one predicting the peak information and the other predicting the latent code, where each CNN contains 5 downsampling conv-blocks with intermediate output channel dimensions of: $(64, 128, 256, 256, 256)$. 
For the sky decoder, the 2D UNet contains 5 downsampling conv-blocks and 5 upsampling conv-blocks, connected with residual links~\cite{resnet}. The intermediate output channel dimensions are $(64, 128, 256, 256, 256, 256, 128, 64, 32, 16)$. 
Each conv-block contains two 2D convolution layers followed by batch normalization and ReLU activation. 

\begin{figure*}[t!]
\centering
\begingroup
\includegraphics[width=0.8\linewidth]{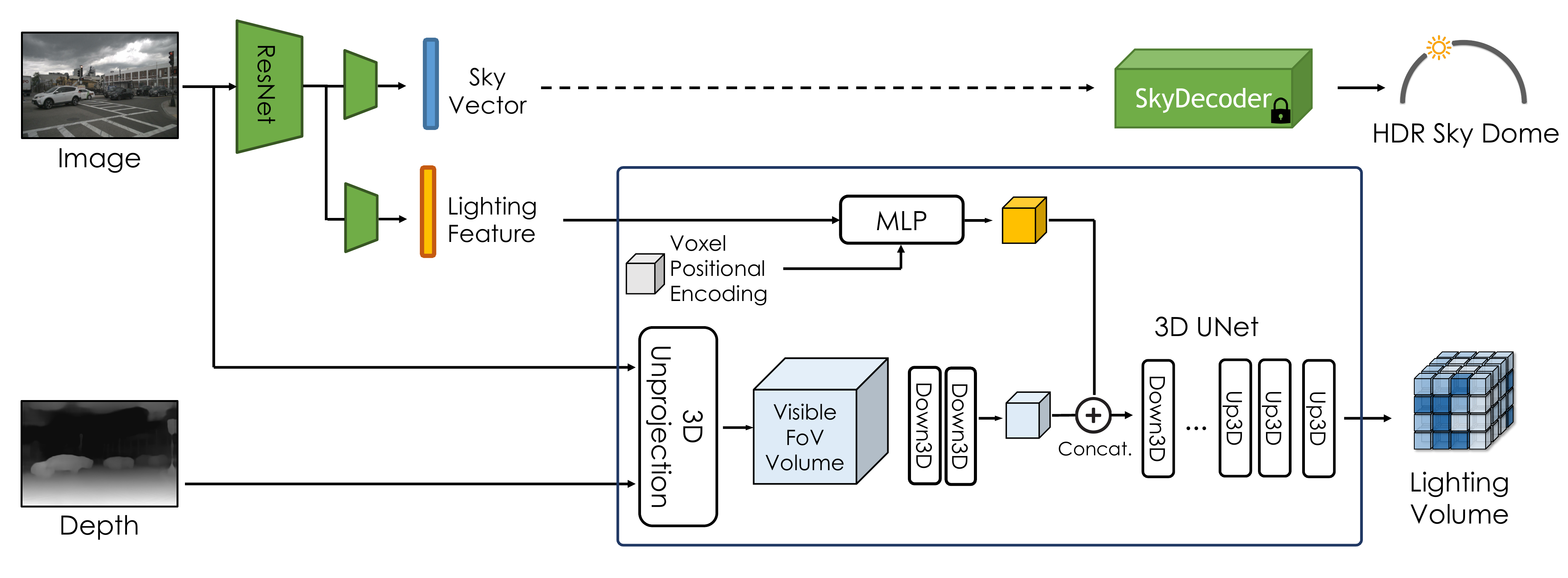} 
\endgroup
\caption{\textbf{Architecture of our hybrid lighting prediction network.} 
For HDR sky dome prediction (top), we directly predict the sky feature vector from the ResNet backbone. 
For lighting volume prediction (bottom), inspired by \cite{wang2021learning}, we unproject the 2D input image into a 3D volume and process it with a 3D UNet. The global lighting feature is converted to a feature volume with an MLP and later fused into the 3D UNet. 
}
\label{fig:lighting_arch} 
\end{figure*} 

\subsection{Hybrid Lighting Prediction Architecture} 

The 2D CNN backbone for the Hybrid Lighting Joint Prediction module (depicted in Figure~\ref{fig:lighting_arch} and main paper Figure~\paperref{2(a)}) is ResNet50~\cite{resnet}, with two separate branches predicting the sky feature vector and the scene lighting feature. 

For the sky prediction branch, the sky feature vector $(\mathbb{R}^{64})$ is then passed into the pre-trained sky decoder, with frozen weights, to decode the HDR sky dome $(\mathbb{R}^{3\times64\times256})$. 

We adapt the architecture used in~\cite{wang2021learning} for the lighting volume prediction branch. 
To encode the visible field-of-view (FoV) information, we unproject the input image into the initial visible surface volume $(\mathbb{R}^{4\times64\times256\times256})$~\cite{wang2021learning}. 
The scene lighting feature $(\mathbb{R}^{128})$ extracted with the ResNet backbone is passed into a coordinate MLP~\cite{mescheder2019occupancy}, and decoded into a global scene feature volume $(\mathbb{R}^{32\times16\times64\times64})$. Here, the coordinate network~\cite{mescheder2019occupancy} contains three residual MLP blocks and the hidden size is $64$. 
We use 3D UNet~\cite{unet} to process the visible surface volume $(\mathbb{R}^{4\times64\times256\times256})$, which contains five downsampling and upsampling conv-blocks with residual connections~\cite{resnet}, and each conv-block contains two 3D convolution layers. The global scene feature volume $(\mathbb{R}^{32\times16\times64\times64})$ is fused into the 3D UNet after 2 downsampling conv-blocks. 
The intermediate output channels of each conv-block have dimensions of $(12, 16, 32, 128, 256, 256, 128, 64, 32, 16)$. The final conv-layer produces the lighting volume prediction $(\mathbb{R}^{8\times64\times256\times256})$.

\subsubsection{Pre-trained depth estimation.} 
In this work, we rely on the existing state-of-the-art off-the-shelf monocular depth estimator PackNet~\cite{packnet} to obtain the 2.5D geometry of the scene. We empirically find that the depth perception is robust with reasonable performance, and provide analysis on the influence of depth prediction error below. 

The predicted depth is first used in lighting volume prediction, where the scene pixel values are unprojected into 3D with the depth prediction. More accurate depth prediction informs more precise geometric information of the scene, and will improve the performance of lighting estimation, \eg the quantitative metric in main paper Table~\paperref{2}. 

Then, during differentiable object insertion, we use depth to decide on the 3D placement of the object we want to insert. Since the location of the inserted object is computed from the depth map, the scale error of the depth prediction leads to error in the size of the inserted objects. To address this, we use the projected sparse LiDAR points as ground-truth depth to rescale the estimated depth by minimizing the L2 error between them. 

When rendering shadows of the inserted object, we rely on the pre-computed depth to compute the 3D location of each scene pixel, which is needed to perform the ray-mesh queries. 
Thus, errors in the estimated depth may result in incorrect shadows. We empirically find that this error is usually negligible as we insert mostly on flat surfaces. 

Prior works that contain submodules with depth prediction~\cite{li2020inverse,wang2021learning} typically do not focus on improving depth perception, but simply use synthetic data with paired ground-truth to train the depth estimation branch, which may easily suffer from the domain gap. 
With a focus on lighting estimation and realistic object insertion, we believe that using the existing mature depth prediction models as a building block is a plausible design choice and does not decrease our technical contribution.

\begin{figure}[t!]
\centering
\setlength{\tabcolsep}{0pt}
\includegraphics[width=0.99\linewidth]{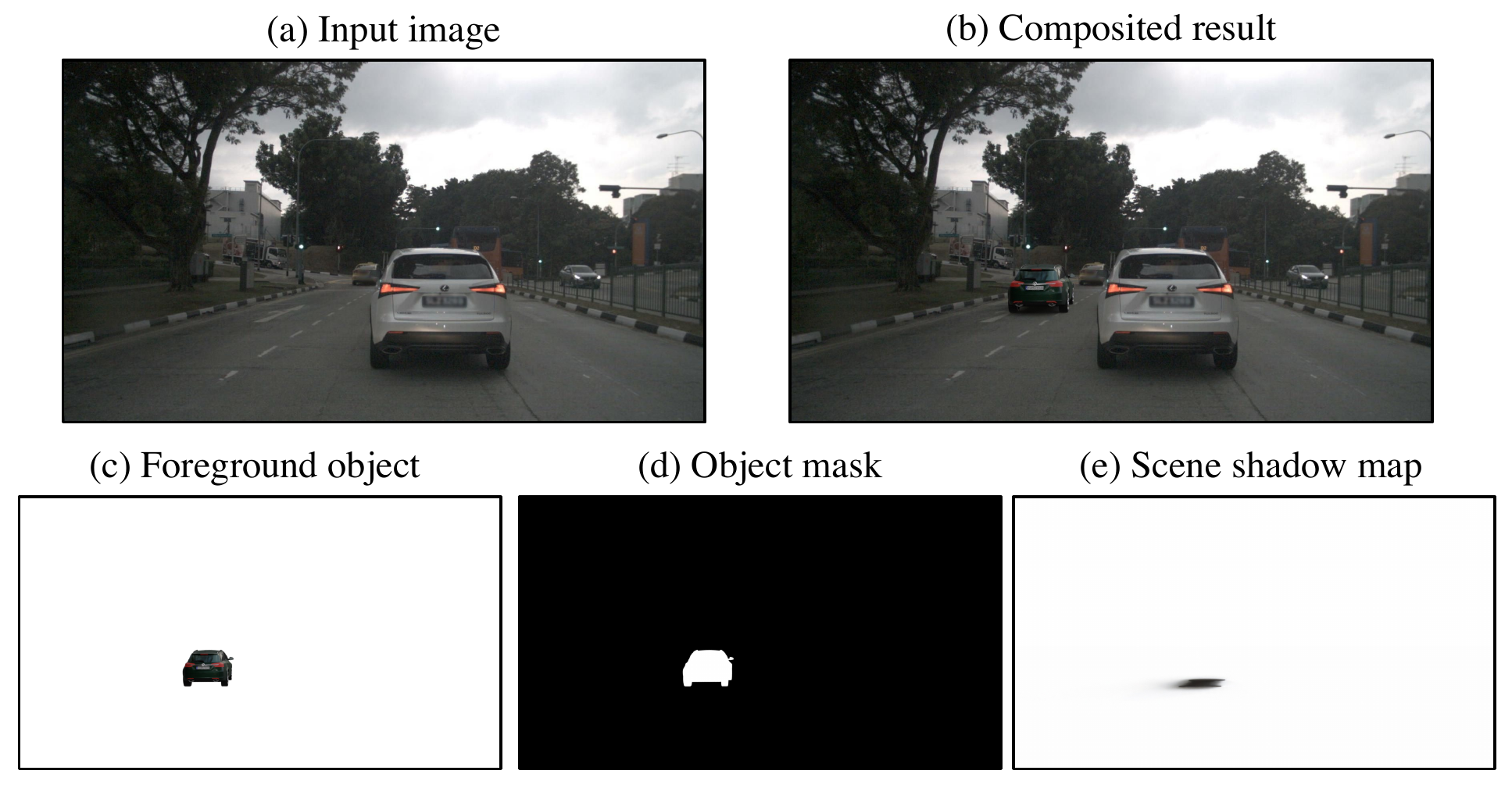} 
\caption{\textbf{Visualization of object insertion composition.} 
Given an input image (a), we estimate our hybrid lighting representation. 
We render the foreground inserted object with standard deferred rendering~\cite{chen2021dibrpp} and obtain the rendering result (c) and object mask (d). 
We render the shadow cast by the inserted object into a ratio map (e) as described in main paper Section~\paperref{3.3}. 
The final editing result (b) is produced by compositing (c-e). 
} 
\label{fig:insertion_buffer} 
\end{figure}

\subsection{Differentiable Object Insertion }

We visualize the object insertion composition process in Figure~\ref{fig:insertion_buffer}. 
We composite the input image $I$, foreground object $I_\text{object}$, alpha mask $M$, scene shadow map $I_\text{shadow}$ into the final editing result $I_\text{edit}$ by
\begin{equation}
    I_\text{edit} = M \odot I_\text{object}  + (1-M) \odot I \odot I_\text{shadow}. 
\end{equation}
We include the implementation details below.

\subsubsection{Rendering details.} 
For rendering the foreground object, we use the BRDF used by Unreal Engine 4, which is a simplified version of Disney BRDF~\cite{burley2012physically,karis2013real}. 
Specifically, we use the base color $c_\text{base} \in \mathbb{R}^3$, metallic $m \in [0, 1]$, roughness $r \in [0, 1]$ and specular $s \in [0, 1]$ to describe material properties of object surfaces. The BRDF is defined as
\begin{equation}
  f(\mathbf{l}, \mathbf{v}) = \frac{c_\text{diffuse}}{\pi} + \frac{DFG}{4(\mathbf{n}\cdot \mathbf{l})(\mathbf{n}\cdot \mathbf{v})},
\end{equation}
where 
\begin{align}
  c_\text{diffuse} &= (1-m)\, c_\text{base} \\
  c_\text{specular} &= (1-m)\, 0.08 s + m\, c_\text{base} \\
  D &= \frac{\alpha^2}{\pi((\mathbf{n}\cdot \mathbf{h})^2 (\alpha^2-1) + 1)^2} \\
  G &= \frac{(\mathbf{n}\cdot \mathbf{l})(\mathbf{n}\cdot \mathbf{v})}{((\mathbf{n}\cdot \mathbf{l})(1-k) + k)((\mathbf{n}\cdot \mathbf{v})(1-k) + k)} \\
  F &= c_\text{specular} + (1 - c_\text{specular}) 2^{(-5.55473(\mathbf{v}\cdot \mathbf{h}) - 6.98316)(\mathbf{v}\cdot \mathbf{h})} \\
  \alpha &= r^2, k = \frac{(r+1)^2}{8}, \mathbf{h} = \frac{\mathbf{l}+\mathbf{v}}{||\mathbf{l}+\mathbf{v}||}.
\end{align}
The rendered raw pixel values is HDR in linear RGB space. 
To convert to LDR sRGB images, we apply gamma correction ($\gamma=2.2$), and do soft clipping following~\cite{wang2021learning} 
\begin{equation}
    \varphi (x) = \left\{
    \begin{aligned}
        &x                                             & \text{if} \ x \le \tau \\
        &1-(1-\tau) e^{-\frac{x-\tau}{1-\tau}}     & \text{if} \ x > \tau 
    \end{aligned}
    \right. 
    \label{eq:hdr2ldr}
\end{equation}
where we set $\tau = 0.95$. 


During training, to save on the computational cost and GPU memory, we uniformly sample $5000$ rays for \emph{object center} alone, and render the foreground object on a 320x180 image canvas. 
For background shadows, we render a 160x90 resolution shadow map. We sample $50$ rays for each scene pixel, and aggregate the 8-neighbors of each pixel to obtain $450$ rays in total. 
The average rendering time for foreground objects and background shadows are 0.2s and 2s respectively. 
The current efficiency bottleneck in the renderer is the ray-mesh query function, which is a CPU implementation using trimesh\footnote{\url{trimsh.org}} with the potential to further speedups. 
It consumes 12G GPU memory during training, including both the forward and the backward pass. 
During inference, we sample rays for \emph{each object pixel} using per-pixel importance sampling following~\cite{chen2021dibrpp} to achieve high quality rendering, where we sample 1024 rays for the diffuse component and 256 rays for the specular component.

\subsubsection{Ray sampling scheme.} 
\begin{figure}[t]
\centering
\setlength{\tabcolsep}{0pt}
\includegraphics[width=0.85\linewidth]{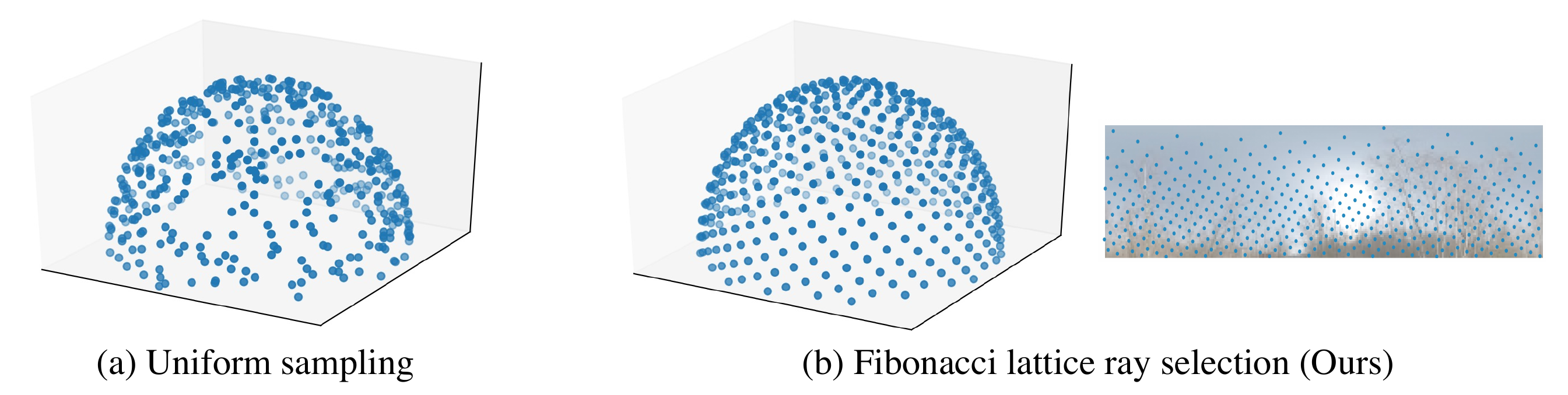} 
\caption{\textbf{Visualization of ray directions.} 
Instead of using naive uniform ray sampling (a), we select equi-spaced rays with Fibonacci lattice (b). This scheme better utilizes the rays and can ensure the sun light is properly sampled. 
} 
\label{fig:rebuttal_ray} 
\end{figure}

A larger number of ray samples may lead to higher quality rendering, but usually limited by the affordable memory and computation, especially for a differentiable rendering module in end-to-end learning tasks. 
To improve the efficiency of ray sampling especially for the process of shadow rendering, we select equi-spaced rays for each pixel on the upper-hemisphere with Fibonacci lattice~\cite{gonzalez2010measurement} to render the spatially-varying shadows. 
As shown in Figure~\ref{fig:rebuttal_ray}, the Fibonacci lattice ray selection strategy better utilizes the rays compared to naive uniform sampling.

\subsubsection{Object insertion details.} 
During training, we use the task of object insertion to provide additional supervision. Specifically, we differentiably insert a virtual object into the photograph and encourage the photorealism of the final image editing results. 
We adopt a relatively conservative scheme to avoid unrealistic editing results due to asset quality and object placement. 
For assets, we collect a set of 283 high quality 3D car models from Turbosquid\footnote{\url{www.turbosquid.com}}. 
To place the cars into plausible locations, we use the dense depth map prediction~\cite{packnet}, lidar semantic segmentation and 3D bounding box annotation in nuScenes~\cite{nuscenes2019}. The location candidates for insertion should satisfy the following conditions: 
(1) Semantics: belong to the semantic class ``driveable surface'', 
(2) Distance: within the range of 10 to 40 meters, 
(3) Collision: candidate location is 1 meter away from the static background classes (such as ``sidewalk'') and 3 meters away from dynamic classes (such as ``vehicles''), 
and (4) Occlusion: the 3D bounding box corners of the inserted object do not get occluded by other objects. 
The car orientation is randomly set to the same or opposite direction as the ego camera, with a Gaussian random perturbation in the yaw angle (sigma value set to $3^\circ$). 

\subsubsection{Object insertion for data augmentation.}  For downstream tasks, we use object insertion as a data augmentation approach, to enhance the data with additional diversity.
We use the 283 car CAD models and randomize the car paint with 60 diverse colors. For diversity, we also include 30 high quality construction vehicle CAD models that are rarely observed on the street. 
We remove the occlusion check mentioned above, and naively handle the occlusion by comparing the depth map of the scene and inserted object. 
For orientation, we increase the Gaussian perturbation with sigma value set to $15^\circ$.

\begin{figure}[t]
\centering
\setlength{\tabcolsep}{0pt}
\includegraphics[width=0.99\linewidth]{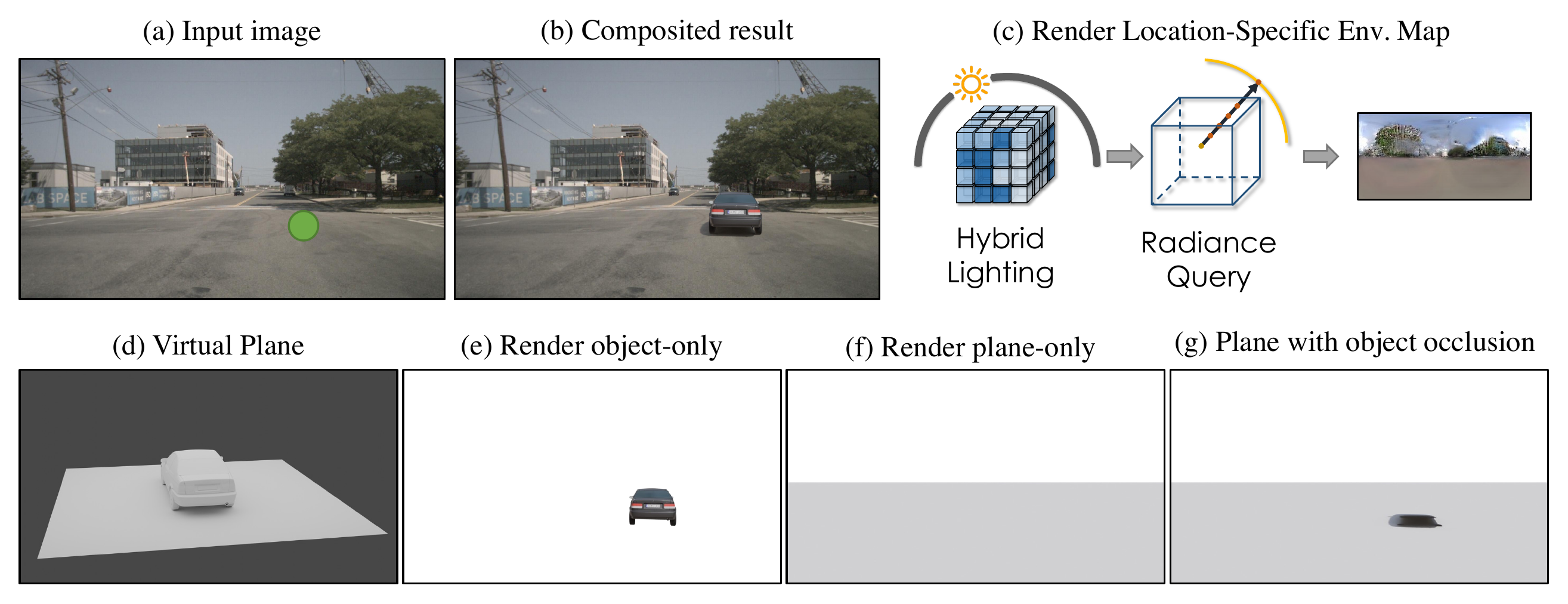} 
\caption{\textbf{Rendering object insertion with Blender.} 
Our lighting representation can be converted to the standard environment map to be compatible with commercial renderers (thus gaining the benefits of richer material support and speed of commercial renderers). 
Given the input image (a) and the 3D location shown in green to insert object, we first convert our hybrid lighting into a local environment map with volume rendering (c). 
We place a virtual plane under the inserted object (d), and render three images -- Object-only (e), plane-only (f) and plane with object occlusion (g). The shadow map can be computed by the ratio of (f) and (g). 
The final editing result is shown in (b). 
} 
\label{fig:blender_insertion} 
\end{figure} 

\subsubsection{Rendering object insertion with commercial renderer.} 
While we adopt our proposed custom differentiable object insertion module during training to provide valuable supervision signal, our rendering is fully physics-based with standard PBR material definition. 
Thus, our estimated lighting can also be made useable in commercial rendering engines such as the Cycles renderer in Blender~\cite{blender}. This allows us to utilize faster rendering as well as richer material support available in the commercial renderers.

Commercial renderers usually only support standard environment map lighting and do not support spatially-varying lighting such as our hybrid lighting representation. In addition, the radiance-level API is not available for shadow rendering. 
We modify our object insertion pipeline to be compatible with Blender Cycles renderer in Figure~\ref{fig:blender_insertion}. 
Given the 3D location to insert the virtual object, we first render our lighting representation into a standard HDR environment map by doing radiance query at the specified 3D location (main paper Eq.~\paperref{1}), and load into Blender as lighting. 
To render cast shadows, we insert a virtual plane beneath the inserted object, and render three images: 
(1) render only inserted object $I_\text{object}$, 
(2) render only virtual plane $I_\text{plane}$, 
(3) render the virtual plane but take into account of the inserted object when doing ray-tracing $I_\text{shadowed}$. 
The shadow map can be computed as $I_\text{shadow} = \frac{I_\text{shadowed}}{I_\text{plane}}$, followed by the image composition in Figure~\ref{fig:insertion_buffer}. 
Note that this maximally preserves scene-level spatially-varying effects, but still ignores the local geometry of object and virtual plane, and thus cannot render localized shadow effects as shown in main paper Figure~\paperref{6}. 

We show qualitative results from Blender in Figure~\ref{fig:qual_blender}. In most cases, with the Cycles renderer at inference time, we can benefit from more bounces of ray-tracing and complex materials such as transparency and sub-surface scattering.

\subsection{Training Details} 

\subsubsection{Sky modeling network.} 
We use a multi-task loss to train the sky encoder-decoder network: 
\begin{equation}
  \mathcal{L}^\text{skym} = \lambda_{\text{dir}} \mathcal{L}_{\text{dir}}^\text{skym} + \lambda_{\text{intensity}} \mathcal{L}_{\text{intensity}}^\text{skym} + \lambda_{\text{hdr}} \mathcal{L}_{\text{hdr}}^\text{skym}
\end{equation}
where the weights are all set to 1. 
During training, we introduce ``teacher forcing'' on HDR reconstruction loss $\mathcal{L}_{\text{hdr}}$ by alternating the input to sky decoder between $(\hat{\mathbf{f}}_\text{dir}, \hat{\mathbf{f}}_\text{intensity}, \hat{\mathbf{f}}_\text{latent})$ and $({\mathbf{f}}_\text{dir}, {\mathbf{f}}_\text{intensity}, \hat{\mathbf{f}}_\text{latent})$. 
This is because the prediction of the peak direction $\hat{\mathbf{f}}_\text{dir}$ and peak intensity $\hat{\mathbf{f}}_\text{intensity}$ are usually inaccurate in the early stages of training. This also helps to efficiently disentangle the peak and the background (Figure~\ref{fig:qual_skymodel}), encouraging the network to accurately reconstruct HDR peak when given groundtruth peak information ${\mathbf{f}}_\text{dir}, {\mathbf{f}}_\text{intensity}$. 
We use the Adam optimizer~\cite{kingma2017adam} and train for 4000 epochs, with the learning rate set to 1e-3 and decaying by $0.3$ every 1000 epochs.

\subsubsection{Lighting prediction network.} 
We use a weighted sum of loss terms introduced in the main paper in Section~\paperref{3.4}, including the sky regression loss $\mathcal{L}_\text{sky}$, radiance and depth reconstruction loss $\mathcal{L}_\text{recon}$, sky separation loss $\mathcal{L}_\text{transmit}$, and adversarial supervision $\mathcal{L}_{\text{adv}}$. 
To address the ambiguity of depth rendering, we also follow~\cite{wang2021learning} and use a regularization loss $\mathcal{L}_{\text{reg}}$ to encourage the alpha channel of the volume to be either 0 or 1. 
The final loss function 
\begin{equation}
  \mathcal{L}^\text{hybrid} = 
  \lambda_\text{sky} \mathcal{L}_\text{sky} + 
  \lambda_\text{recon} \mathcal{L}_\text{recon} + 
  \lambda_\text{reg} \mathcal{L}_\text{reg} + 
  \lambda_\text{transmit} \mathcal{L}_\text{transmit} + 
  \lambda_\text{adv} \mathcal{L}_\text{adv} 
\end{equation}
where we set $\lambda_\text{sky}, \lambda_\text{recon}, \lambda_\text{transmit}$ to 1, $\lambda_\text{reg}$ to 1e-4, and $\lambda_\text{adv}$ to 3e-3. 
For the sky regression loss $\mathcal{L}_\text{sky}$, we linearly fade out the L1 loss for latent code in the first 50 epochs. 
For the adversarial supervision $\mathcal{L}_{\text{adv}}$, 
the discriminator $\mathcal{D}$ is a 5-layer PatchGAN with spectral norm~\cite{pix2pix2017,park2019SPADE}.
We use hinge loss for the discriminator 
\begin{align}
    \mathcal{L}^{\text{D}} =& \max(0, 1-\mathcal{D}(I_\text{real})) + \max(0, 1+\mathcal{D}(\hat{I}_\text{edit})). 
\end{align}
where we use the training set of real world images $\{I_\text{real}\}$ from nuScenes~\cite{nuscenes2019} and perspective image crops from HoliCity street views~\cite{zhou2020holicity} as positive examples. 
We set the batch size to 1. The full model takes 20G GPU memory during training (8G for network inference and 12G for rendering). 
We first pre-train without $\mathcal{L}_\text{adv}$ for 50 epochs and then jointly train another 50 epochs. We train with the Adam optimizer~\cite{kingma2017adam} with learning rate of 3e-4, decaying by $0.3$ every 30 epochs.

\subsection{Experimental Settings} 

\subsubsection{Data processing.} 
We train our model on nuScenes~\cite{nuscenes2019}, HoliCity \cite{zhou2020holicity}, and a set of 724 outdoor HDR panoramas collected from three data sources: HDRIHaven\footnote{\url{polyhaven.com/hdris} (License: CC0)}, DoschDesign\footnote{\url{doschdesign.com} (License: \url{doschdesign.com/information.php?p=2})} 
and HDRMaps\footnote{\url{hdrmaps.com} (License: Royalty-Free)}.
For HDRI data, we use 90\% and 10\% for training and evaluation. During training, we apply random flipping and random azimuth shifting as data augmentation. 
For nuScenes, we use the official split containing 700 scenes for training and 150 scenes for evaluation. For each key frame, we take the front camera as the input image and use the captured views at a novel viewpoint (1.5 seconds after the input frame) to supervise lighting. 
For HoliCity dataset, we follow~\cite{hold2017deep} and detect the sun location as the centriod of largest connected region after thresholding with the 98-th percentile. 
For evaluation, we manually annotated the sun location for the test set. 
Considering that the peak direction is ambiguous for an almost uniform sky dome, we only use a subset of the HoliCity data with a strong peak to train and evaluate the peak direction prediction. Specifically, we pass the LDR panoramas to the pre-trained sky encoder and get the intensity prediction $\tilde{\mathbf{f}}_\text{intensity}$. Loss is only used when the peak intensity is greater than 10. This results in 1897 panoramas for training and 183 panoramas for evaluation.

\subsubsection{Human perception study details.} 

\begin{figure*}[t!]
\centering
\includegraphics[width=\linewidth]{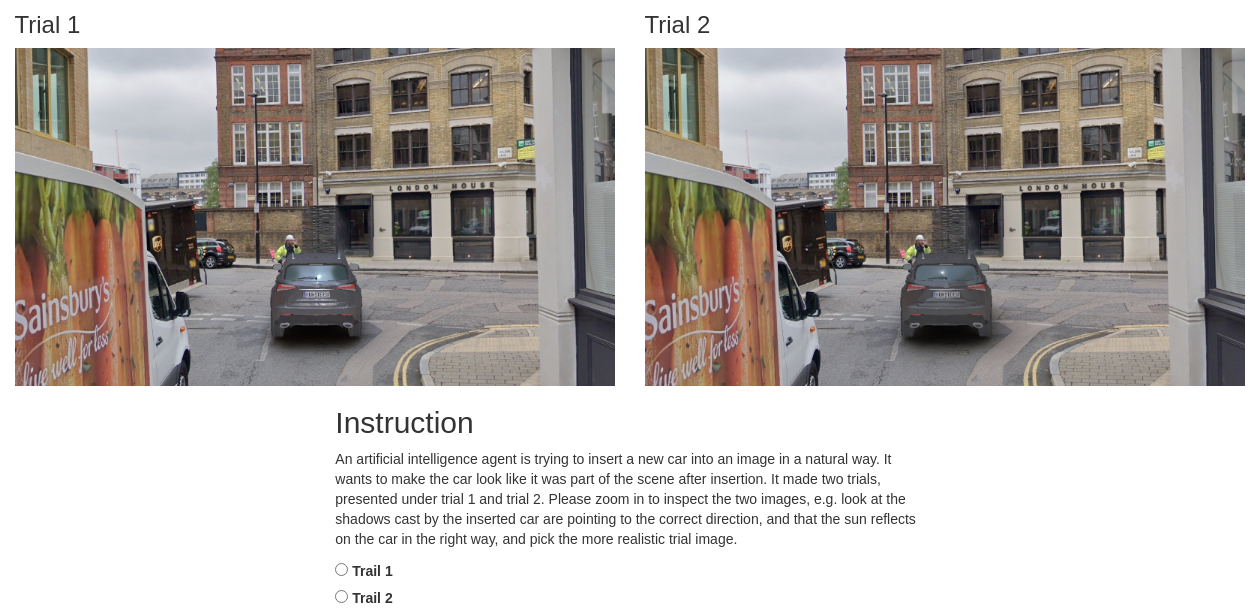}
\caption{\textbf{User Study Interface (AMT):} We insert two cars, which are rendered with different lighting approaches, into the same background image, and ask a human participant to select the more realistic one. We conduct 3 comparisons, where we first compare our full model with the baseline lighting methods~\cite{wang2021learning,hold2019deep}, then ablate against our method without adversarial supervision. We randomize their order in each HIT. 
}
\label{fig:userstudy} 
\end{figure*}

In this section, we provide additional details of our user study. We perform the user study on Amazon Mechanical Turk (AMT) and visualize the interface in Figure~\ref{fig:userstudy}. In each HIT, we provide two images and ask the user to select the more realistic one. Among the two images, one is relighted by our approach and the other is either one of the baselines~\cite{wang2021learning,hold2019deep} or the ablation setting (Ours w/o adversarial supervision). To avoid bias in order, we always randomize which of the methods is shown on the left vs right. We provide instructions as follows:

\emph{An artificial intelligence agent is trying to insert a new car into an image in a natural way. It wants to make the car look like it was part of the scene after insertion. It made two trials, presented under trial 1 and trial 2. Please zoom in to inspect the two images, e.g. look at the shadows cast by the inserted car are pointing to the correct direction, and that the sun reflects on the car in the right way, and pick the more realistic trial image.}

We synthesize 23 insertion examples. For each example, we ask 15 users to judge the realism of the inserted object and adopt a majority vote to compute the final preference. In summary, it results in $3 \text{ comparisons} \times 23 \text{ examples} \times 15 \text{ selections} \times 3 \text{ repeat experiments} = 3105 \text{ HITs}$. We provide results in Table \paperref{3} in the main paper, demonstrating that users prefer our approach over the baselines~\cite{wang2021learning,hold2019deep}, and adding adversarial supervision further improves the realism of object insertion.

\section{Additional Results} 
\label{sec:results}

\subsection{Sky modeling} 

The sky encoder-decoder takes as input an LDR sky panorama, and produces a low-dimensional vector representation of the sky $\hat{\mathbf{f}}$ and a reconstructed HDR sky dome. 
The qualitative results are shown in Figure~\ref{fig:qual_skymodel}. 
Our model can accurately reconstruct HDR sky, especially the peaks with extreme intensity values. 
As the sky vector $\hat{\mathbf{f}} = (\hat{\mathbf{f}}_\text{intensity}, \hat{\mathbf{f}}_\text{dir}, \hat{\mathbf{f}}_\text{latent})$ contains explicit peak information, it enables peak editing by feeding into edited sky vector $(\hat{\mathbf{f}}_\text{intensity}^\text{edit}, \hat{\mathbf{f}}_\text{dir}^\text{edit}, \hat{\mathbf{f}}_\text{latent})$. 
As the results shown, our sky decoder can efficiently disentangle the HDR peak from the background, and enables precise peak editing. The controllable property allows for potential post-editing of inaccurate predictions.

We quantitatively ablate our architecture design and show the results in Table~\ref{table:quant_skymodel}. 
We evaluate by comparing the MSE of reconstructed HDR sky dome and the ground truth, and the median angular error between the reconstructed peak and the ground-truth peak direction. 
For the ablated version, ``ours w/o encoding'' removes the positional encoding and peak encoding concatenated to the Sky Encoder and Sky Decoder, and ``ours w/o peak information'' removes the peak direction and peak intensity and only relies on the latent code to reconstruct HDR sky. The quantitative results validated the effectiveness of our design choices to achieve the best performance. 
We also compare under the same experiment setting with the sky representation used in \cite{hold2019deep}, where the HDR sky is represented with a latent vector and explicit azimuth of the sun. 
Our method outperforms \cite{hold2019deep}, which shows the benefits from the explicit representation and supervision of peak direction and intensity.

\begin{table}[ht]
\centering
\resizebox{\linewidth}{!}{
\begin{tabular}{|l|c|c|}
\hline
Method & HDR reconstruction MSE $(\times10^{-2})$ $\downarrow$ & Peak direction median angular error $\downarrow$ \\
\hline
Latent code w/ explicit azimuth~\cite{hold2019deep} & $10.56$ 	    & $5.67^\circ$   \\
Ours                                                & $\bm{7.49}$ 	& $\bm{3.38}^\circ$   \\
Ours (w/o encoding)	                                & $8.31$        & $4.09^\circ$   \\
Ours (w/o peak information)	                        & $11.01$       & $7.93^\circ$   \\
\hline
\end{tabular}
} 
\vspace{1mm} 
\caption{\textbf{Quantitative results of sky modeling. } We compare the HDR reconstruction MSE and the median angular error of the reconstructed peak direction. 
} 
\label{table:quant_skymodel} 
\end{table}

\subsection{Auxiliary Quantitative Evalutation of Lighting Estimation} 

Prior works~\cite{zhang2019all} that ignore spatially-varying effects may also enforce that a known rendered virtual scene $\mathcal{O}_\text{scene}$ (\eg a sphere) appears consistent when using the predicted envenvironment map and groundtruth envenvironment map, i.e., $||\text{Render}(L_\text{pred}, \mathcal{O}_\text{sphere}) - \text{Render}(L_\text{gt}, \mathcal{O}_\text{sphere})||$. 
Compared to the direct envenvironment map regression loss $||L_\text{pred} - L_\text{gt}||$, this is still regression but smartly weighted. 
This can be used as both training loss and evaluation metric, when prior works simplify the lighting to only one environment map. 

However, for spatially-varying lighting estimation which is the focus of our work, there is no groundtruth lighting provided for the specific 3D location of the inserted object, and thus we cannot directly utilize this as a training loss. 

Despite not a precise metric, we show the quantitative evaluation of this metric as an auxiliary result. 
Specifically, we take a cropped perspective image of groundtruth HDR panorama as input, and insert a diffuse or specular sphere with the estimated lighting. We report MSE between insertion results generated with predicted lighting and groundtruth HDR panorama, as shown in Table.~\ref{table:rebuttal_quantitative_insertion}. 
Hold-Geoffroy \etal~\cite{hold2019deep} ignores spatially-varying effects and usually cannot recover the high-frequency details, and thus lead to inferior performance on specular sphere insertion. 
Wang \etal~\cite{wang2021learning} cannot reconstruct the HDR component well and especially suffers when inserting a diffuse sphere. 
Our method outperforms baselines with better quantitative performance. 

\begin{table}[ht]
\centering
\resizebox{0.75\linewidth}{!}{
\addtolength{\tabcolsep}{6pt}
\begin{tabular}{|l|c|c|}
\hline
Rendering MSE ($\times 10^{-3}$) $\downarrow$       & Diffuse Sphere   & Specular Sphere \\
\hline 
Hold-Geoffroy \etal~\cite{hold2019deep}  & 2.30   & 4.44  \\ 
Wang \etal~\cite{wang2021learning}           & 3.36   & 3.13  \\
Ours                                 & \textbf{1.79}   & \textbf{2.41}  \\
\hline
\end{tabular}
} 
\vspace{1mm} 
\caption{\textbf{Rendering error of inserted objects.} 
Note that this evaluation ignores spatially-varying effects. 
} 
\label{table:rebuttal_quantitative_insertion} 
\end{table}

\subsection{Quantitative Ablation of Multi-view Input} 
While we focus on monocular estimation, our model is extendable to multi-view input, such as the six surrounding perspective cameras in nuScenes sensor rig~\cite{Ost_2021_CVPR}. 
For extremely challenging cases such as predicting precise shadow boundaries, multi-view input images can provide more field-of-view information and predict more accurate lighting, as shown in main paper Figure~\paperref{6}. 
We additionally provide quantitative result in Table~\ref{table:quantitative_multiview}, following the experiment settings in main paper Table~\paperref{1, 2}. 
Although the six surrounding views still only cover a subset of the panorama, it can significantly improve upon monocular lighting prediction. 

\begin{table}[ht]
\centering
\resizebox{0.8\linewidth}{!}{
\addtolength{\tabcolsep}{6pt}
\begin{tabular}{|l|c|c|c|}
\hline
Method          & \tabincell{c}{HoliCity~\cite{zhou2020holicity} sun location \\ Median angular error $\downarrow$}   & \tabincell{c}{nuScenes~\cite{nuscenes2019} \\ PSNR $\uparrow$} & \tabincell{c}{nuScenes~\cite{nuscenes2019} \\ si-PSNR $\uparrow$} \\
\hline 
Ours            & $22.43^\circ$          & $14.49$      & $15.35$      \\
Ours (6 views)  & $\bm{19.91}^\circ$     & $\bm{17.96}$ & $\bm{18.45}$  \\
\hline
\end{tabular}
} 
\vspace{1mm} 
\caption{Quantitative ablation study of multi-view input.
} 
\label{table:quantitative_multiview} 
\end{table}

\subsection{Analysis of the Adversarial Supervision} 
The adversarial supervision (main paper Section~\paperref{3.4}, \emph{Training Lighting via Object Insertion}) uses a discriminator to encourage the photorealism of the lighting-aware image editing results, where this signal is backpropagated to the predicted lighting through the Differentiable Object Insertion module (main paper Section~\paperref{3.3}).

To understand the ``photorealism'' implicitly perceived by the discriminator during the training process, we perform test-time optimization on the object insertion results and visualize the optimization process (main paper Figure~\paperref{7}). 
Specifically, we take the network weights of the lighting prediction network $\Theta$ and the discriminator $\mathcal{D}$ in the middle of the training process (10-th epoch), and optimize the lighting prediction network weights $\Theta$ to minimize the discriminator loss $\mathcal{L}_\text{adv}=-\mathcal{D}(\hat{I}_\text{edit})$.
Note that the discriminator and the pre-trained sky decoder are kept frozen. 

We use Adam optimizer~\cite{kingma2017adam} with learning rate 1e-4, and show the optimized results after 5 and 10 iterations in Figure~\ref{fig:qual_adv}. 
We refer to the accompanied video for animated results. 

In high-level, the scene appearance (encoded by image pixel values) is an interaction between scene geometry, material and lighting, where this process can be approximated by the rendering equation. 
With known groundtruth image distribution implicitly learned with a discriminator, we utilize groundtruth material and geometry to supervise lighting prediction, by inserting artist-designed virtual objects into the scene images. 
This objective matches the end goal of our predicted lighting -- the photorealism of image editing results. 
Both quantitative results and qualitative visualization indicate its value as a complementary supervision signal to existing datasets.

\subsection{Improving Downstream Perception with Data Augmentation} 

\subsubsection{Qualitative results of data augmentation. } 
We show the augmented data using our AR pipeline in Figure~\ref{fig:qual_dataaug},~\ref{fig:qual_dataaug_cv},~\ref{fig:qual_dataaug_occlusion},~\ref{fig:qual_dataaug_night}. 
The 3D assets are provided courtesy of TurboSquid and their artists Hum3D, be fast, rabser, FirelightCGStudio, amaranthus, 3DTree\_LLC, 3dferomon and Pipon3D.
Note that the 3D bounding box and orientation of the inserted objects are known and become free labels to train a 3D object detector. 

As shown in Figure~\ref{fig:qual_dataaug}, our data augmentation inserts a diverse set of car assets into captured images from the nuScenes dataset~\cite{nuscenes2019} based on the estimated lighting. The data generation process is fully automatic. 
Our editing results can produce realistic lighting effects, 
such as cast shadows, which requires accurate HDR prediction, and ``clear coat'' reflection on the car body, which requires high-frequency lighting prediction and HDR highlights. 

While many existing image manipulation methods~\cite{chen2021geosim,Ost_2021_CVPR} have underlying assumptions about object categories, 
our method predicts physics-based scene lighting information, and thus is agnostic to the category of the virtual object to insert. 
This enables editing with object classes rarely observed in real world -- a critical use case for data augmentation. 
Figure~\ref{fig:qual_dataaug_cv} shows insertion results of construction vehicles, where our method demonstrates consistent performance. 

In Figure~\ref{fig:qual_dataaug_occlusion}, we show results of occlusion handling. 
We adopt a simple strategy by comparing the depth of the inserted object (cached in G-buffer) with the scene depth map, and take the closer pixel to display. 

While we primarily focus on daytime outdoor scenes, our method can also predict reasonable results for night-time scenes, as shown in Figure~\ref{fig:qual_dataaug_night}. 
Note that our model has no HDR data supervision for this domain, due to the scarcity of ground truth lighting for night time street scenes, and only relies on limited FoV LDR data and the adversarial supervision to learn.

\subsubsection{Quantitative results.} 
We compare against the baseline that uses real-world data only, and a strong baseline that augments virtual objects with a fixed dome lighting. 
In Table~\ref{tab:supp.quantitative_results_downstream}, we can observe  that the performance of the detector improves by $2\%$ comparing to the same detector when trained on real data alone. Moreover, we can also see that while naively adding objects leads to a $1\%$ improvement, another $1\%$ is a result of  having better light estimation. We believe that further improvements are possible with more sophisticated placement strategies as well as using more assets (both for each class, and more classes). 
Interestingly, notice that the performance of the object detector also improves in other object categories even though we do not directly augment those. We attribute this to various factors, including providing the detector with more challenging negatives for other classes, 
 improving the detector's confidence in classes that may get confused with construction vehicles (e.g. bus vs construction vehicles and trailer vs construction vehicles), as well as the way certain objects are annotated. For example,  we notice that cranes and extremities of construction vehicles are only included in nuScenes annotations if they interfere with traffic (see Figure~\ref{fig:nuscenes_cv}), therefore what constitutes a 3D bounding box for that object category is not uniquely defined. We additionally notice that in nuScenes trucks used to hauling rocks or building materials are considered as truck rather than construction vehicles. These factors might also explain why we do not see a big improvement in the class of construction vehicles even though we directly augment it.

\begin{table}[htbp]
\centering
  \resizebox{0.99\linewidth}{!}{
  \addtolength{\tabcolsep}{0pt}
  \centering
  \begin{tabular}{|l|c|c|c|c|c|c|c|c|c|c|c|}
  \hline
  \textbf{Method} &  {  mAP } &   { {car}} &  { {truck}} &  { {bus}} & { {trailer}} &  { {const.veh.}} &    {pedestrian}  &  { {motorcycle}} &  { {bicycle}} &  { {traffic cone}} &  { {barrier}} \\
   \hline
   Real Data   & 0.190  & 0.356 & 0.112 & 0.124 & 0.011 & 0.016 & 0.327 & 0.127 & 0.116 & 0.389 & 0.317 \\
   \hline
   + Aug No Light & 0.201 & 0.363 & \underline{0.149} & 0.163 & 0.029 & \underline{0.021} & 0.311 & 0.146 & 0.120  & 0.394 & 0.309 \\
   +  Aug Light  & \textbf{0.211} & \underline{0.369} & 0.146 & \underline{0.182} & \underline{0.036} & 0.020  & \underline{0.317} & \underline{0.161} & \underline{0.146} & \underline{0.400} & \underline{0.332} \\
  \hline
  \end{tabular}%
}
\vspace{1mm}
\caption{Performance of a SoTA 3D object detector on the nuScenes 3D object detection task. mAP represents the mean for the 10 object categories. We additionally report the average precision (AP) for all categories. \emph{Real Data} corresponds to a subset of the nuScenes training set (10\%). Performance of the detector improves by $2\%$ when comparing to Real Data, and 1\% is due to better lighting estimation.}
\label{tab:supp.quantitative_results_downstream}%
\end{table}%

\begin{figure}[htbp]
\footnotesize
\centering
\begingroup
\setlength{\tabcolsep}{1pt}
\resizebox{0.998\linewidth}{!}{
\begin{tabular}{cc}
(a) Model Trained on Real Data & (b) Model Trained on Real Data + Lighting Aug \\
\includegraphics[width=0.6\linewidth]{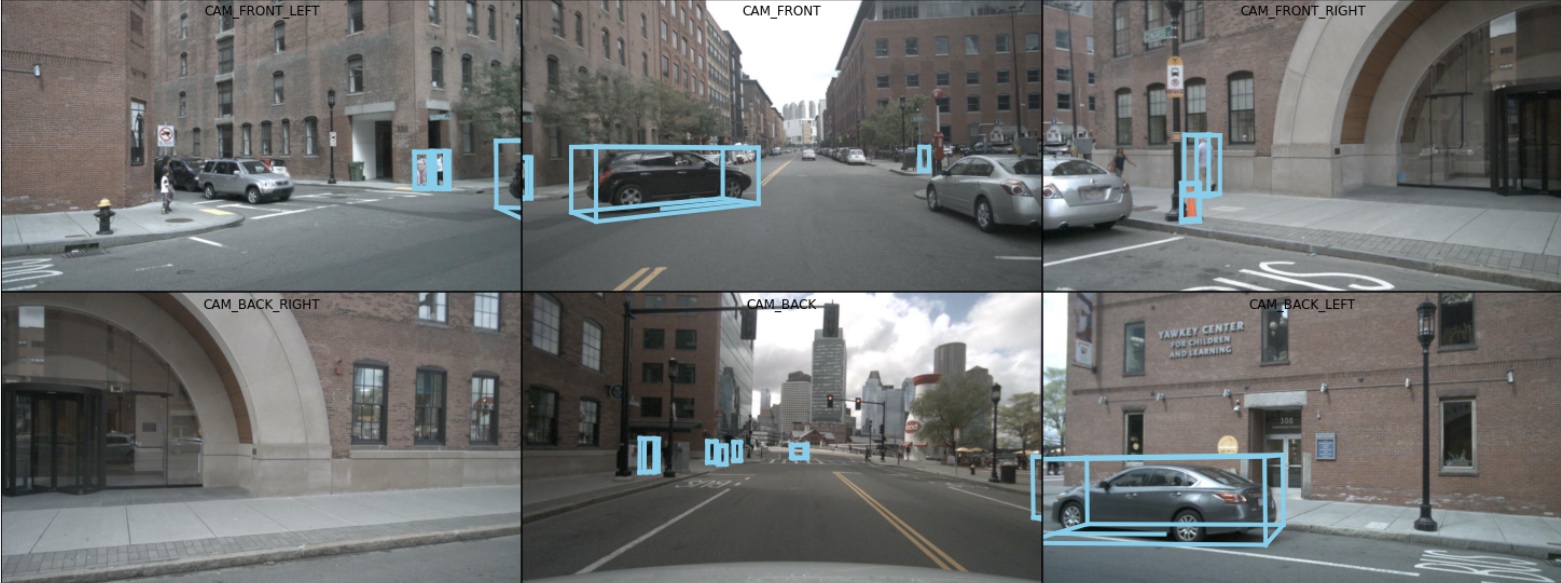} & 
\includegraphics[width=0.6\linewidth]{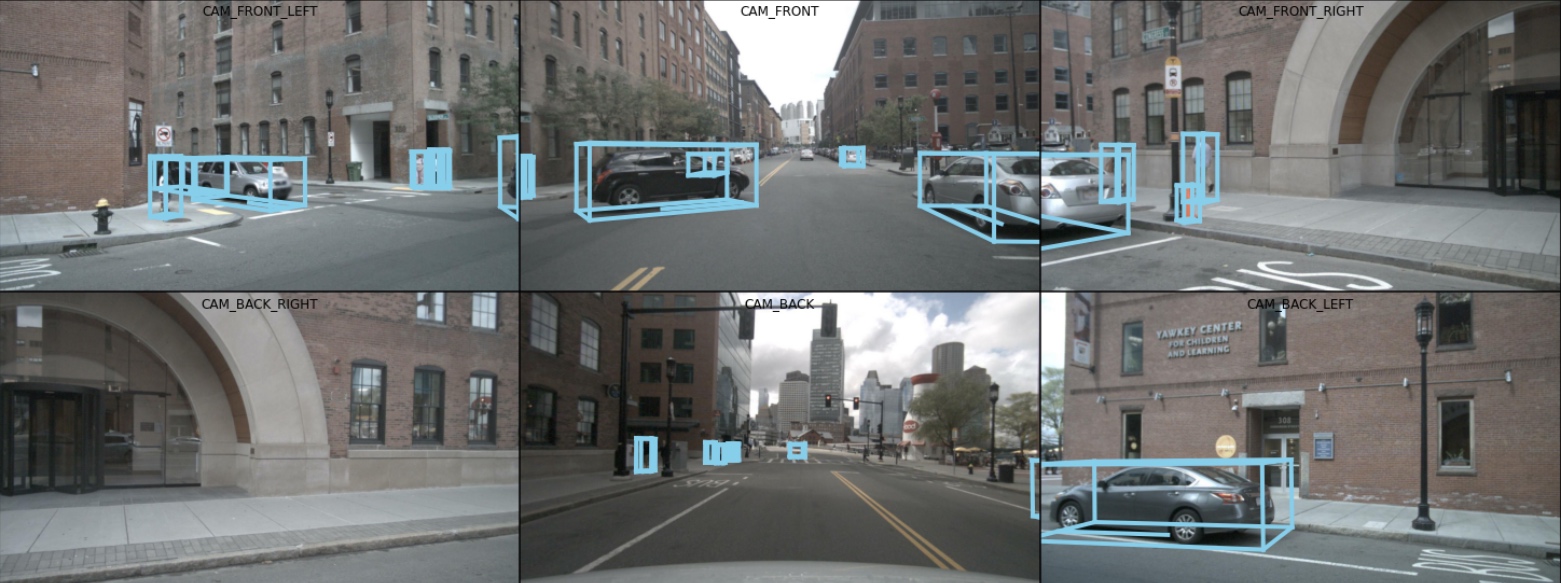}
\end{tabular}
}
\endgroup
\caption{\footnotesize \textbf{Downstream 3D Detection.} \textbf{(a)} Results of a 3D object detector trained on Real Data.  \textbf{(b)} Results of the same detector trained on Real Data + Aug Lighting. We obtain improved detections after training on data generated with our method. }
\label{fig:qual_perception3d} 
\end{figure}

\begin{figure*}[htbp]
\centering
\begingroup
\setlength{\tabcolsep}{0.5pt}
\begin{tabular}{cc}
\includegraphics[width=0.335\linewidth]{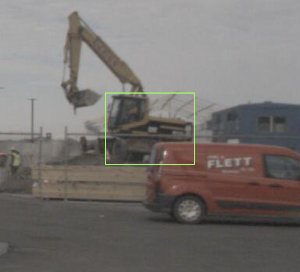} & 
\includegraphics[width=0.325\linewidth]{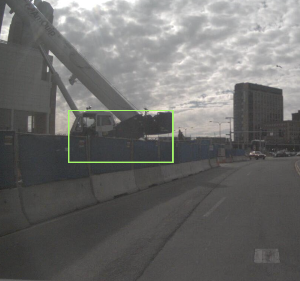} \\
\end{tabular}
\endgroup
\caption{\textbf{Example of nuScenes annotations of construction vehicles}. Extremities of construction vehicles are only included in nuScenes annotations if they interfere with traffic. Therefore, 3D bounding boxes for that object category are not uniquely defined.
} 
\label{fig:nuscenes_cv} 
\end{figure*}

\section{Discussion} 
\label{sec:discussion}

\subsubsection{Failure cases and future works.}
\begin{figure*}[t!]
\centering
\begingroup
\setlength{\tabcolsep}{0.5pt}
\resizebox{0.99\textwidth}{!}{
\begin{tabular}{ccc}
(a) Sensor effects & (b) Non-Lambertian shadows & (c) Occlusion \\
\includegraphics[width=0.335\linewidth]{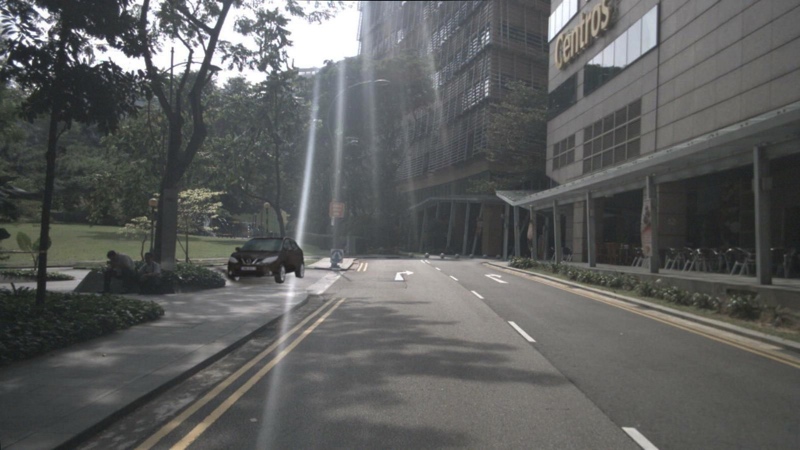} & 
\includegraphics[width=0.335\linewidth]{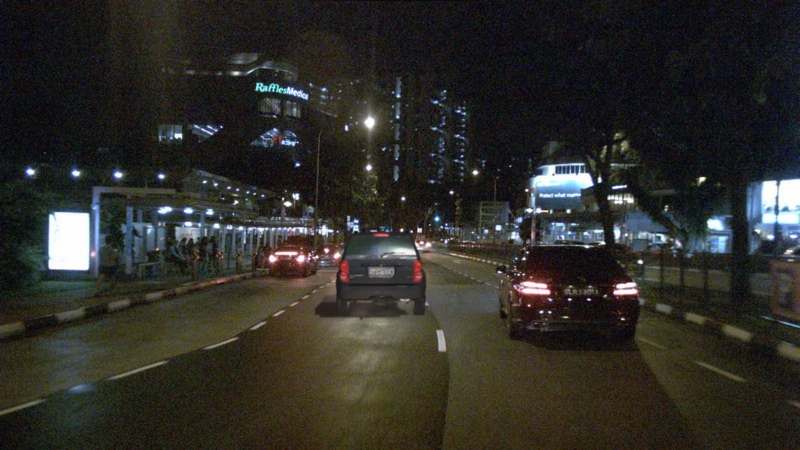} & 
\includegraphics[width=0.335\linewidth]{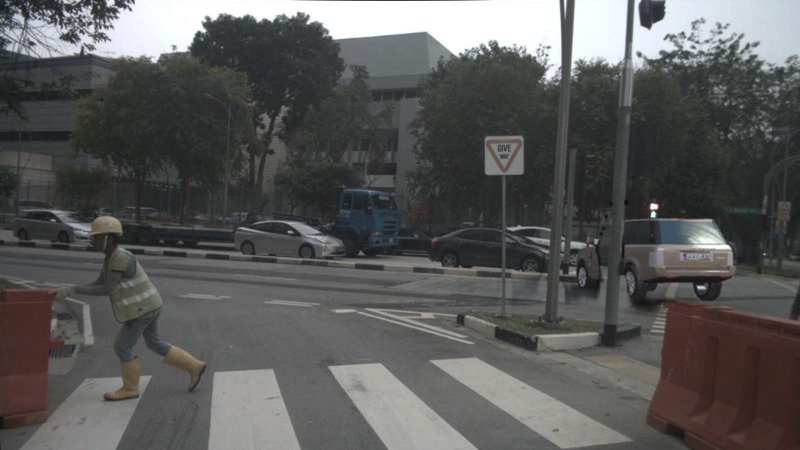} \\
\includegraphics[width=0.335\linewidth]{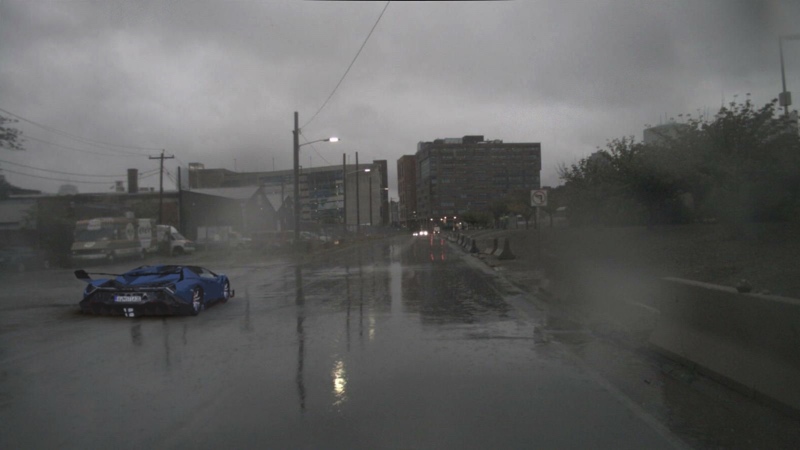} & 
\includegraphics[width=0.335\linewidth]{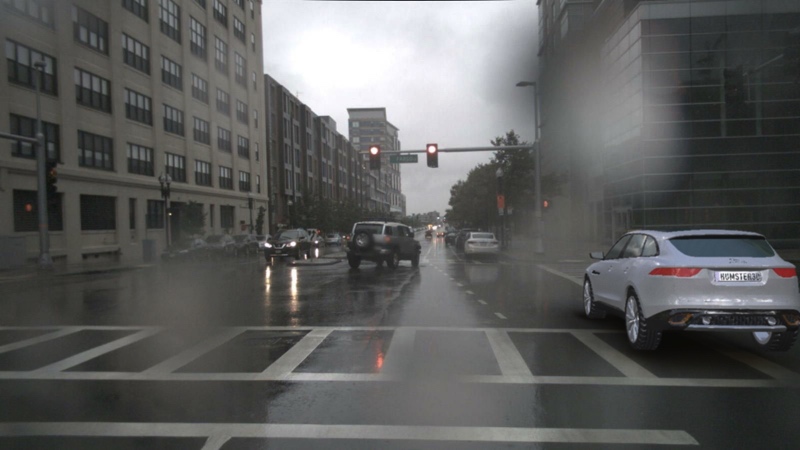} & 
\includegraphics[width=0.335\linewidth]{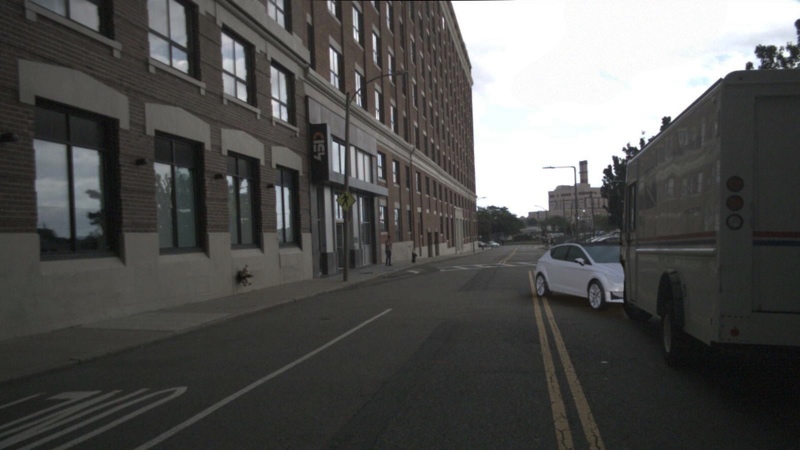} \\
\end{tabular}
}
\endgroup
\caption{\textbf{Qualitative results -- failure cases.} 
(a) The virtually inserted object may not pass exactly the same environment and signal processing pipeline as the real-world captured scene pixels, \eg the virtual object cannot reconstruct the halo effects of the sun and blurry regions caused by rain drops on the camera. 
(b) We assume a Lambertian scene surface when rendering the shadows, and thus the quality of editing results may decrease for wet specular road surfaces on rainy days. 
(c) Our occlusion handling relies on depth ordering, and may generate artifacts when the scene depth map prediction is inaccurate. 
} 
\label{fig:qual_failure} 
\end{figure*}

We show failure cases of our method in Figure~\ref{fig:qual_failure} and describe below. 

In our current object insertion pipeline, the inserted object pixels will not pass exactly the same image capturing process as the background scene pixels, and thus cannot simulate the sun halo effects and blurry rain-drop effects. 
In addition, a better modeling of the camera image signal processor (ISP) pipeline, such as tone-mapping, could potentially lead to more realistic object insertion. 
We believe it is an interesting future work to learn camera sensor parameters, and handle the diverse weather such as rainy, snowy and foggy effects. 

With unknown scene material properties, our current shadow map rendering (main paper Eq.~\paperref{4}) assumes a Lambertian scene surface. This assumption enables measuring the residual effects caused by the inserted object with a ratio image. 
Although the Lambertian assumption is usually a good approximation and widely adopted in prior works~\cite{neuralSengupta19,wang2021learning}, the shadow map quality will decrease when this assumption no longer holds. For example, the wet road surfaces in rainy days become quite specular and should reflect the appearance of the inserted object, as shown in the second column of Figure~\ref{fig:qual_failure}. It is an interesting direction to jointly estimate scene material properties to address such complex effects. 
In the third column of Figure~\ref{fig:qual_failure}, our editing results exhibit occlusion artifacts when the scene depth map is inaccurate. While predicting more accurate depth is out of the scope of our paper, we believe an additional geometry refinement adopted in prior works~\cite{chen2021geosim} could be beneficial to address this issue.

\subsubsection{Broader impact.} 
Our paper focuses on a neural Augmented Reality (AR) pipeline for outdoor scenes, which first estimates the lighting information and insert virtual objects into the input image. 
We show that our carefully designed hybrid lighting representation handles both the spatially-varying effects and the extreme HDR intensity of outdoor scenes. 
The differentiable object insertion formulation with an adversarial discriminator serves as a valuable supervision signal, which for the first time enables lighting supervision by jointly leveraging ground truth material, geometry and real world images. 
We show the benefits of our AR approach on a downstream 3D object detector, indicating its potential as a valuable data augmentation technique for safety-critical applications such as autonomous driving. 

Our method falls into the category of works that enable image editing. While we believe that there are many positive implications, however -- just like with the deep fake technology, we can also foresee nefarious use cases, such as rendering offensive content into photographs. Technology targeting detecting offensive content could help alleviate such use cases.

\begin{figure*}[t!]
\centering
\setlength{\tabcolsep}{1pt}
\renewcommand{\arraystretch}{0.5}
\footnotesize
\resizebox{0.99\textwidth}{!}{
\begin{tabular}{lcc}
& Example 1 & Example 2 \\
Input LDR \& GT HDR
  & \multicolumn{1}{m{0.38\linewidth}}{\includegraphics[width=\linewidth]{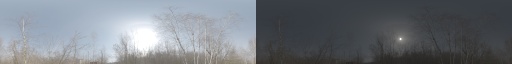}} 
  & \multicolumn{1}{m{0.38\linewidth}}{\includegraphics[width=\linewidth]{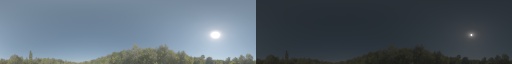}} \\
Reconstructed HDR sky 
  & \multicolumn{1}{m{0.38\linewidth}}{\includegraphics[width=\linewidth]{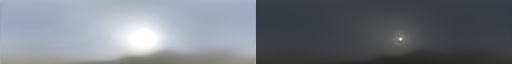}} 
  & \multicolumn{1}{m{0.38\linewidth}}{\includegraphics[width=\linewidth]{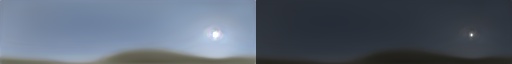}} \\
Peak direction editing 
  & \multicolumn{1}{m{0.38\linewidth}}{\includegraphics[width=\linewidth]{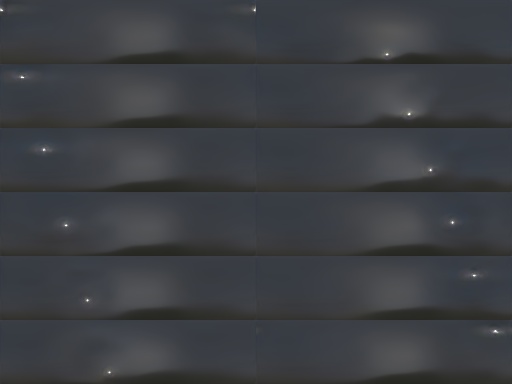}} 
  & \multicolumn{1}{m{0.38\linewidth}}{\includegraphics[width=\linewidth]{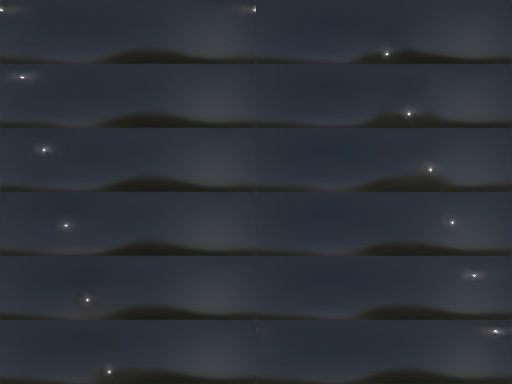}} \\
Input LDR \& GT HDR
  & \multicolumn{1}{m{0.38\linewidth}}{\includegraphics[width=\linewidth]{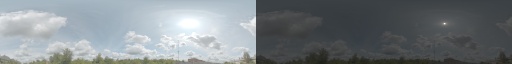}} 
  & \multicolumn{1}{m{0.38\linewidth}}{\includegraphics[width=\linewidth]{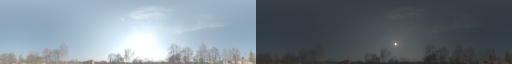}} \\
Reconstructed HDR sky  
  & \multicolumn{1}{m{0.38\linewidth}}{\includegraphics[width=\linewidth]{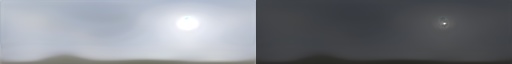}} 
  & \multicolumn{1}{m{0.38\linewidth}}{\includegraphics[width=\linewidth]{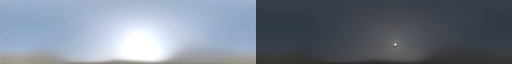}} \\
Peak direction editing 
  & \multicolumn{1}{m{0.38\linewidth}}{\includegraphics[width=\linewidth]{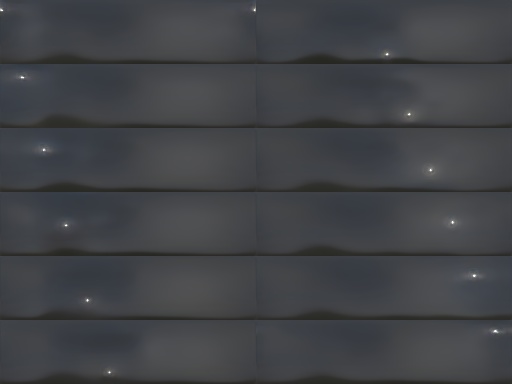}} 
  & \multicolumn{1}{m{0.38\linewidth}}{\includegraphics[width=\linewidth]{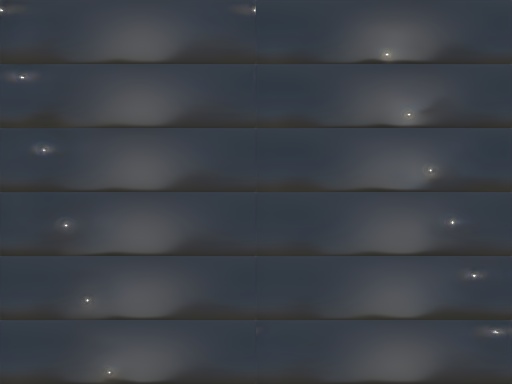}} \\
Input LDR \& GT HDR
  & \multicolumn{1}{m{0.38\linewidth}}{\includegraphics[width=\linewidth]{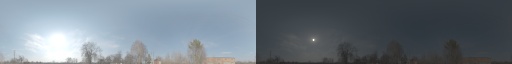}} 
  & \multicolumn{1}{m{0.38\linewidth}}{\includegraphics[width=\linewidth]{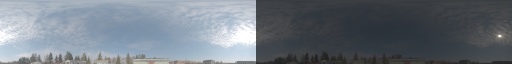}} \\
Reconstructed HDR sky  
  & \multicolumn{1}{m{0.38\linewidth}}{\includegraphics[width=\linewidth]{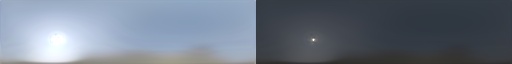}} 
  & \multicolumn{1}{m{0.38\linewidth}}{\includegraphics[width=\linewidth]{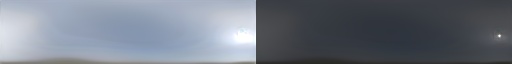}} \\
Peak direction editing 
  & \multicolumn{1}{m{0.38\linewidth}}{\includegraphics[width=\linewidth]{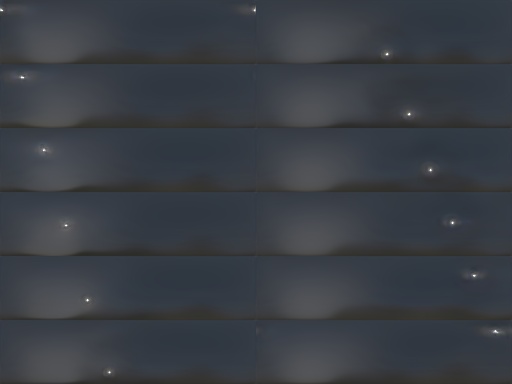}} 
  & \multicolumn{1}{m{0.38\linewidth}}{\includegraphics[width=\linewidth]{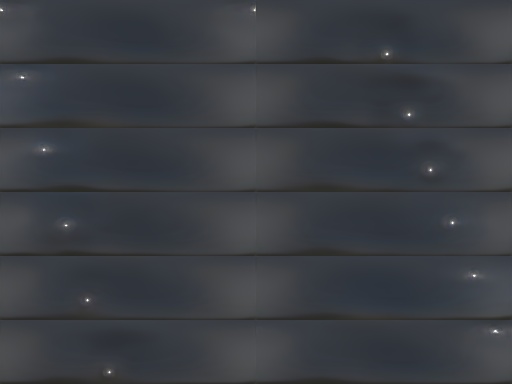}} \\
\end{tabular}
}
\caption{\textbf{Qualitative results of the sky modeling network.} We show two examples each row, and visualize with two exposure values to show LDR and HDR. Given an LDR sky panorama as input, our sky modeling network can reconstruct HDR peaks with extreme intensity values. We also provide peak editing results by changing the peak direction of the sky feature vector, where we change $\mathbf{f}_\text{dir}$ but fix $\mathbf{f}_\text{intensity},\mathbf{f}_\text{latent}$. As a result, we successfully generate HDR sky map with the same content but different sun directions. This allows for potential post-editing.
} 
\label{fig:qual_skymodel} 
\end{figure*}

\begin{figure*}[t!]
\small
\centering
\begingroup
\setlength{\tabcolsep}{0.5pt}
\resizebox{0.99\textwidth}{!}{
\begin{tabular}{ccc}
\includegraphics[width=0.335\linewidth]{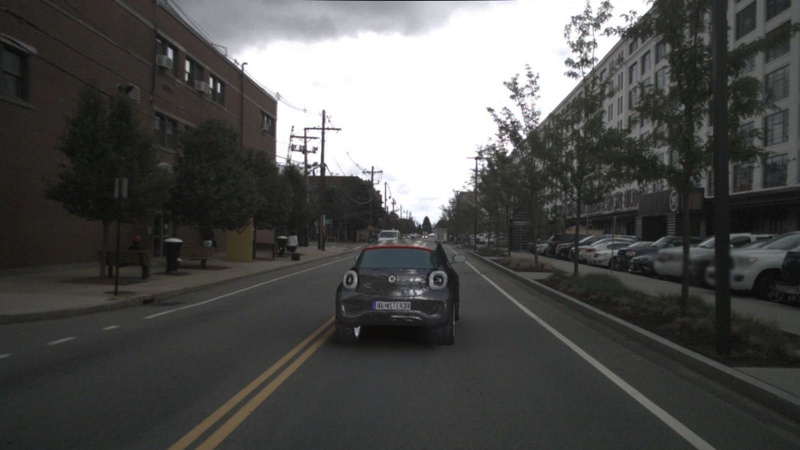} & 
\includegraphics[width=0.335\linewidth]{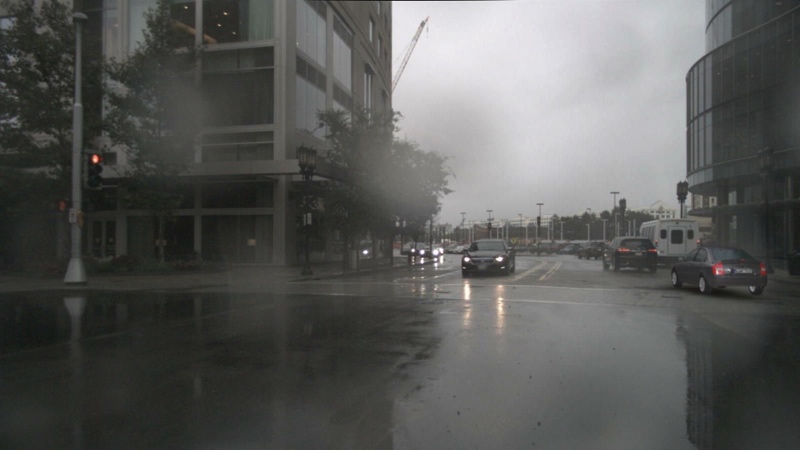} & 
\includegraphics[width=0.335\linewidth]{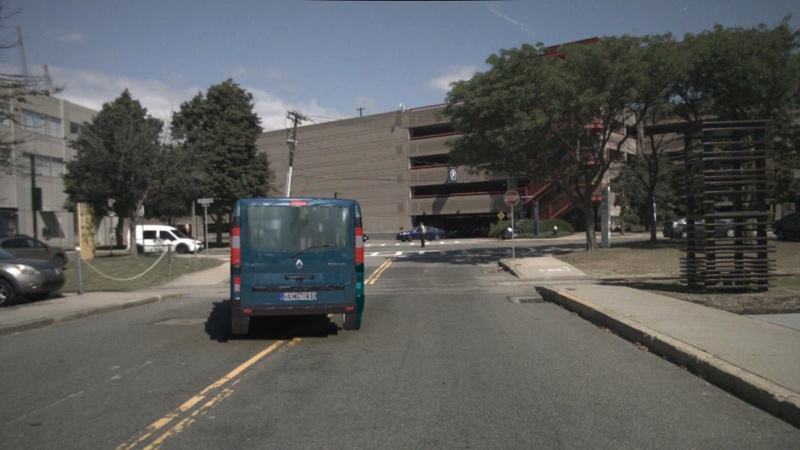} \\
\includegraphics[width=0.335\linewidth]{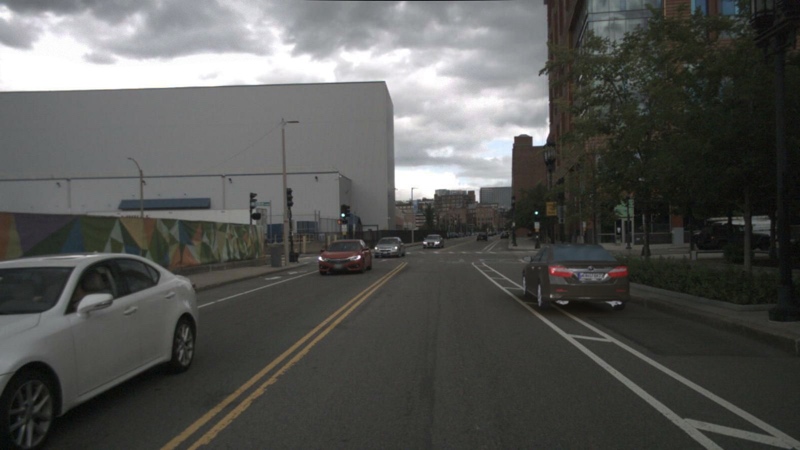} & 
\includegraphics[width=0.335\linewidth]{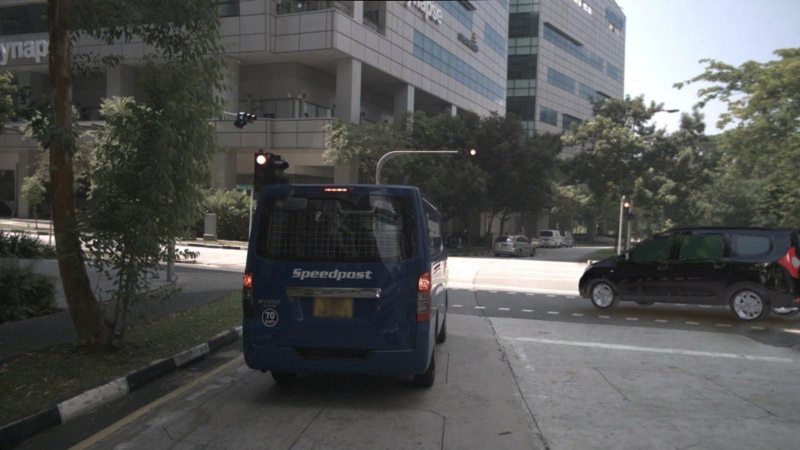} & 
\includegraphics[width=0.335\linewidth]{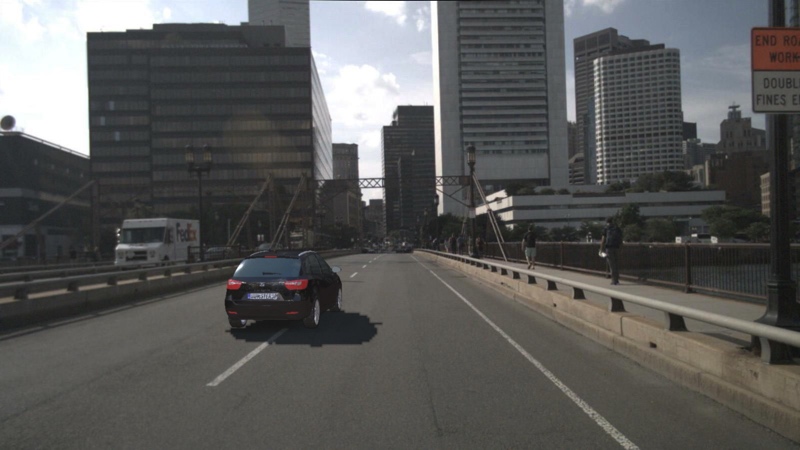} \\
\includegraphics[width=0.335\linewidth]{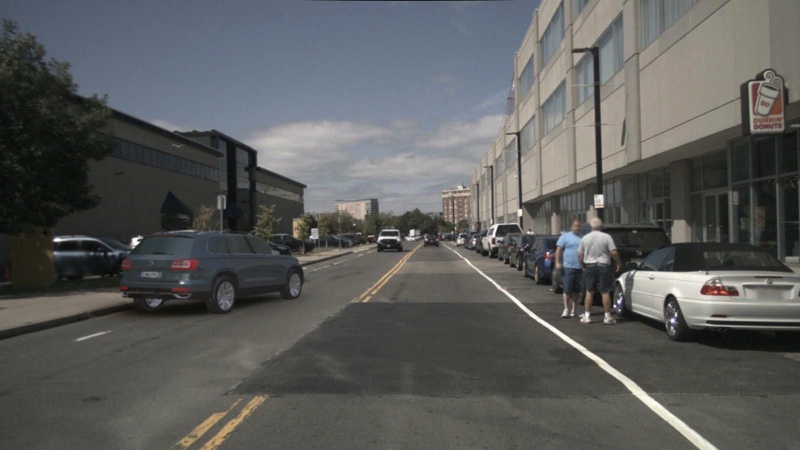} & 
\includegraphics[width=0.335\linewidth]{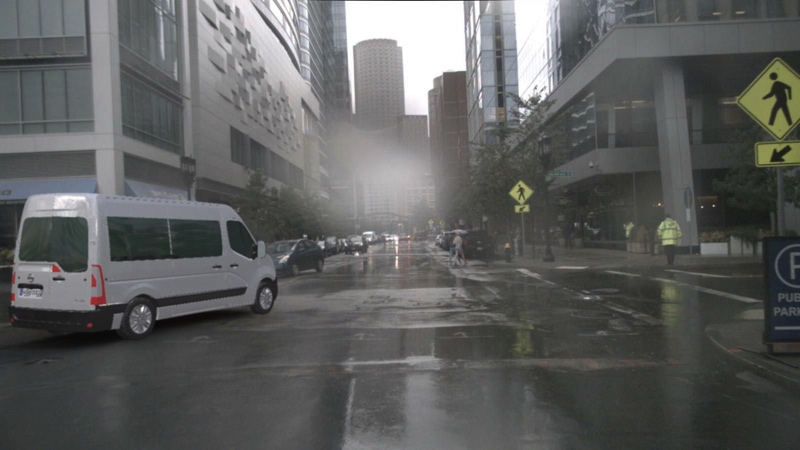} & 
\includegraphics[width=0.335\linewidth]{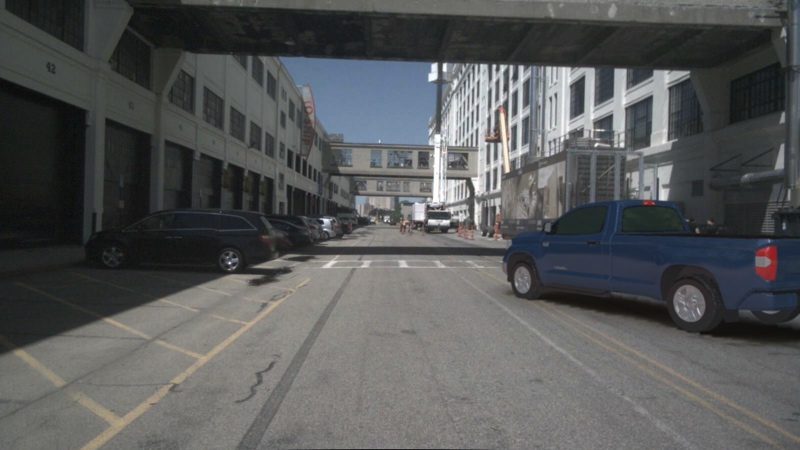} \\
\includegraphics[width=0.335\linewidth]{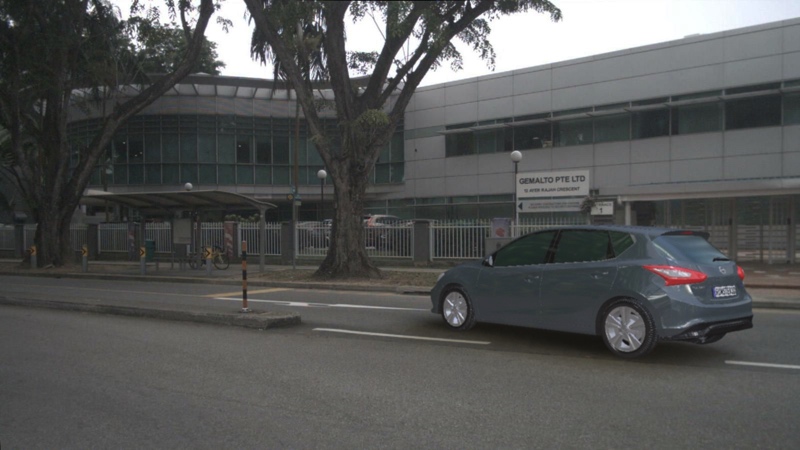} & 
\includegraphics[width=0.335\linewidth]{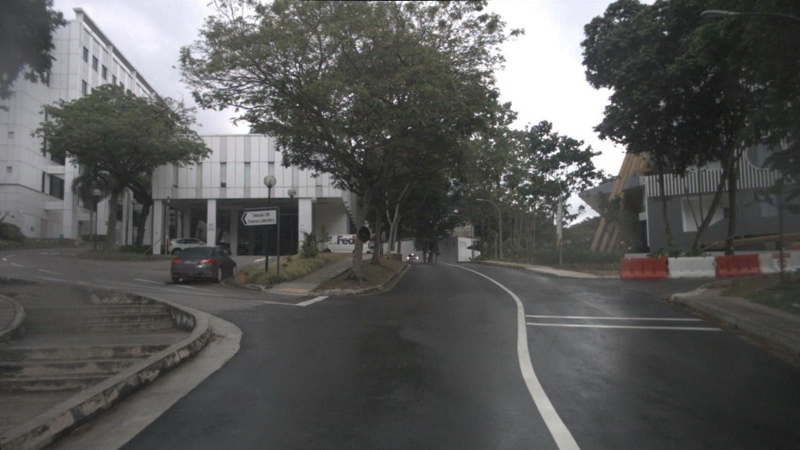} & 
\includegraphics[width=0.335\linewidth]{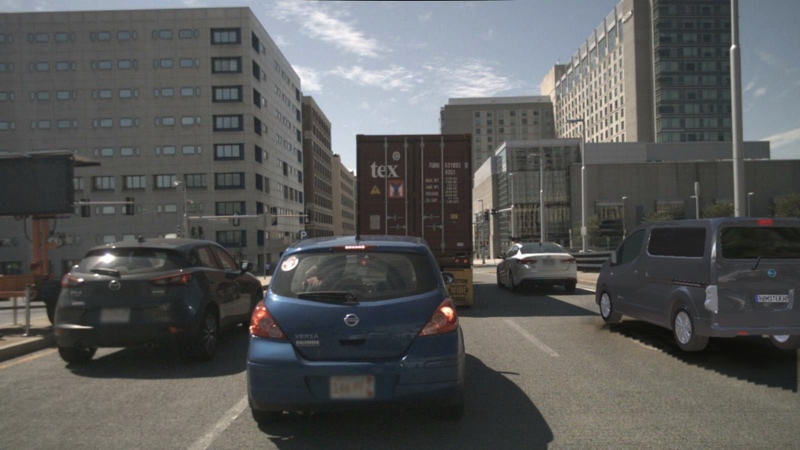} \\
\includegraphics[width=0.335\linewidth]{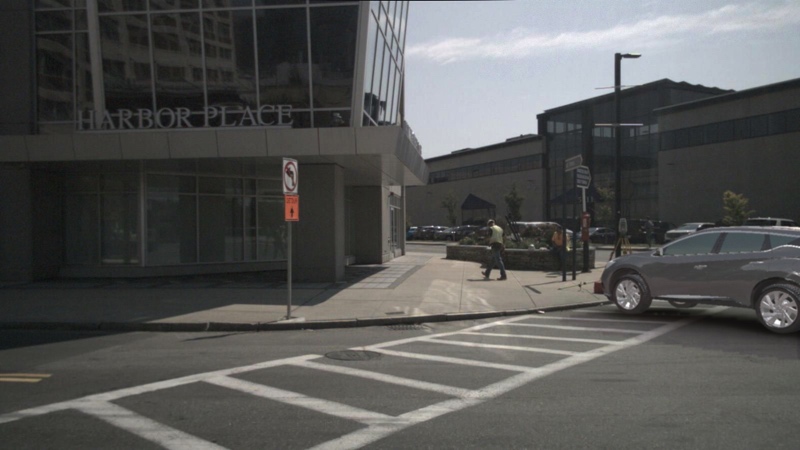} & 
\includegraphics[width=0.335\linewidth]{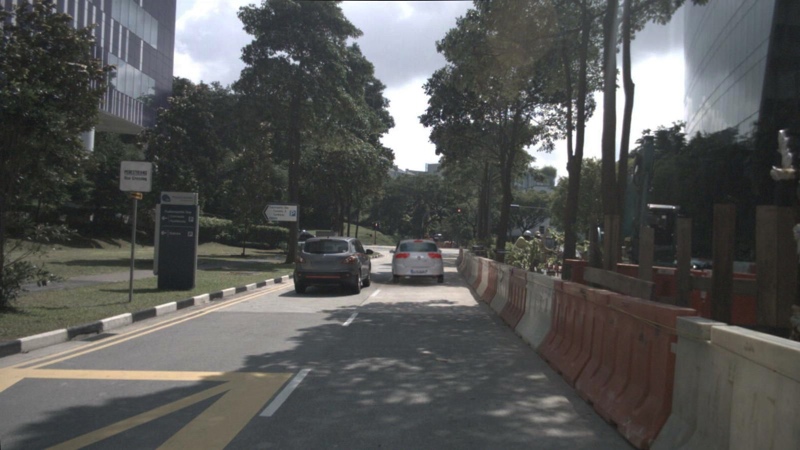} & 
\includegraphics[width=0.335\linewidth]{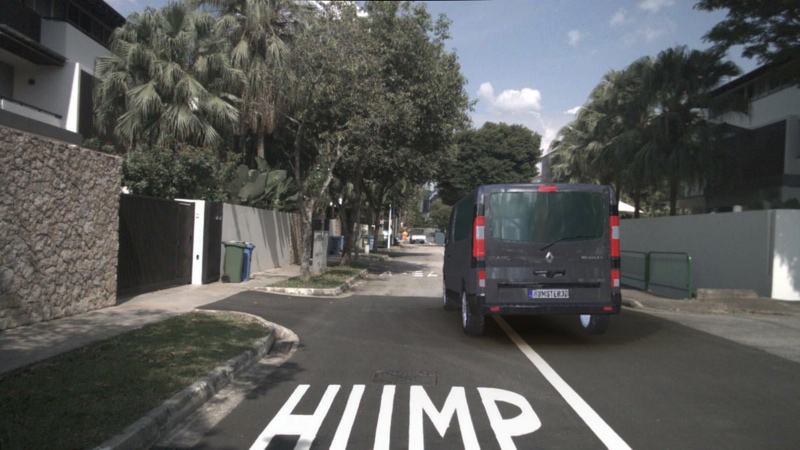} \\
\end{tabular}
}
\endgroup
\caption{\textbf{Qualitative results of automatic data augmentation on nuScenes dataset~\cite{nuscenes2019}.} Each image contains one inserted virtual car rendered with our neural AR approach. } 
\label{fig:qual_dataaug} 
\end{figure*}

\begin{figure*}[t!]
\small
\centering
\begingroup
\setlength{\tabcolsep}{0.5pt}
\resizebox{0.99\textwidth}{!}{
\begin{tabular}{ccc}
\includegraphics[width=0.335\linewidth]{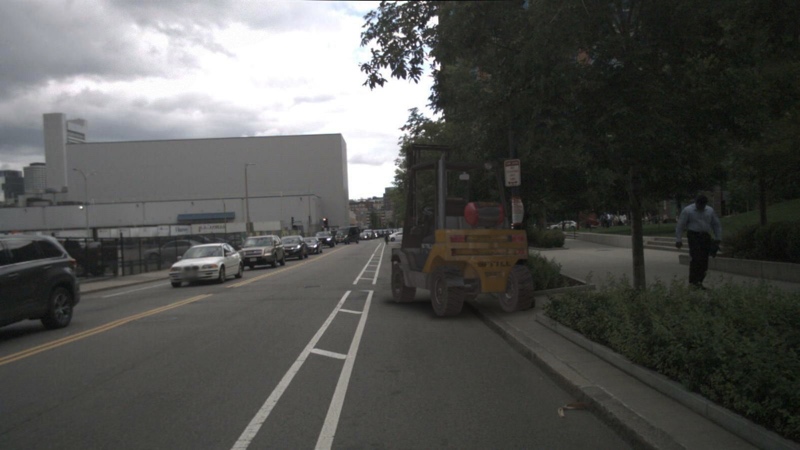} & 
\includegraphics[width=0.335\linewidth]{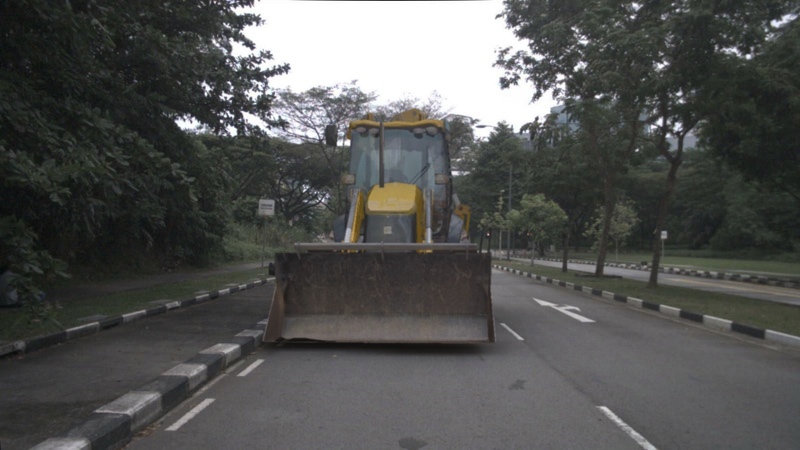} & 
\includegraphics[width=0.335\linewidth]{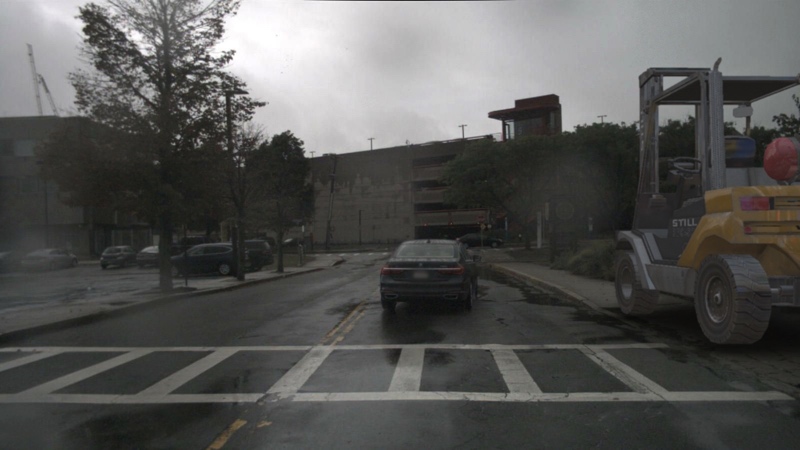} \\
\includegraphics[width=0.335\linewidth]{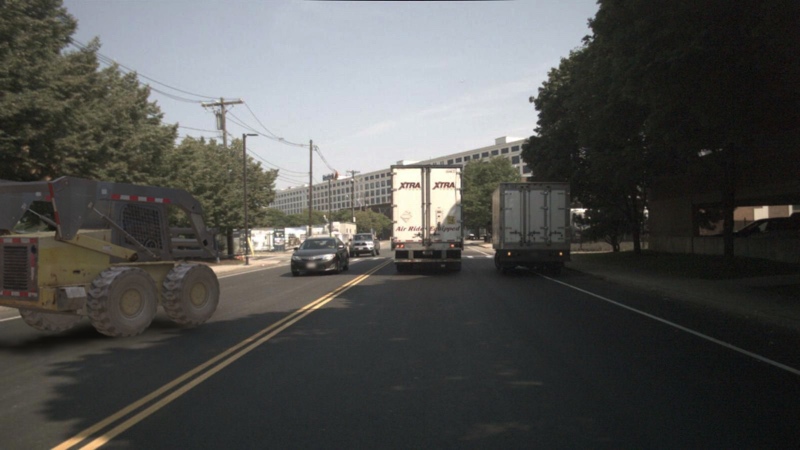} & 
\includegraphics[width=0.335\linewidth]{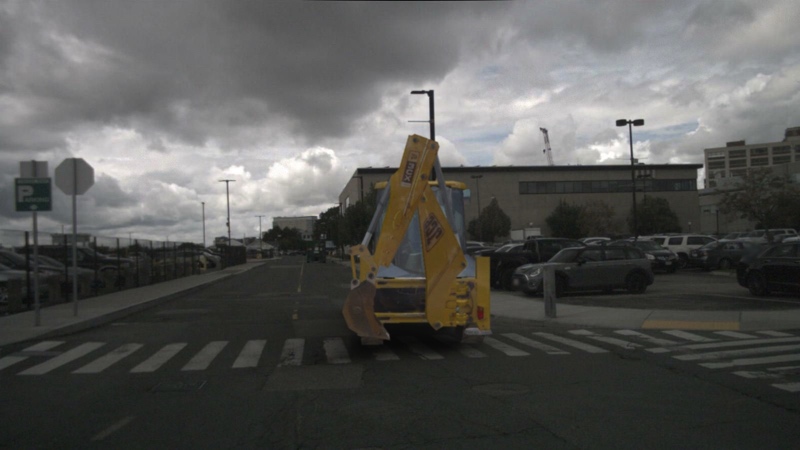} & 
\includegraphics[width=0.335\linewidth]{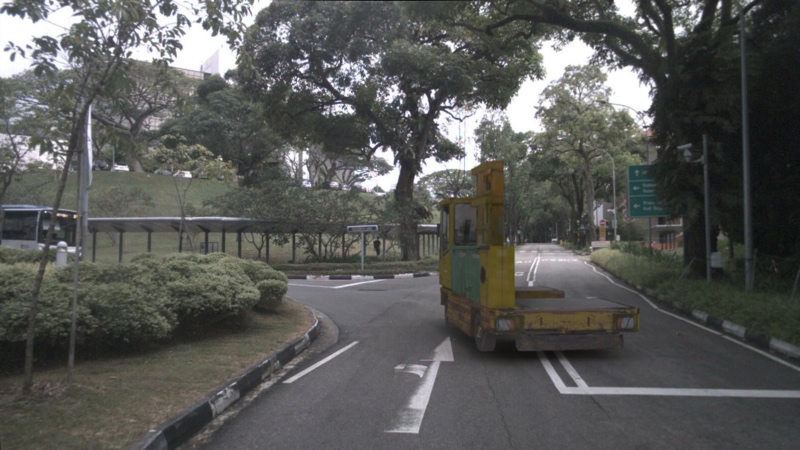} \\
\end{tabular}
}
\endgroup
\caption{\textbf{Construction vehicle insertion results of automatic data augmentation on nuScenes dataset~\cite{nuscenes2019}.} Our neural AR approach is agnostic to the category of 3D assets and can realistically insert virtual objects belonging to rare classes. } 
\label{fig:qual_dataaug_cv} 
\end{figure*}

\begin{figure*}[t!]
\small
\centering
\begingroup
\setlength{\tabcolsep}{0.5pt}
\resizebox{0.99\textwidth}{!}{
\begin{tabular}{ccc}
\includegraphics[width=0.335\linewidth]{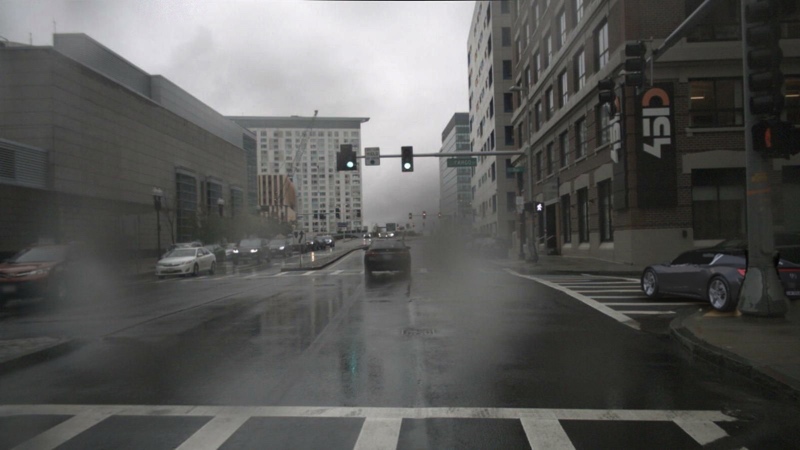} & 
\includegraphics[width=0.335\linewidth]{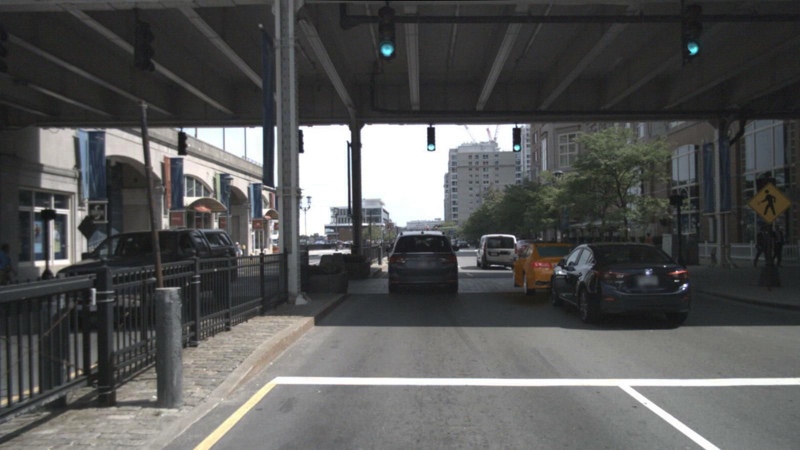} & 
\includegraphics[width=0.335\linewidth]{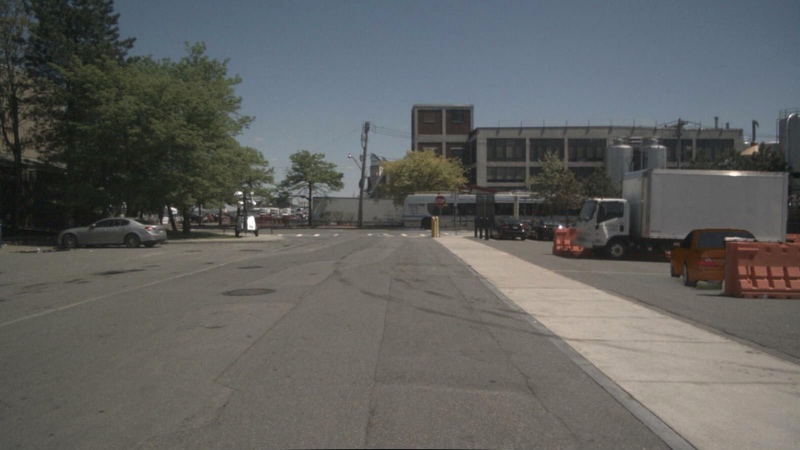} \\
\includegraphics[width=0.335\linewidth]{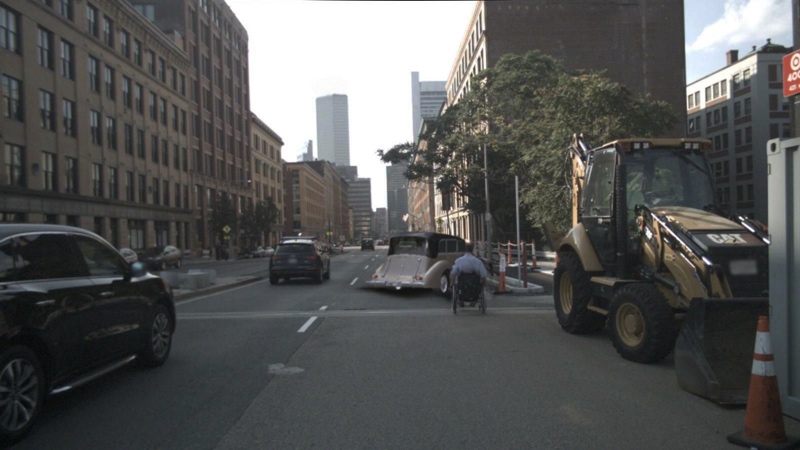} & 
\includegraphics[width=0.335\linewidth]{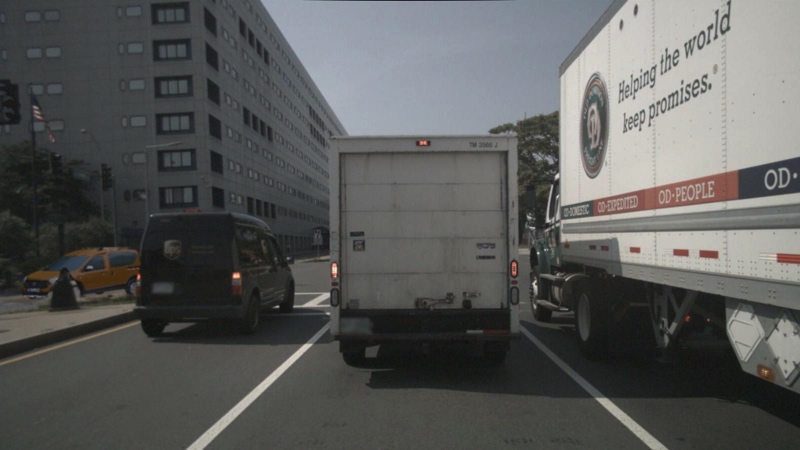} & 
\includegraphics[width=0.335\linewidth]{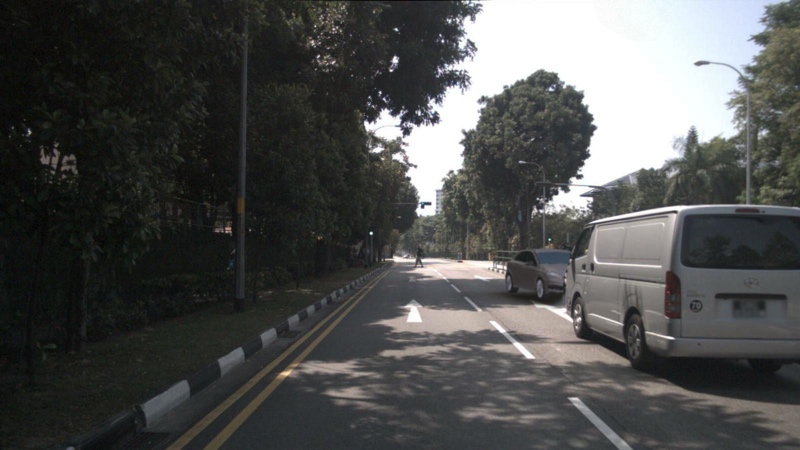} \\
\end{tabular}
}
\endgroup
\caption{\textbf{Occlusion results of automatic data augmentation on nuScenes~\cite{nuscenes2019}.} We naively handle occlusion by comparing the scene depth map and object Z buffer. Scene depth map is predicted with a pre-trained state-of-the-art monocular depth estimation model PackNet~\cite{packnet}. } 
\label{fig:qual_dataaug_occlusion} 
\end{figure*}

\begin{figure*}[t!]
\centering
\begingroup
\setlength{\tabcolsep}{0.5pt}
\resizebox{0.99\textwidth}{!}{
\begin{tabular}{ccc}
\includegraphics[width=0.335\linewidth]{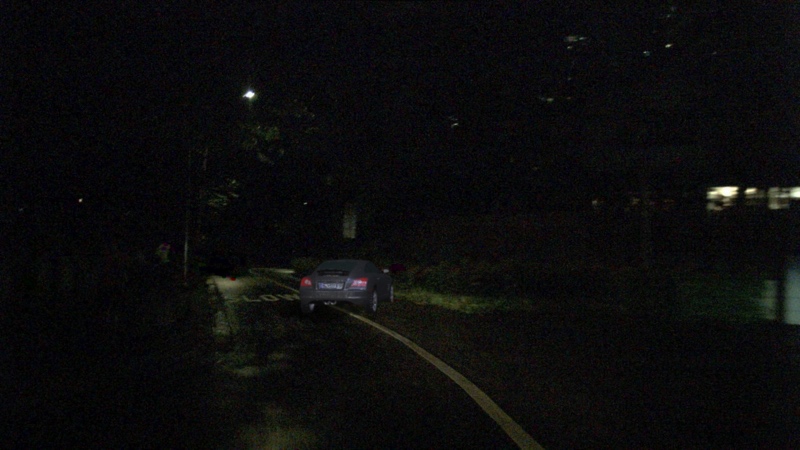} & 
\includegraphics[width=0.335\linewidth]{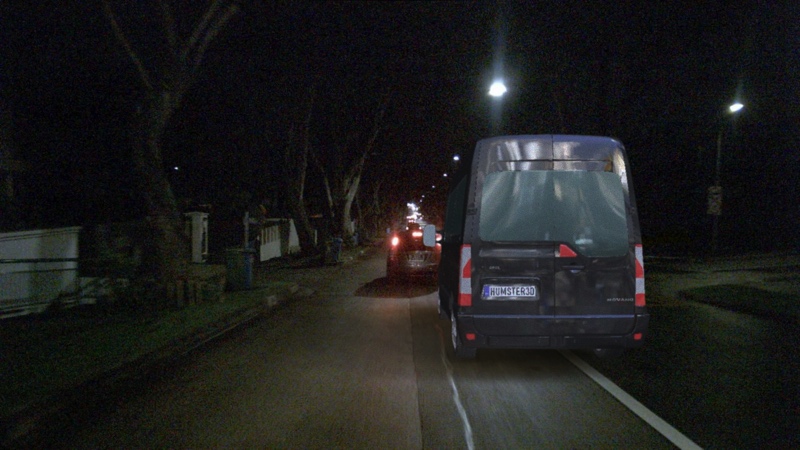} & 
\includegraphics[width=0.335\linewidth]{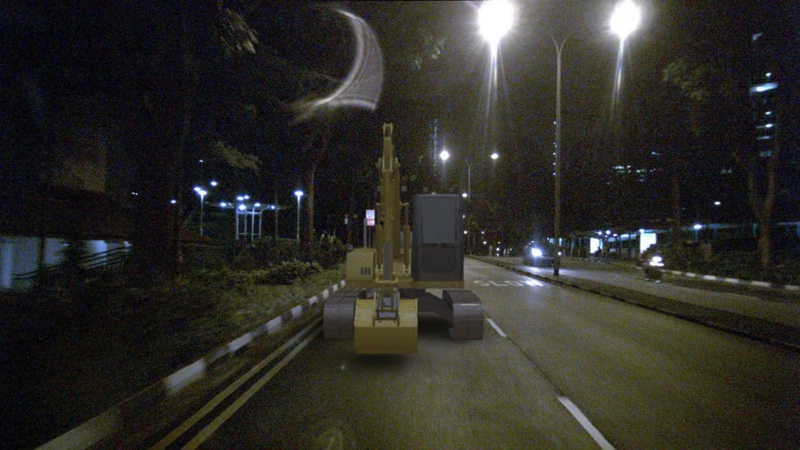} \\
\end{tabular}
}
\endgroup
\caption{\textbf{Night time results of automatic data augmentation on the nuScenes dataset~\cite{nuscenes2019}.} 
Our method can produce reasonable results without any HDR data supervision for the night time outdoor scenes. 
} 
\label{fig:qual_dataaug_night} 
\end{figure*}

\begin{figure*}[t!]
\small
\centering
\begingroup
\setlength{\tabcolsep}{0.5pt}
\resizebox{0.99\textwidth}{!}{
\begin{tabular}{ccc}
\includegraphics[width=0.335\linewidth]{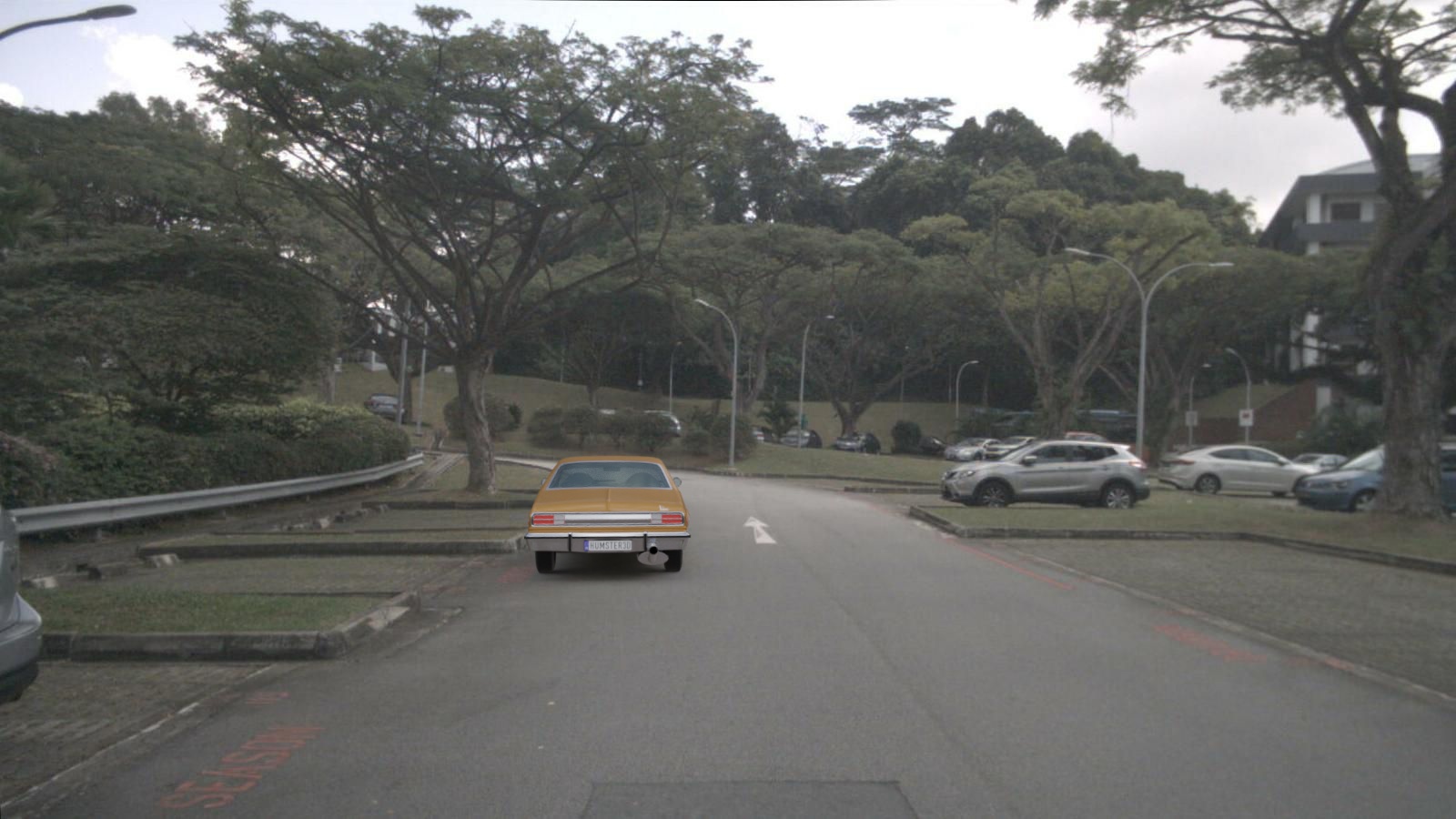} & 
\includegraphics[width=0.335\linewidth]{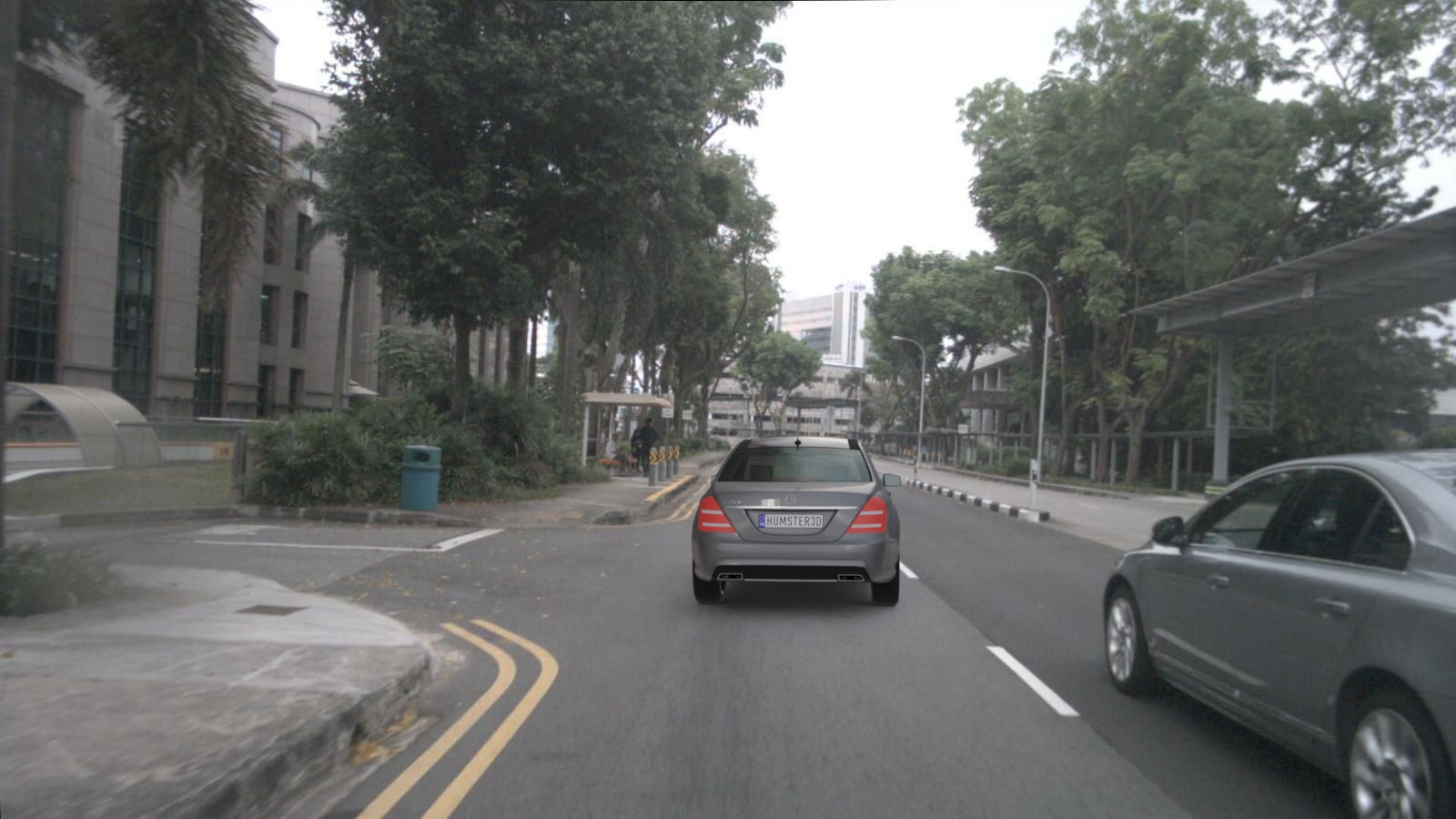} & 
\includegraphics[width=0.335\linewidth]{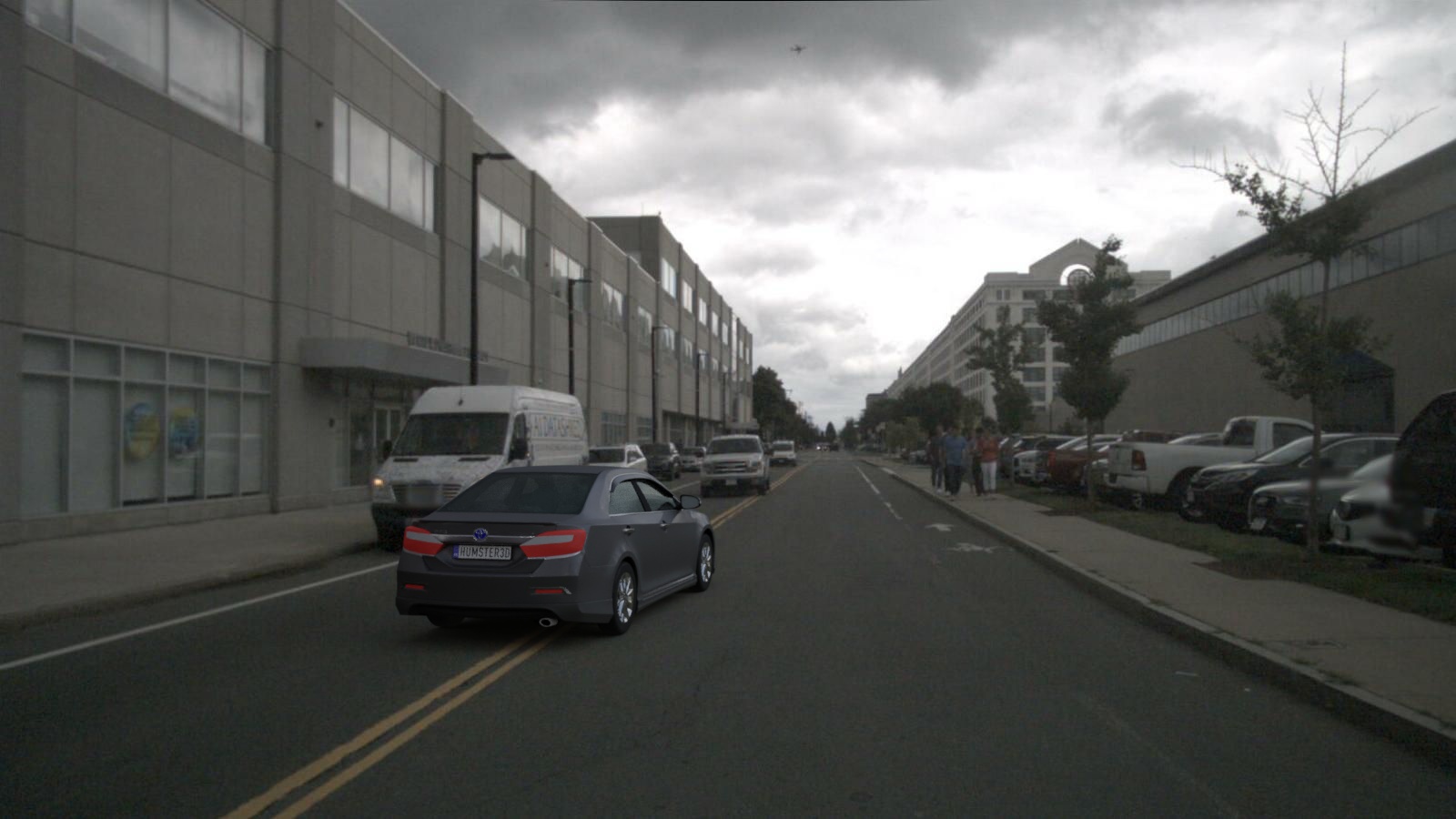} \\
\includegraphics[width=0.335\linewidth]{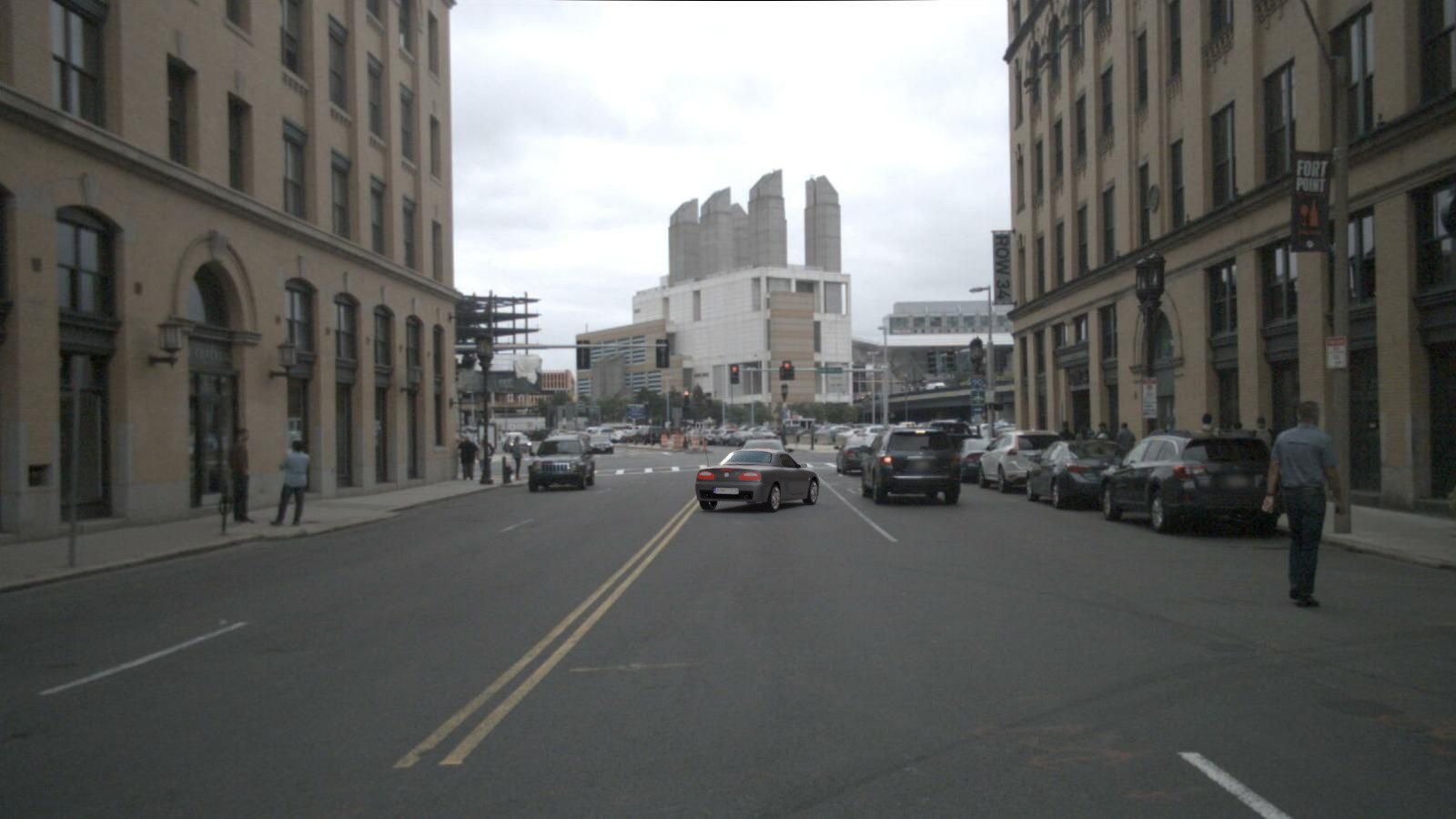} & 
\includegraphics[width=0.335\linewidth]{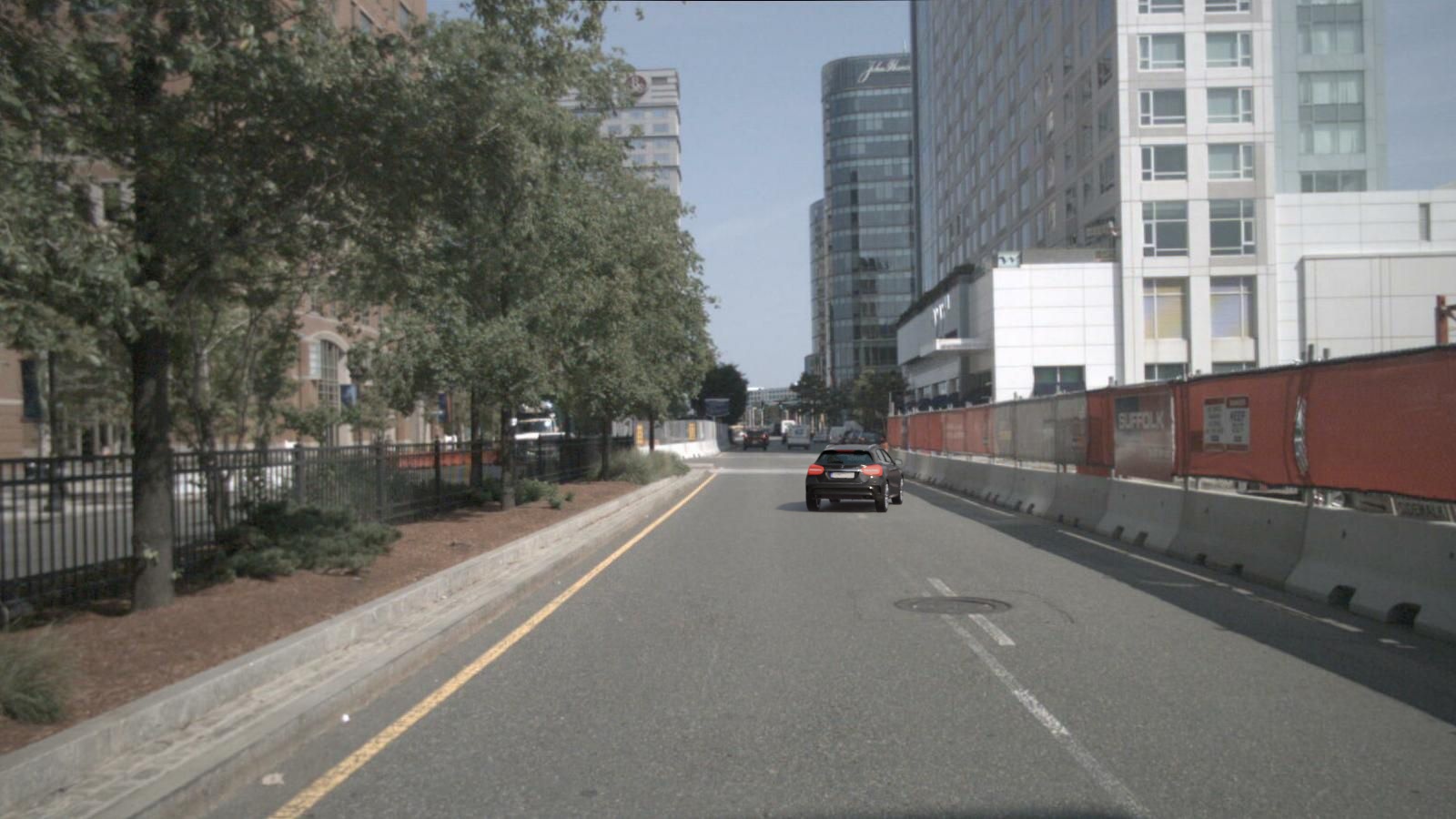} & 
\includegraphics[width=0.335\linewidth]{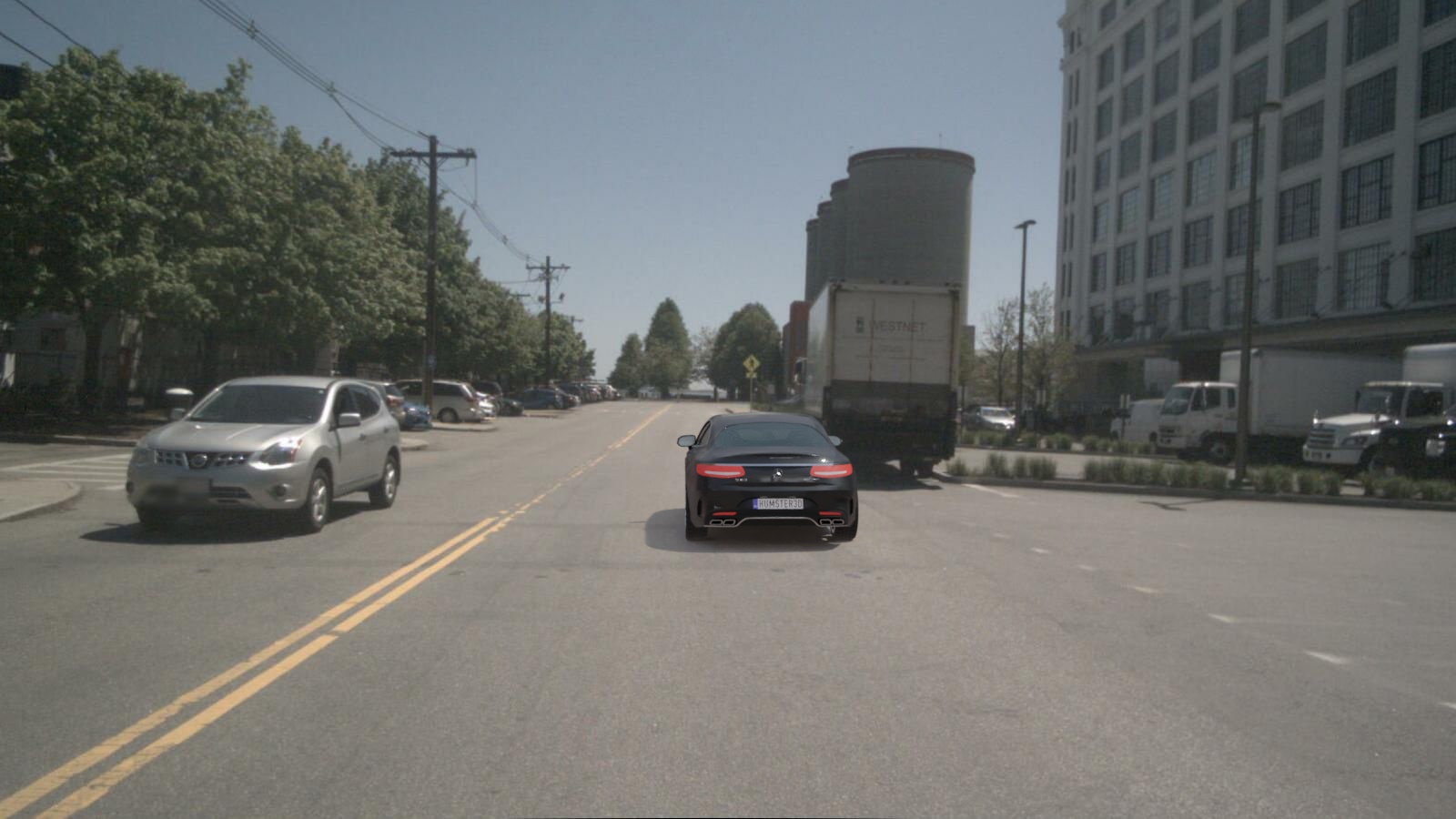} \\
\end{tabular}
}
\endgroup
\caption{\textbf{Qualitative results of object insertion rendered by Blender.} 
Our lighting representation can be rendered into location-specific environment map, which is compatible with commercial renderers. Integrated with powerful rendering engines, our method can leverage complex materials such as transparency and multi-bounce ray-tracing. 
} 
\label{fig:qual_blender} 
\end{figure*}

\begin{figure*}[t!]
\centering
\begingroup
\setlength{\tabcolsep}{0.5pt}
\resizebox{0.99\textwidth}{!}{
\begin{tabular}{ccc}
Initial editing results & 5 iterations & 10 iterations \\
\includegraphics[width=0.335\linewidth]{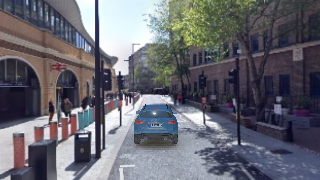} & 
\includegraphics[width=0.335\linewidth]{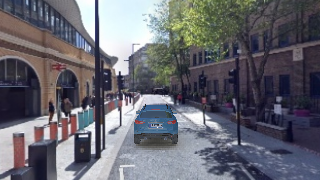} & 
\includegraphics[width=0.335\linewidth]{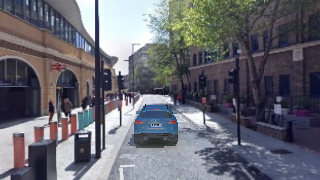} \\
\includegraphics[width=0.335\linewidth]{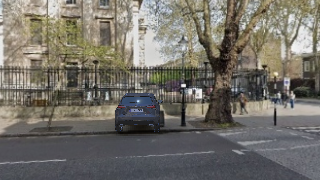} & 
\includegraphics[width=0.335\linewidth]{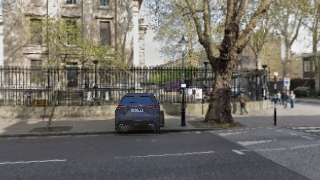} & 
\includegraphics[width=0.335\linewidth]{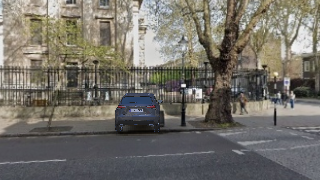} \\
\end{tabular}
}
\endgroup
\caption{\textbf{Qualitative visualization of the test-time optimization.} 
To understand the behaviour of the discriminator, we perform test-time optimization for the network parameters, where the only objective is to minimize the discriminator loss $\mathcal{L}_\text{adv}$. 
On the first column, we display the 320x180 resolution image editing results, which are consumed by the discriminator. 
The second and the third column shows the image editing results after 5 and 10 iterations. 
Note how the discriminator corrects the shadow direction in the first row, and removes obvious erroneous highlight in the second row. 
(Best viewed zooming in. We refer to additional animated result in the accompanied video.)
} 
\label{fig:qual_adv} 
\end{figure*}

\clearpage
%
%
\bibliographystyle{splncs04}
\bibliography{egbib}

%% file: intro.tex
\section{Introduction}
\label{sec:intro}

\begin{figure}[t]
\centering
\setlength{\tabcolsep}{0.1pt}
\resizebox{0.99\textwidth}{!}{
\begin{tabular}{cccc}
\includegraphics[width=0.28\linewidth,height=2.1cm,trim=110 10 0 0,clip]{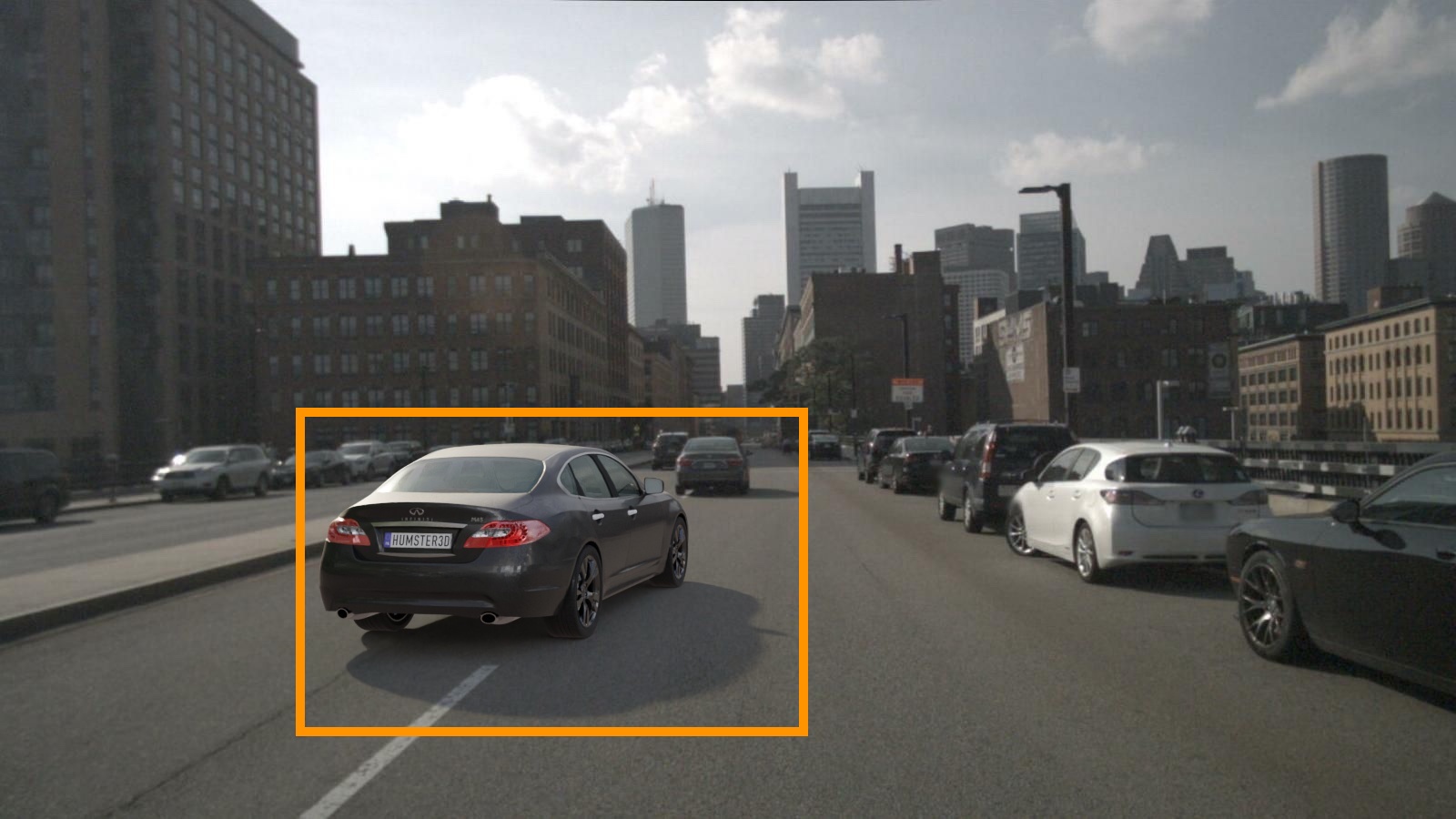} &
\includegraphics[width=0.28\linewidth,height=2.1cm,trim=40 10 30 0,clip]{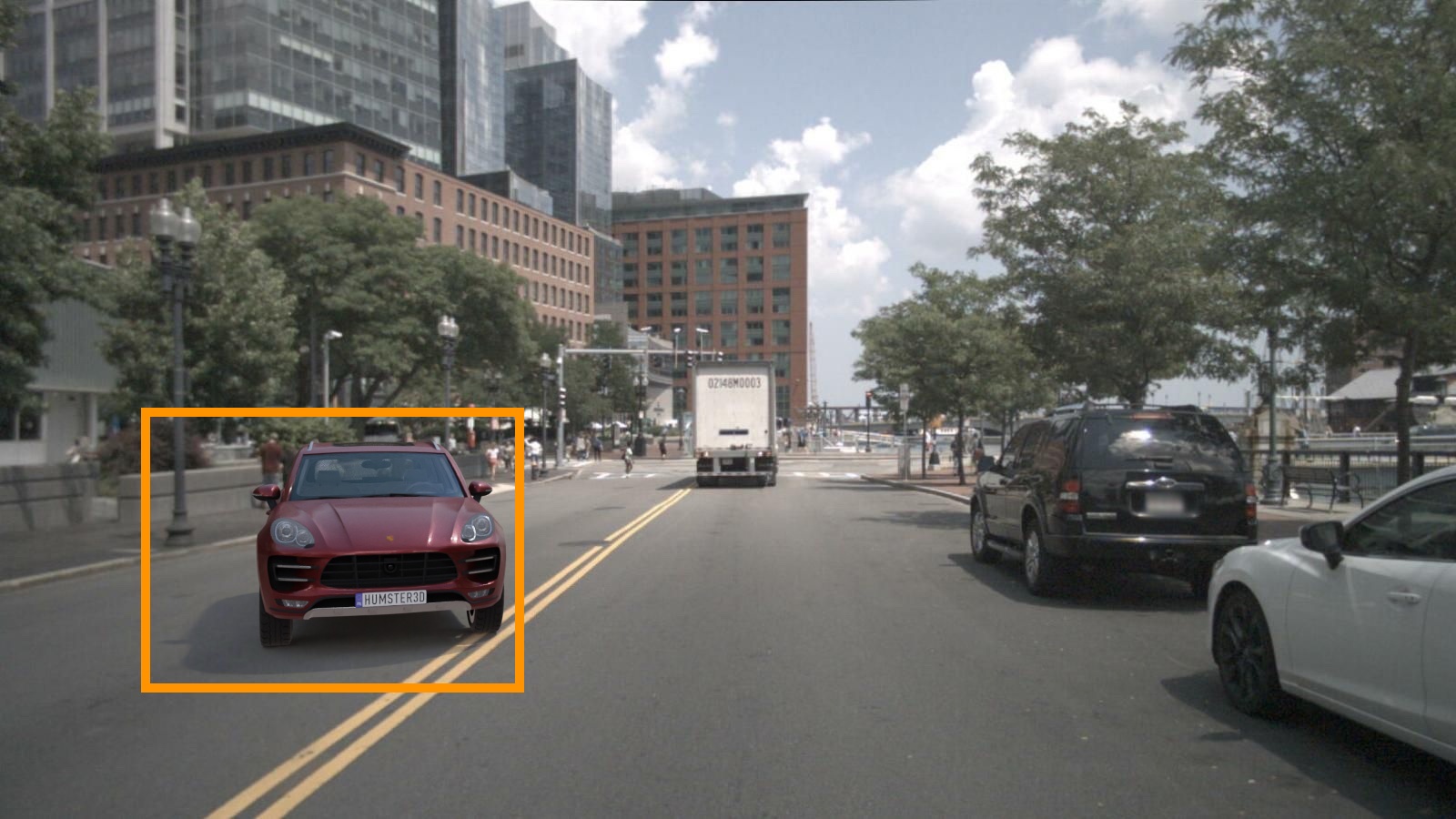} &
\includegraphics[width=0.28\linewidth,height=2.1cm,trim=40 10 40 0,clip]{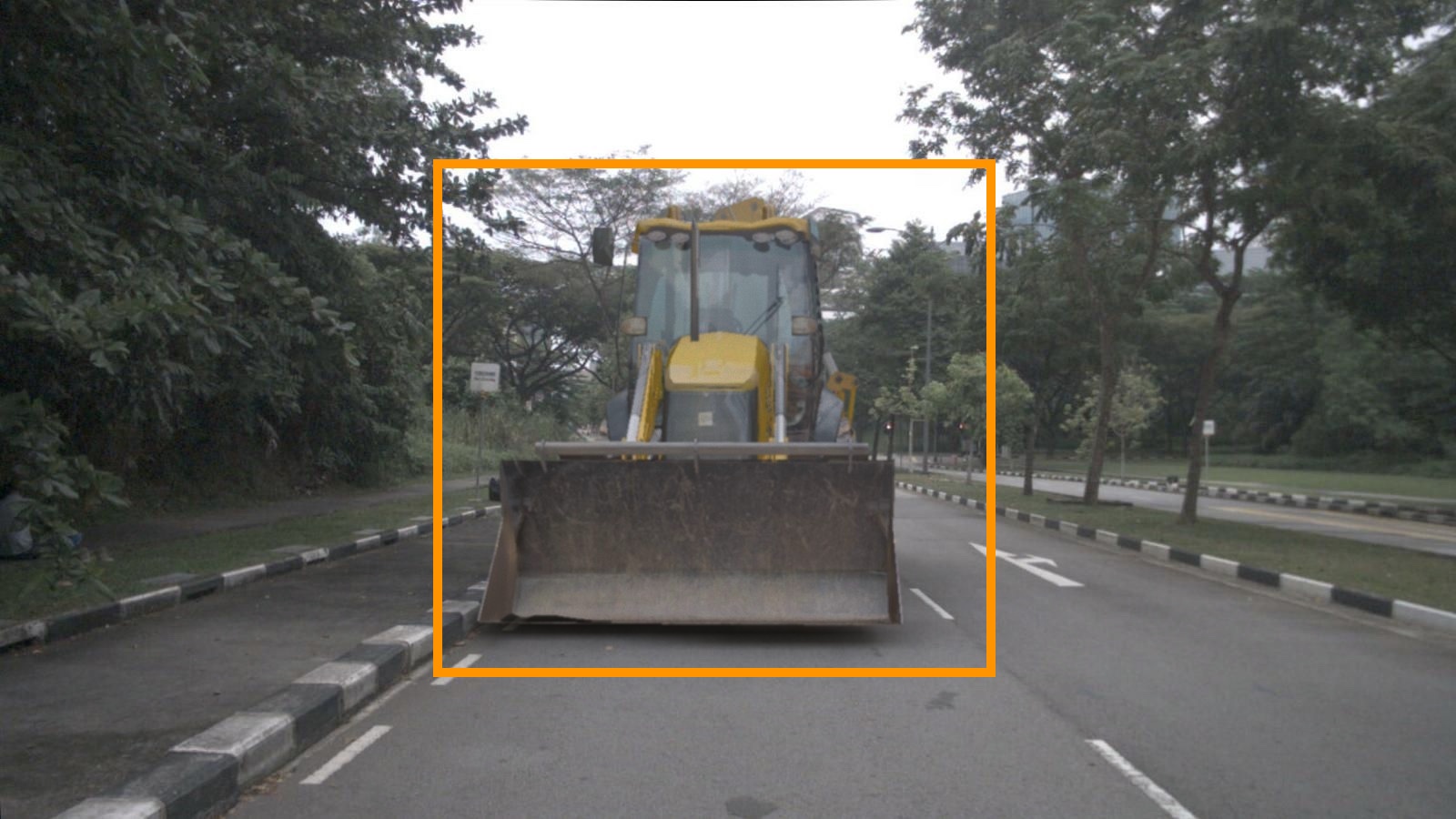} &
\includegraphics[width=0.28\linewidth,height=2.1cm,trim=40 10 30 0,clip]{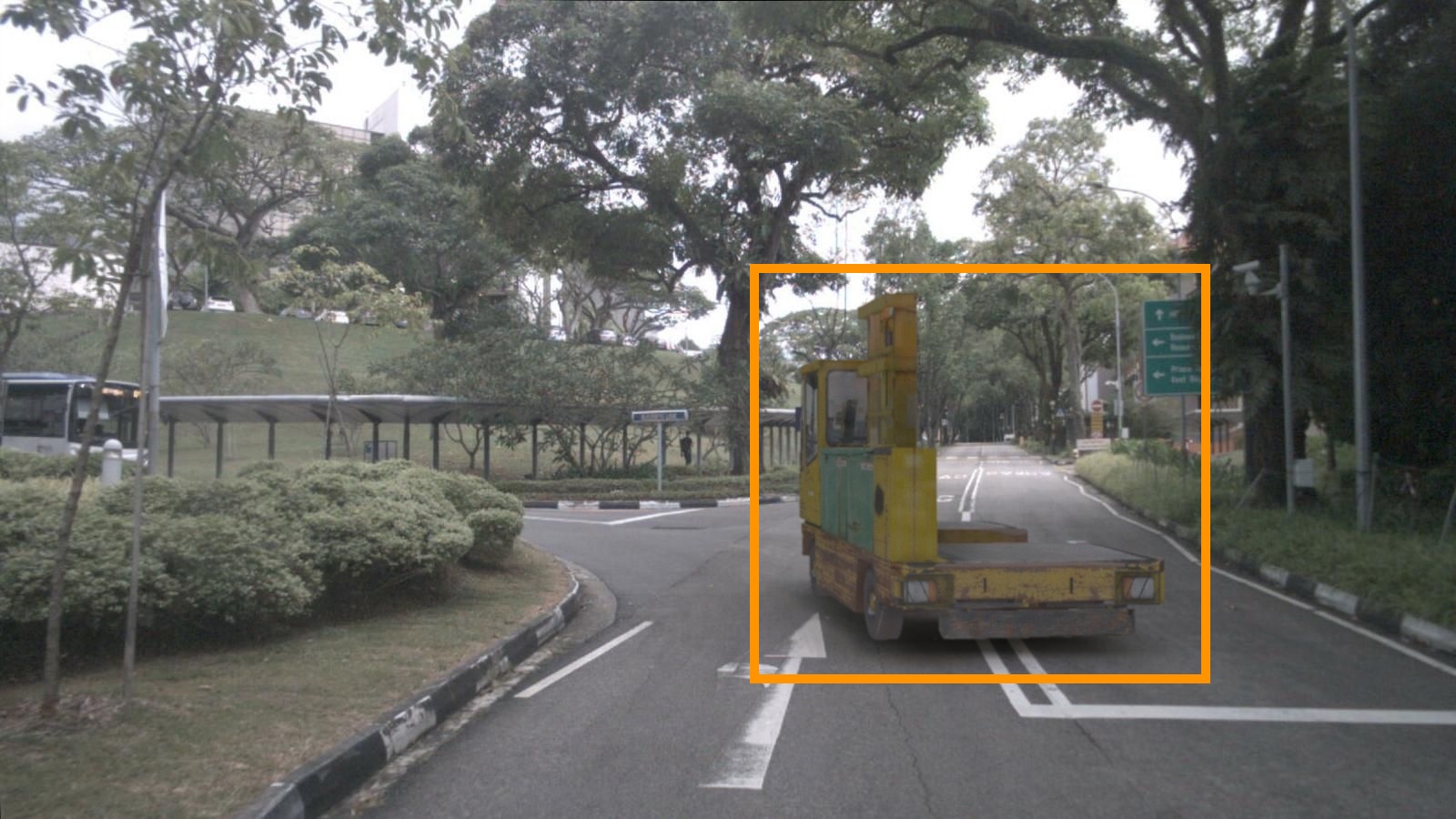} \\ 
\vspace{-4.5mm} \\ 
\includegraphics[width=0.28\linewidth,height=2.1cm,trim=110 10 0 0,clip]{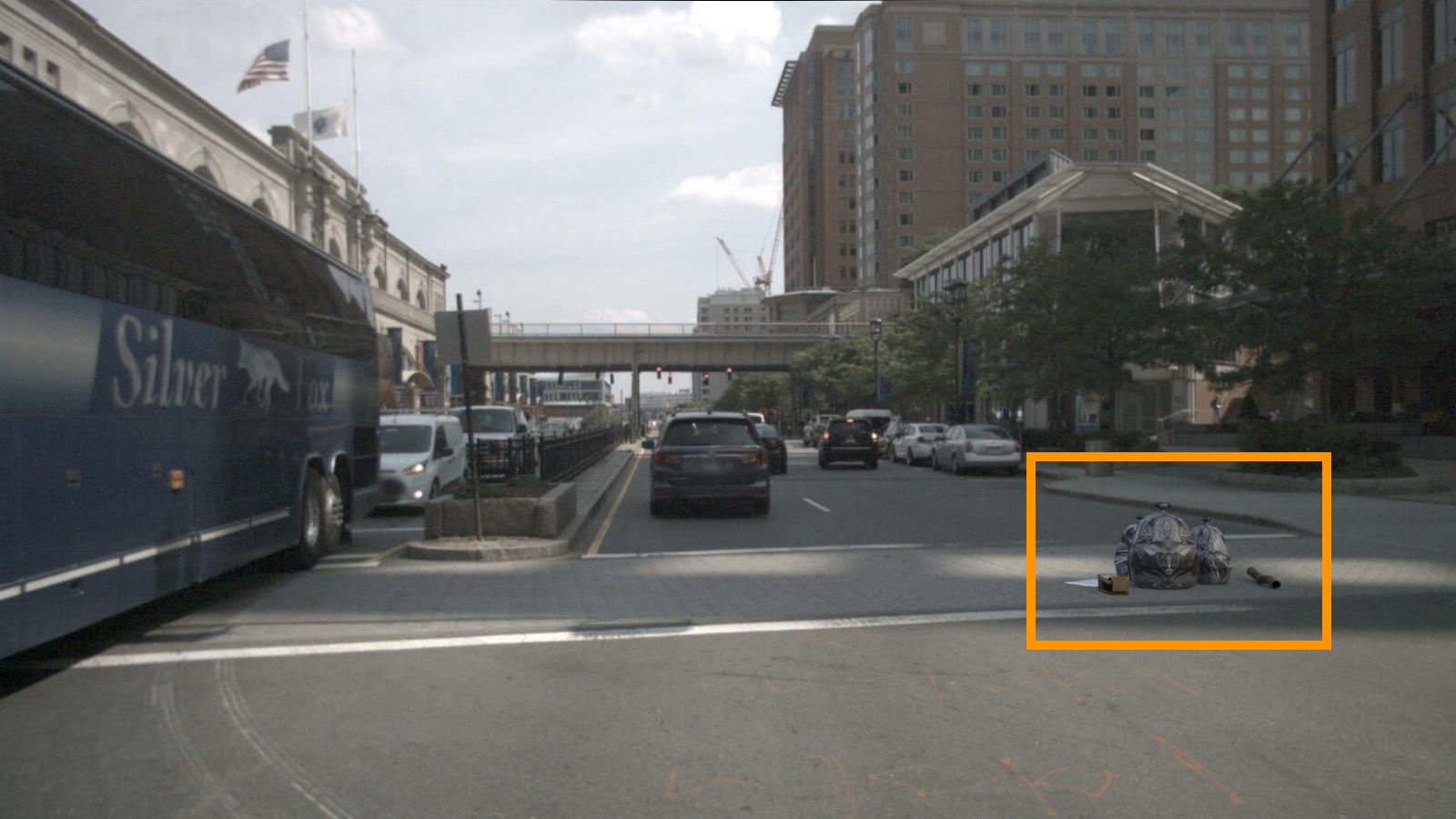} &
\includegraphics[width=0.28\linewidth,height=2.1cm,trim=40 10 30 0,clip]{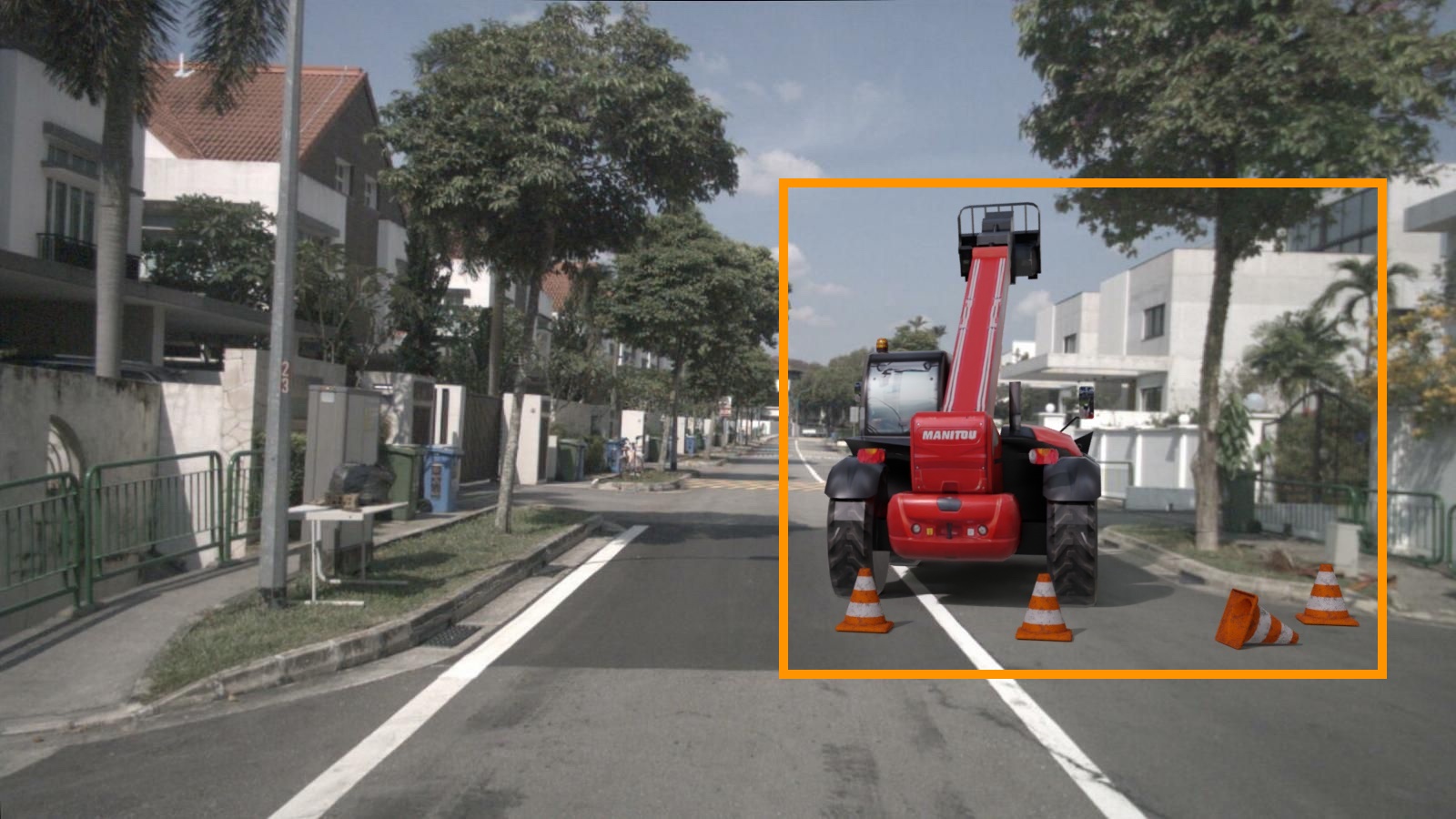} &
\includegraphics[width=0.28\linewidth,height=2.1cm,trim=110 20 0 0,clip]{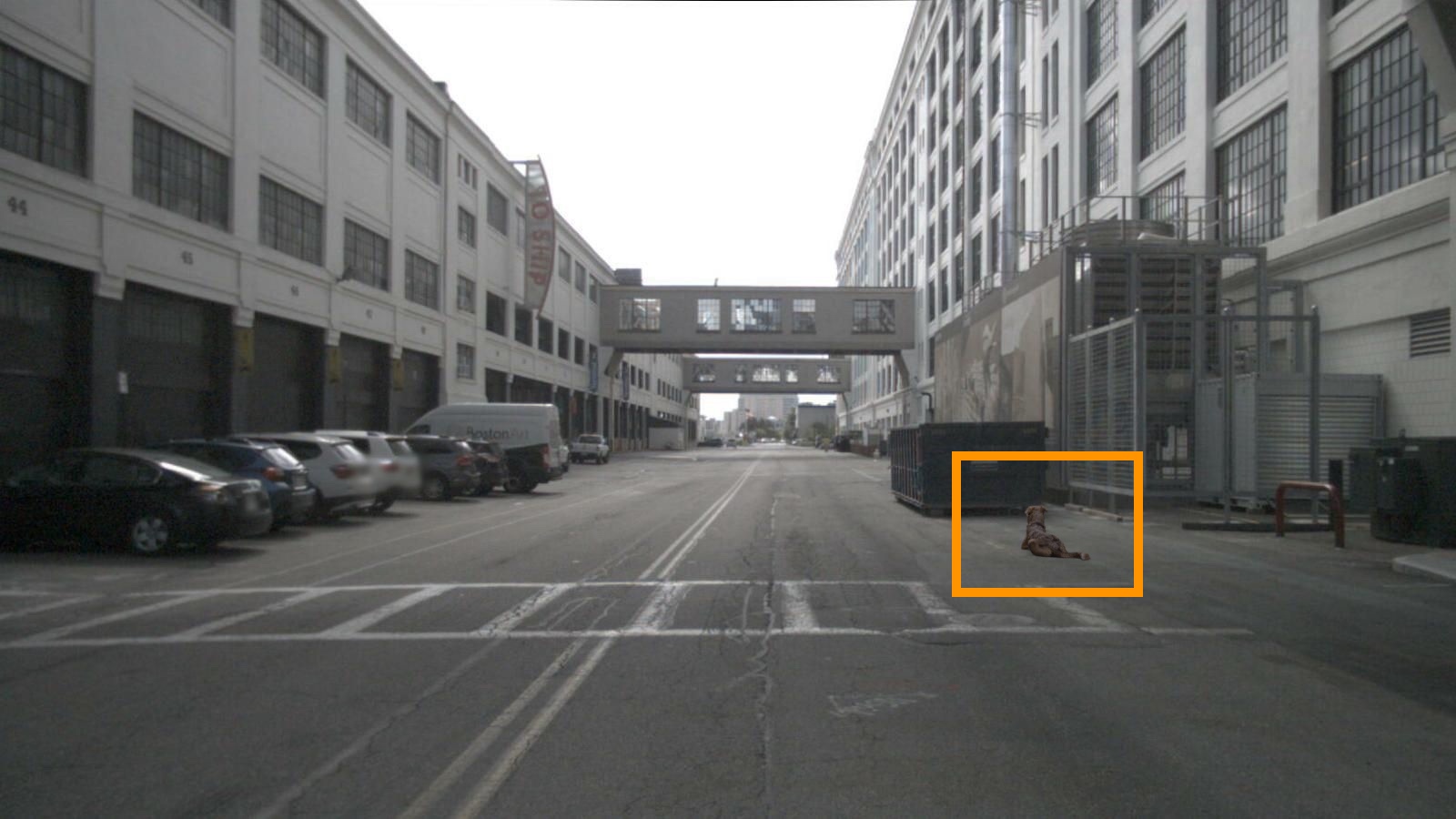} &
\includegraphics[width=0.28\linewidth,height=2.1cm,trim=50 25 50 0,clip]{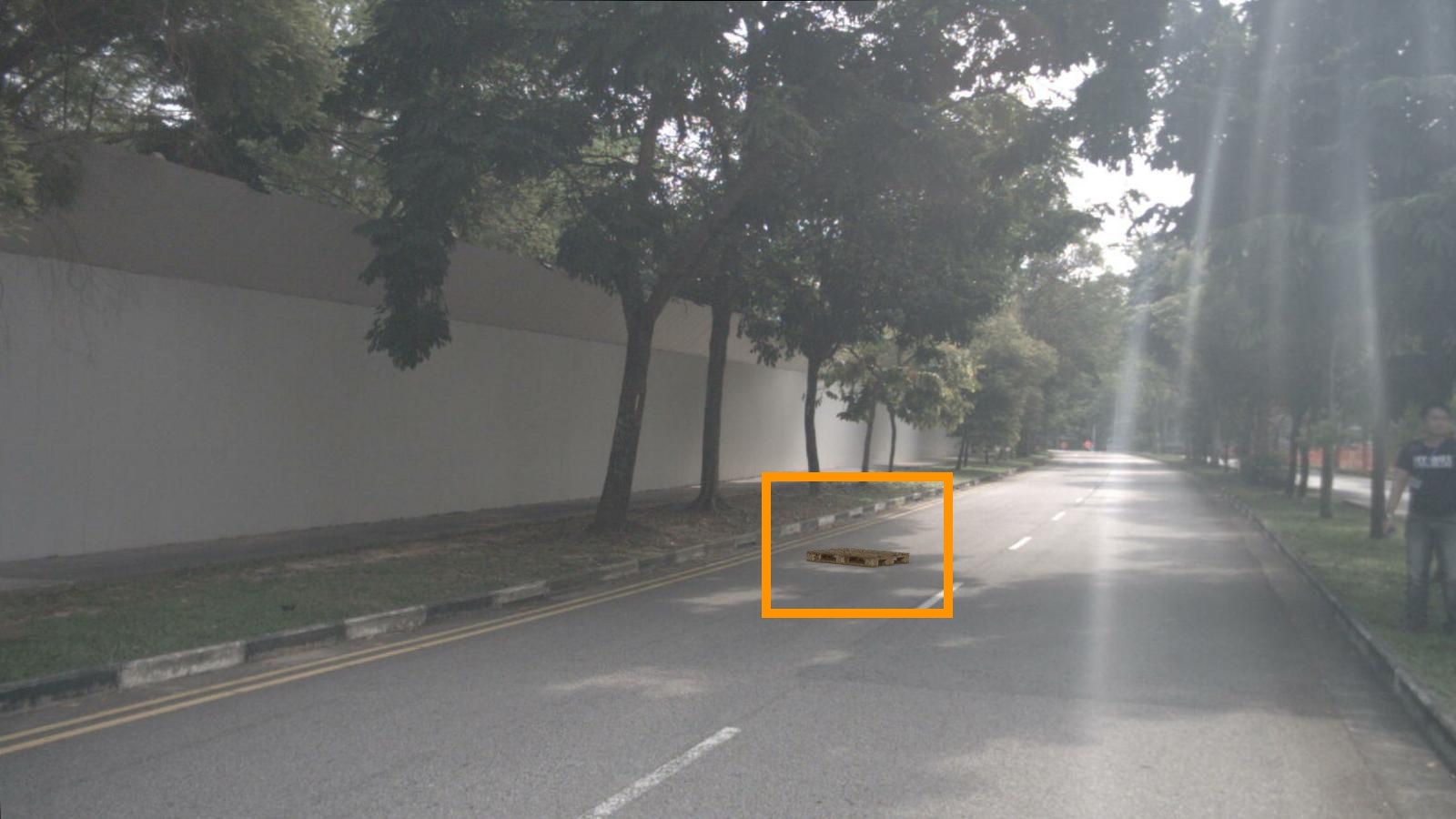} \\
\end{tabular}
}
\vspace{-1mm}
\caption{We estimate lighting and perform virtual objects insertion in real street scenery. We insert cars and heavy vehicles (top), and composite rare but safety-critical scenarios with garbage, construction site, a dog, and debris (bottom).
3D assets provided courtesy of TurboSquid and their artists Hum3D, be fast, rabser, FirelightCGStudio, amaranthus and 3DTree\_LLC.
}
\label{fig:teaser}
\end{figure}

In this work, we address the task of outdoor lighting estimation from monocular imagery, specifically focusing on street scenes, as shown in Fig.~\ref{fig:teaser}. This is an important task, as it enables virtual object insertion 
that can cater to many downstream domains~\cite{chen2021geosim,Ost_2021_CVPR,hong2018learning,Su_2015_ICCV,Dwibedi_2017_ICCV}, such as virtually inserting newly planned buildings for architectural visualization, realistically rendering game characters into surroundings, or as a way to augment real datasets with objects that are otherwise hard to record in the real world, such as road debris and exotic animals, for the purpose of training more robust and performant computer vision models. 

Lighting estimation for AR applications needs to account for complex 5D light transport~\cite{Adelson1991}, \ie a function of spatial location and viewing direction. 
With usually a limited field-of-view observed from input, the task of estimating the light field is challenging and ill-posed. 
An additional challenge encountered for outdoor scenes, in contrast to indoor scenes, is the extreme high dynamic range (HDR) of the sun, which is critical to estimate correctly in order to render cast shadows. 
Existing literature on lighting estimation usually tackles a simplified problem setting. \cite{wang2021learning,srinivasan2020lighthouse} focus on spatially-varying effects but do not handle HDR intensities. 
In contrast, methods that focus on HDR and predict parametric sky~\cite{hold2017deep,zhang2019all} or utilize learned sky models~\cite{hold2019deep} typically ignore the spatially-varying effects and lack high-frequency details. 
These limitations not only result in inaccurate lighting estimation but also hamper virtual object insertion effects.

In this paper, we propose a unified approach that overcomes the previously mentioned limitations, estimating the HDR scene light field from a single image. 
Tailored to outdoor scenes, we estimate a hybrid lighting representation that comprises of two components: an HDR sky dome and a volumetric lighting representation for the surrounding scene. 
We employ a learned latent vector to represent the sky dome inspired by~\cite{hold2019deep}, which can be decoded into an HDR environment map that is designed to model the strong intensity of the sun. We adopt the volumetric spherical Gaussian representation~\cite{wang2021learning} to represent the non-infinity surroundings such as road and buildings. 
The two components naturally combine with volume rendering and demonstrate superiority over prior works~\cite{hold2019deep,wang2021learning}. 
We further design a physics-based object insertion formulation that renders the inserted objects and their shadows cast on the scene. We utilize ray-tracing to capture the second-order lighting effects, which is fully differentiable with respect to the lighting parameters. We train our method with supervised and self-supervised losses, and show that adversarial training over the composited AR images provides complementary supervisory signal to improve lighting estimation.

Our method outperforms prior work in the tasks of lighting estimation and photorealistic object insertion, which we show through numerical results and a user study. 
We further showcase our virtual object insertion through the application of 3D object detection in autonomous driving. Our approach, which can render synthetic 3D  objects into real imagery in a realistic way, provides useful data augmentation that leads to notable performance gains over the vanilla dataset, and a naive insertion method that does not account for lighting.

%% file: related.tex
\section{Related Work}

\vspace{-1mm}
\mysubsubsection{Lighting estimation} 
aims to predict an HDR light field from image observations. Due to the ill-posed nature of the problem, 
prior works often tackle a simplified task and ignore spatially-varying effects, using lighting representations such as spherical lobes~\cite{boss2020two,li2018learning}, light probes~\cite{legendre2019deeplight}, sky parameters~\cite{hold2019deep,hold2017deep,zhang2019all}, and environment maps~\cite{gardner2017learning,neuralSengupta19,wei2020object,hdr-environment-map-estimation}. 
Recent works explored various representations for capturing spatially-varying lighting, including per-pixel spherical lobes~\cite{garon2019fast,li2020inverse,zhao2020pointar}, light source parameters~\cite{gardner2019deep}, per-location environment map \cite{song2019neural,zhu2021cvpr} and 3D volumetric lighting~\cite{srinivasan2020lighthouse,wang2021learning}. These works usually train with synthetic data due to the scarcity of available groundtruth HDR light field on real-world captures. 

Outdoor scenes require special attention for the extreme High Dynamic Range (HDR) lighting intensity, \eg the sun's intensity, which is several orders higher in magnitude than typical light sources found in indoor scenes. 
Prior works employed sky parameters~\cite{hold2017deep,zhang2019all}, or learned sky models~\cite{hold2019deep} with an encoder-decoder architecture for modeling HDR sky. However, these works typically only focus on modeling the sky, and ignore the high-frequency and spatially-varying effects, which are equally important to get right for AR applications. In this work, we propose a unified representation that can handle both HDR sky intensity as well as spatially varying effects in outdoor scenes to achieve better performance.

\mysubsubsection{Self-supervised lighting estimation methods} apply differentiable rendering to provide gradient for lighting estimation~\cite{DIBR19,chen2021dibrpp,StyleGAN3D,Li:2018:DMC,NimierDavidVicini2019Mitsuba2}.
Differentiable rasterization-based renderers~\cite{DIBR19,StyleGAN3D,chen2021dibrpp} are typically limited to images of single objects and ignores the spatially-varying effects. 
Physically-based rendering (PBR) methods~\cite{Li:2018:DMC,NimierDavidVicini2019Mitsuba2} require intensive memory and running time, and are thus limited to optimization tasks. 
Note that existing differentiable renderers~\cite{DIBR19,chen2021dibrpp,NimierDavidVicini2019Mitsuba2} typically do not provide direct functionality for object insertion, which is an image editing task. 
We propose a novel differentiable object insertion formulation, providing valuable supervision signal for lighting estimation.

\mysubsubsection{Image manipulation.} 
Related to ours is also work that aims to insert synthetic objects into images using alternative techniques, such as adversarial methods~\cite{ling2020variational,kim2021grivegan}, or by perturbing real scenes using recent advances in neural rendering~\cite{Ost_2021_CVPR}. 
Alhaija \etal~\cite{alhaija2018augmented} assumes known lighting and propose to use AR as a data generation technique by inserting synthetic assets into real world scenes.
Naive copy-paste object insertion has also been shown to boost downstream object recognition accuracy~\cite{Su_2015_ICCV,Dwibedi_2017_ICCV}. 
Neural Scene Graph~\cite{Ost_2021_CVPR} optimize neural implicit functions for each object in the scene. Despite realistic editing results, lighting effects are baked into the representation and thus swapping assets from one scene to another is not easily possible.
GeoSim~\cite{chen2021geosim} reconstructs assets such as cars from real-world driving sequences, and inserts them into a given image using a classical renderer followed by a shallow neural renderer that fixes errors such as unrealistic compositing effects. 
While achieving impressive results on car insertion, it remains difficult to apply on less frequent objects. 
In contrast, our method supports inserting 3D assets of various classes (Fig.~\ref{fig:teaser}).

%% file: method.tex

\section{Method} 
\label{sec:method}

\begin{figure*}[t!]
\begingroup
\centering
{\includegraphics[width=0.99\textwidth]{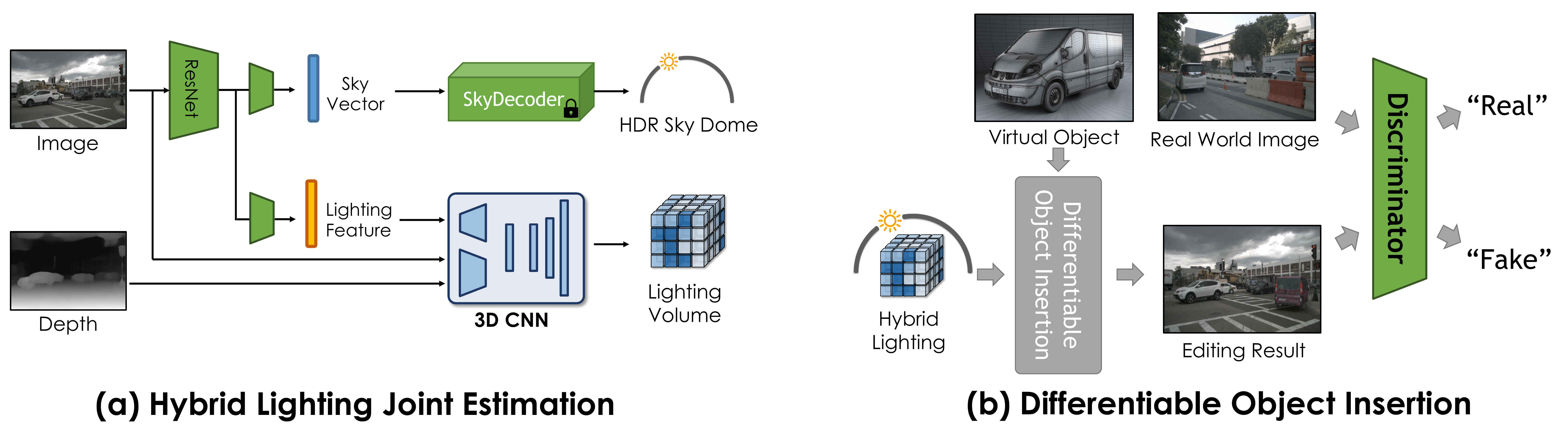}}
\vspace{-3mm}
\caption{\textbf{Model overview.} 
Our monocular lighting estimation model (a) predicts a hybrid lighting representation containing an HDR sky dome (top) representing sky and sun at infinity, and a lighting volume (bottom) representing the surrounding scene. 
The depth (a, left) for lighting volume prediction comes from off-the-shelf monocular depth estimator \cite{packnet}. 
With the predicted lighting, our object insertion module (b) renders a 3D asset into a given image and is fully differentiable w.r.t. lighting parameters, thus enabling end-to-end training with adversarial objective for photorealism. 
}
\label{fig:model} 
\endgroup
\end{figure*}

We aim to estimate scene lighting so as to render synthetic 3D assets into images realistically. 
In what follows, we introduce the hybrid representation, which comprises an HDR sky dome and a volumetric scene lighting (Sec.~\ref{sec:lighingrepresent}), the hybrid lighting prediction model (Sec.~\ref{sec:lighingnetwork}, Fig.~\ref{fig:model}a), the differentiable object insertion module that renders a virtual object into an image (Sec.~\ref{sec:objectinsertion}, Fig.~\ref{fig:model}b), and the training schema (Sec.~\ref{sec:training}).

\subsection{Hybrid Lighting Representation} 
\label{sec:lighingrepresent}

Our goal is to model the 5D light field, which maps a spatial location $\mathbf{x} \in \mathbb{R}^3$ and light direction $\mathbf{l} \in \mathcal{S}^2$ into an HDR radiance value $\mathbf{r} \in \mathbb{R}_+^3$. In contrast to indoor, outdoor scenes require simultaneously modeling the extreme HDR sky as well as the surrounding environment. The peak magnitude of the former (sun) can be several orders higher than the latter. 
To address this, we propose to use a hybrid lighting representation that separately models the sky at infinity and the surrounding scene. 
This decomposition allows us to capture both, the extreme intensity of the sky while preserving the spatially-varying effects of the scene. 

\mysubsubsection{{HDR sky representation.}}  
The sky dome typically contains a relatively simple structure, \ie sun, sky, and possibly clouds, which affords a much lower dimensional representation than that of a typical environment map. 
Thus, instead of directly predicting a high resolution 2D environment map, we learn a feature space of the sky, and represent the sky dome with a sky feature vector $\textbf{f} \in \mathbb{R}^d$, which can further be passed to a pretrained CNN decoder to decode into an HDR environment map, as shown in Fig.~\ref{fig:model}a (top).  The sky feature space learning is described in Sec.~\ref{sec:lighingnetwork}, \textit{sky modeling} part.

\mysubsubsection{{Spatially-varying surrounding scene representation.}}  
The outdoor scenes generally consist of complex geometric structures resulting in location-dependent lighting effects like shadows and reflections, which cannot be simply modeled as an environment map. 
To address this, we use a volumetric spherical Gaussian (VSG)~\cite{wang2021learning} to represent the nearby surrounding scene. 
VSG is a 8-channel volumetric tensor $L_\text{VSG} \in \mathbb{R}^{8 \times X \times Y \times Z}$, augmenting the RGB$\alpha$ volume with view-dependent spherical Gaussian lobes.
Each voxel of VSG contains a spherical Gaussian lobe 
$G(\mathbf{l}) = \bm{c}e^{-(1 - \mathbf{l} \cdot \bm{\mu}) / \sigma^2}$, 
where $\mathbf{l}$ is viewing direction and $\bm{\xi} = \{\bm{c}, \bm{\mu}, \sigma\}$ are 7-dimensional parameters to model the exiting radiance of the corresponding 3D location. Each voxel also has an alpha channel $\alpha \in [0, 1]$ to represent occupancy. 
The 5D light field can be queried with volume rendering which we detail below. 

\mysubsubsection{{Radiance query function.}} 
With our hybrid lighting representation utilizing both sky dome and volumetric lighting, the lighting intensity at any 3D point along any ray direction can be queried.
To compute the radiance of a ray that starts inside the volume, we first shoot the ray through the lighting volume $L_\text{VSG}$ and finally hit the HDR sky dome $L_\text{env}$. We adopt alpha compositing to combine the two lighting effects.
Specifically, to compute the lighting intensity for the ray emitted from the location $\mathbf{x} \in \mathbb{R}^3$ in the direction $\mathbf{l} \in \mathcal{S}^2$, we select $K$ equi-spaced locations along the ray and use nearest neighbor interpolation to get the voxel values $\{\alpha_{k}, \bm{\xi}_k\}_{k=1}^K$ from $L_\text{VSG}$. We then query the intensity of $L_\text{env}$ in the direction $\mathbf{l}$, referred to as $L_\text{env}(\mathbf{l})$, via bilinear interpolation. The final HDR light intensity $L(\mathbf{x}, \mathbf{l}) \in \mathbb{R}_+^3$ can be computed using volume rendering:
\begin{equation}
	\label{eq:alphacomp}
	L(\mathbf{x}, \mathbf{l}) = \Big(\sum_{k=1}^{K} \tau_{k-1} \alpha_{k} G(-\mathbf{l}; \bm{\xi}_k)\Big) + \tau_{K} L_\text{env}(\mathbf{l}) 
\end{equation}
where $\tau_k = \prod_{i=1}^{k} (1-\alpha_{i})$ is the transmittance. This function will be used for rendering object insertion in Eq.~\ref{eq:MC_shading} and shadows in Eq.~\ref{eq:ratio}.

\begin{figure*}[t!]
\begingroup
\centering
{\includegraphics[width=0.99\textwidth]{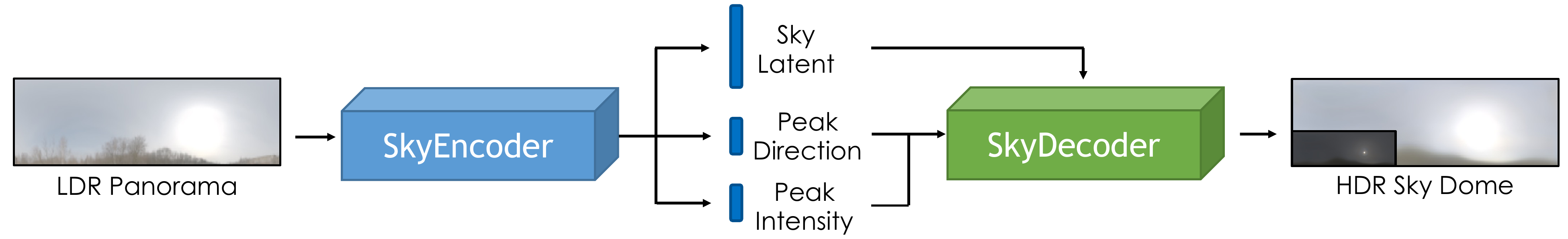}}
\vspace{-3mm}
\caption{\textbf{Sky modeling network} takes as input an LDR panorama and produces an HDR sky with an encoder-decoder structure, where the network also learns to compress sky information into the intermediate vector representation $\textbf{f} \in \mathbb{R}^d$. 
The sky vector consists of explicit feature of peak intensity and direction, and a latent feature vector. 
}
\label{fig:skymodel} 
\endgroup
\end{figure*}

\subsection{Network Architecture}
\label{sec:lighingnetwork}

In this section, we introduce the network architecture for the pre-trained sky model and lighting prediction. 

\mysubsubsection{{Sky modeling.}}
To learn the sky feature space, we design a sky modeling network as shown in Fig.~\ref{fig:skymodel}. 
Specifically, the encoder compresses the input LDR panorama into a feature vector, then the decoder decode it to the original HDR sky dome. 
Our sky feature vector $\textbf{f}$ contains explicit HDR peak intensity ${\mathbf{f}}_\text{intensity} \in \mathbb{R}^3$, peak direction ${\mathbf{f}}_\text{dir} \in \mathbb{R}^3$, and a latent vector $\textbf{f}_\text{latent} \in \mathbb{R}^{d-6}$ encoding the content of the sky dome. This design allows us to control the lighting of the sun by controlling ${\mathbf{f}}_\text{dir}$ and ${\mathbf{f}}_\text{intensity}$, which is convenient for applications that allow manual editing. 

We pre-train the sky encoder-decoder network on a set of outdoor HDR panoramas and keep it freezed. 
Once trained, we integrate the fixed decoder into the sky prediction branch (Fig.~\ref{fig:model}a top). 
The pre-trained sky modeling network, which maps an LDR panorama into HDR, will also be used to generate HDR pseudo labels for supervision (Sec.~\ref{sec:training}) due to the lack of HDR data.

\mysubsubsection{Hybrid lighting prediction.}
As shown in Fig.~\ref{fig:model}, we use a two-branch network to predict the sky dome and the lighting volume. We detail each branch below. 

\myparagraph{{HDR sky prediction branch.}}
Given an input image, the sky branch directly predicts the sky feature vector $\mathbf{f}$ from the ResNet~\cite{resnet} backbone. The pre-trained fixed sky decoder then maps $\mathbf{f}$ to an HDR sky environment map.

\myparagraph{{Lighting volume prediction branch.}} 
We adapt from a subnetwork of~\cite{wang2021learning} for VSG prediction. 
Specifically, we first use an MLP~\cite{mescheder2019occupancy} to map the lighting feature extracted by ResNet backbone into a feature volume, and then unproject the input image into a RGB$\alpha$ volume. 
We adopt a 3D UNet to fuse the two volumes and predict the VSG lighting representation. 
Since unprojection requires depth information, we adopt an off-the-shelf monocular depth estimator PackNet~\cite{packnet} to predict a dense depth map. 
Architecture details are included in the Appendix. 

\subsection{Differentiable Object Insertion} 
\label{sec:objectinsertion}

We now study the task of realistic object insertion, which is key in AR and our main final objective. Given an estimated lighting representation $L$, our goal is to composite a virtual object with known geometry $\mathcal{M}$ and material $\Theta$ into a real image $I$ with known depth $D$ and camera intrinsics.
To achieve realistic lighting effects in this process, not only the inserted object should be influenced by the scene lighting, but it should also affect the scene to create cast shadows. 

In our work, we aim to make this insertion module differentiable, including shadow rendering, such that it can afford gradient backpropagation from the losses defined on the composited image, back to our lighting parameters. 
Since groundtruth light field is not easily available for outdoor scenes, we argue that a plausible complementary supervision signal is to use the quality of object insertion as an objective. Thus, carefully designed adversarial training on the composite images can be a powerful approach to supervise lighting estimation. We also argue that even if groundtruth lighting information would be available, optimizing for the quality of object insertion end-to-end will likely lead to improved results.

\mysubsubsection{Foreground object appearance rendering.} 
We render the virtual object with our predicted hybrid lighting representation $L$ using a physically-based renderer. We adopt the 
Disney BRDF~\cite{burley2012physically,karis2013real,chen2021dibrpp} for enhanced realism. 

Specifically, we first shoot rays from the camera origin to the scene, where we apply ray-mesh intersection detection for the rays and the inserted object $\mathcal{M}$. For each intersected ray, we create a G-buffer for the location of the intersection $\mathbf{x}$, the surface normal $\mathbf{n}$ and the material properties $\mathbf{\theta}$. We bounce multiple rays at $\mathbf{x}$ and render with Monte-Carlo numerical integration:
\begin{equation} 
\label{eq:MC_shading}
	I_\mathbf{x} = \frac{1}{N} \sum_{k=1}^{N} 
    	\frac{f(\mathbf{l}_k, \mathbf{v}; \mathbf{\theta}) L(\mathbf{x}, \mathbf{l}_k) (\mathbf{n} \cdot \mathbf{l}_k)^+}
    	{p(\mathbf{l}_k)} 
\end{equation} 
where $L(\cdot,\cdot)$ is the radiance query function as defined in Eq.~\ref{eq:alphacomp}, $N$ and $\mathbf{l}_k$ is the number and direction of sampled lighting, $\mathbf{v}$ is the viewing direction of the camera ray, and $f$ is the Disney BRDF.

\mysubsubsection{Background shadow map rendering.} 
The inserted object changes the light transport in the scene and affects the appearance of the background scene pixels, which typically causes shadows. We adopt ray tracing to generate faithful ratio shadow maps for the inserted object, inspired by classic ratio imaging techniques~\cite{peers2007post}. 
~
Specifically, for each scene pixel $p$, we compute its 3D location $\mathbf{x}$ from the depth map $D$. We first compute the lighting distribution before object insertion $\{L(\mathbf{x}_\text{s}, \mathbf{l}_k)\}_{k=1}^{N_\text{s}}$, where $\{\mathbf{l}_k\}_{k=1}^{N_\text{s}}$ are uniformly selected light directions on the upper hemisphere. 
After object insertion, the rays 
may potentially get occluded by the inserted objects, resulting in a post-insertion lighting distribution $\{L^{'}(\mathbf{x}_\text{s}, \mathbf{l}_k)\}_{k=1}^{N_\text{s}}$. 
To compute $\{L^{'}(\mathbf{x}_\text{s}, \mathbf{l}_k)\}_{k=1}^{N_\text{s}}$, we perform ray-mesh query for all  rays. If ray $(\mathbf{x}_\text{s}, \mathbf{l}_k)$ is occluded by the inserted object, we set the lighting intensity value to an ambient value $I_\text{a}$ which we empirically set to 0.1, while the lighting intensities of unoccluded rays remain the same as the original radiance. We then define the shadow effects as the ratio of the pixel intensity values before ($I$) and after ($I^{'}$) object insertion, 
\begin{equation}
\label{eq:shadow1}
\begin{aligned}
S_{p} &= \frac{I^{'}_{p}}{I_{p}} 
 =  \frac{\sum_{k=1}^{N_\text{s}} f_\text{scene}(\mathbf{x}_\text{s}, \mathbf{l}_k, \mathbf{v}; \mathbf{\theta}) L^{'}(\mathbf{x}_\text{s}, \mathbf{l}_k) (\mathbf{n} \cdot \mathbf{l}_k)^+} 
 {\sum_{k=1}^{N_\text{s}} f_\text{scene}(\mathbf{x}_\text{s}, \mathbf{l}_k, \mathbf{v}; \mathbf{\theta}) L(\mathbf{x}_\text{s}, \mathbf{l}_k) (\mathbf{n} \cdot \mathbf{l}_k)^+} 
\end{aligned}
\end{equation}
where the BRDF of the scene pixel $f_\text{scene}$ and normal direction $\mathbf{n}$ are unknown. 
As our object insertion and shadows occur on flat surfaces in typical street scenes we consider, we simplify it by assuming the normal direction is pointing upward, and assume the scene surface is Lambertian with constant diffuse albedo $f_\text{scene}(\mathbf{x}_\text{s}, \mathbf{l}_k, \mathbf{v}) = f_\text{d}$. 
As a result, we can move the BRDF term outside the sum in Eq.~\ref{eq:shadow1} and cancel it out to obtain a simpler term: 
\begin{equation}
\label{eq:ratio}
  S_{p}\!=\! \frac{f_\text{d}\! \sum\limits_{k=1}^{N_\text{s}}\! L^{'}\!(\mathbf{x}_\text{s}, \mathbf{l}_k) (\mathbf{n} \!\cdot\! \mathbf{l}_k)^+} 
	{f_\text{d}\! \sum\limits_{k=1}^{N_\text{s}}\! L(\mathbf{x}_\text{s}, \mathbf{l}_k) (\mathbf{n}\! \cdot\!  \mathbf{l}_k)^+}
	= \frac{\sum\limits_{k=1}^{N_\text{s}}\! L^{'}\!(\mathbf{x}_\text{s}, \mathbf{l}_k) (\mathbf{n}\! \cdot\! \mathbf{l}_k)^+}
	{\sum\limits_{k=1}^{N_\text{s}}\! L(\mathbf{x}_\text{s}, \mathbf{l}_k) (\mathbf{n}\! \cdot\! \mathbf{l}_k)^+} 
\end{equation}
which can be computed with the estimated lighting $L$. Scene pixels after insertion can then be computed by multiplying the ratio shadow map $I^{'} = S \odot I$.

\mysubsubsection{Gradient propagation.}
We design the forward rendering process to be differentiable for both foreground object and background shadows, which allows us to back propagate gradients from image pixels to the lighting parameters.

For each foreground pixel, the rendered appearance of the inserted object is differentiable via Eq.~\ref{eq:MC_shading}. 
Gradients from background pixels $I^{'}$ with respect to the lighting $L$, \ie $\frac{\partial I^{'}}{\partial L}$, can be computed via $\frac{\partial I^{'}}{\partial L} = \frac{\partial S}{\partial L} I$, where the shadow ratio $S$ in Eq.~\ref{eq:ratio} is also differentiable wrt. lighting $L$. 
Intuitively, if we want the shadows around the object to be perceptually darker, this will encourage the occluded light directions to have stronger intensity.

\subsection{Training} 
\label{sec:training}
We first pre-train the sky modeling network 
on a collection of outdoor HDR panoramas, and then keep it fixed in the following training process for our hybrid lighting prediction. 
The supervision for our hybrid lighting joint estimation module comes from two parts: 
(1) the direction supervision that learns lighting information from the training data, and 
(2) the adversarial supervision that applies on the final editing results and optimizes for realism. 

\mysubsubsection{Sky Modeling Supervision.}
\label{sec:skymodeltrain}
We train the sky modeling encoder-decoder network (Fig.~\ref{fig:skymodel}) on a collection of outdoor HDR panoramas. 
For each HDR panorama $I_\text{HDR}$, we compute its ground-truth peak intensity ${\mathbf{f}}_\text{intensity}$ and direction ${\mathbf{f}}_\text{dir}$. 
We train the encoder-decoder with the LDR-HDR pair $(I_\text{LDR}, I_\text{HDR})$, where the input LDR panorama $I_\text{LDR}$ is converted from the HDR panorama via gamma correction and intensity clipping. 
We supervise the network with a combination of three losses, 
including peak direction loss $\mathcal{L}_{\text{dir}}$ with L1 angular error, 
and peak intensity loss $\mathcal{L}_{\text{intensity}}$ and HDR reconstruction loss $\mathcal{L}_{\text{hdr}}$ using log-encoded L2 error defined as $\text{LogEncodedL2}(\hat{x}, x) = ||\log(1+\hat{x}) - \log(1+x)||_2^2$.

\mysubsubsection{Supervision for Hybrid Lighting Prediction.}
\label{sec:directsupervision}
To supervise lighting prediction, we use two complementary datasets: the self-driving dataset nuScenes~\cite{nuscenes2019}, and the panoramic street view dataset HoliCity~\cite{zhou2020holicity}. 
As both datasets are LDR, we predict HDR pseudo labels from the pre-trained modeling network (Fig.~\ref{fig:skymodel}) by lifting HoliCity LDR panoramas into HDR. 
In what follows, we descibe the supervision for the sky prediction branch, the lighting volume prediction branch, and the loss signal to combine the two representation. 

\myparagraph{HDR sky branch losses.} We train our sky 
branch on HoliCity~\cite{zhou2020holicity}, which contains LDR panoramas $I_\text{pano}$ with sky masks $M_\text{sky}$ and sun location $\mathbf{f}_\text{dir}$. 
To compute HDR pseudo labels, we feed the LDR panorama $I_\text{pano}$ into the pre-trained sky  network, and get the estimated sky peak intensity $\tilde{\mathbf{f}}_\text{intensity}$ and latent code $\tilde{\mathbf{f}}_\text{latent}$ as pseudo groundtruth to supervise our sky prediction branch. 
In a training step, we crop a perspective image $I_\text{crop}$ from the panorama as input to our lighting estimation network, and predict the sky vector output $(\hat{\mathbf{f}}_\text{intensity}, \hat{\mathbf{f}}_\text{dir}, \hat{\mathbf{f}}_\text{latent})$ and the reconstructed HDR sky image $\hat{I}_\text{pano}$. 
We use a combination of the log-encoded L2 loss for peak intensity $(\hat{\mathbf{f}}_\text{intensity}, \tilde{\mathbf{f}}_\text{intensity})$, L1 loss for latent code $(\hat{\mathbf{f}}_\text{latent}, \tilde{\mathbf{f}}_\text{latent})$, L1 angular loss for peak direction $(\hat{\mathbf{f}}_\text{dir}, {\mathbf{f}}_\text{dir})$, and L1 reconstruction loss between $(\hat{I}_\text{pano} \odot M_\text{sky}, {I}_\text{pano} \odot M_\text{sky})$ within the LDR sky region indicated by $M_\text{sky}$.

\myparagraph{Lighting volume branch loss.} 
Recall that images in the datasets are inherently groundtruth of a subset of the light field captured by camera sensor rays, \eg the videos captured by self-driving cars in nuScenes~\cite{nuscenes2019} and the panoramas in HoliCity~\cite{zhou2020holicity}. Meanwhile, the predicted hybrid lighting representation supports radiance query along arbitrary rays as shown in Eq.~\ref{eq:alphacomp}. 
Thus, we can enforce the consistency of radiance between our predicted light field and captured image groundtruth, given known camera pose and intrinsics. 

Specifically, we sample images from nuScenes, and crop perspective images from HoliCity panoramas as input images. We then predict the corresponding lighting volume together with the sky dome, query the radiance of the camera rays, and enforce consistency with ground truth captured images using L2 loss. Following~\cite{wang2021learning}, we also render the alpha channel into a depth map and enforce consistency with groundtruth depth. 

\myparagraph{Sky separation loss.} 
Intuitively, the real world camera rays that directly reach the sky should also transmit through the lighting volume and hit the sky environment map. 
With the losses mentioned above, the model may still fall into the degenerate case where the lighting volume completely occludes the sky. 
To address this, we use the sky mask $M_\text{sky}$ information to supervise the sky transmittance $\tau_{K}$ in Eq.~\ref{eq:alphacomp} with binary cross entropy loss.

\mysubsubsection{Training Lighting via Object Insertion.}
\label{sec:advsupervision}
\label{sec:insertion}
Our final goal is to realistically insert virtual objects into images. 
We formulate the object insertion process in an end-to-end fashion and use a discriminator to supervise the perceptual lighting effects on the image editing results. 

Specifically, we collect a set of high quality 3D car models from Turbosquid\footnote{\url{www.turbosquid.com}}. Given an input image, we estimate scene lighting, randomly select a 3D asset, and insert it into the scene using our object insertion module to get  $\hat{I}_\text{edit}$. 
We use the map information available in nuScenes to place the car on a driveable surface. We also perform  collision and occlusion checking with the depth map to avoid unrealistic object insertion due to erroneous placement. 
As shown in Fig.~\ref{fig:model}b, we use a discriminator to judge the quality of $\hat{I}_\text{edit}$ compared to real cars, and employ adversarial supervision to optimize for realism of insertion: $\mathcal{L}_\text{adv} = -\mathcal{D}(\hat{I}_\text{edit})$. 
Intuitively, a discriminator could easily detect erroneous shadow direction and intensity, and error in specular highlights.
Through the adversarial supervision, the estimated lighting is encouraged to produce object insertion results similar to real world image samples. 
We refer to further analysis in the Appendix.

%% file: results.tex
\section{Experiments}
\label{sec:results}

We extensively evaluate our method both qualitatively and quantitatively. We first provide experiment details (Sec.~\ref{sec:trainingdetials}). 
We then compare lighting estimation, evaluate the quality of object insertion (Sec.~\ref{sec:lightingcomp})
and perform ablation study (Sec.~\ref{sec:ablationstydt}). 
Finally, we show that our AR data helps downstream self-driving perception tasks (Sec.~\ref{sec:downstream}).

\begin{table}[t!]
\begin{minipage}{0.48\linewidth}
\centering
\resizebox{1\textwidth}{!}{
	\begin{tabular}{|l|c|}
		\hline
		Method & \tabincell{c}{Median angular error $\downarrow$}  \\
		\hline
		Hold-Geoffroy \etal~\cite{hold2019deep} 	& $24.88^\circ$      \\
		Wang \etal~\cite{wang2021learning}			& $53.86^\circ$      \\
		Ours                            	& $\bm{22.43}^\circ$ \\
		Ours (w/o sky modeling)  			& $31.45^\circ$ \\
		Ours (w/o adv. supervision)  		& $24.16^\circ$ \\
		\hline
	\end{tabular}
} 
\end{minipage}
\hspace{1mm}
\begin{minipage}{0.47\linewidth}
\caption{Quantitative results of peak direction on HoliCity~\cite{zhou2020holicity}. We outperform past work, and each component (sky modeling, adversarial supervision) helps.
} 
\label{table:quant_peak} 
\end{minipage}
\end{table} 

\begin{table}[t!]
\begin{minipage}{0.48\linewidth}
\centering
\resizebox{1\textwidth}{!}{
\begin{tabular}{|l|c|c|}
\hline
Method & \tabincell{c}{PSNR $\uparrow$}	& \tabincell{c}{si-PSNR $\uparrow$}  \\
\hline
Hold-Geoffroy \etal~\cite{hold2019deep} 	& $9.33$	& $10.73$    \\
Hold-Geoffroy \etal~\cite{hold2019deep}* 	& $10.81$	& $14.20$    \\
Wang \etal~\cite{wang2021learning}			& $14.06$	& $15.28$    \\
Ours (w/o adv. supervision)  				& $14.23$	& $15.31$ 	\\
Ours                            			& $\bm{14.49}$	& $\bm{15.35}$ 	\\
\hline
\end{tabular}
} 
\end{minipage}
\hspace{1mm}
\begin{minipage}{0.47\linewidth}
\caption{\footnotesize Quantitative results of LDR appearance on the nuScenes dataset~\cite{nuscenes2019}. * indicates constraining the evaluation on the upper hemisphere. 
} 
\label{table:quant_ldr} 
\end{minipage}
\end{table}

\subsection{Experimental Details} 
\label{sec:trainingdetials}

\mysubsubsection{Lighting estimation.} 
Our lighting representation combines a sky feature vector and a VSG lighting volume. We set the dimension of the sky vector to be 64 and decode it to a 64x256 HDR sky dome. Different from ~\cite{wang2021learning}, we tailor the size of VSG lighting volume to be 256x256x64 (xyz) to accommodate 300x300x80 (meters$^3$) outdoor scenes. 
As outdoor scenes are larger in scale while visible scene surfaces are relatively dense in close-to-camera regions, we employ log projection to map the volume representation to a 3D scene location. 
Inference time of the lighting estimation network is 180ms per image, clocked on a TITAN V GPU.

\mysubsubsection{Object insertion.} 
Our object insertion module relies on a differentiable rendering process of both foreground object and background shadows. 
During training, we sample 5000 rays for foreground objects. For background shadows, we render a 160x90 resolution shadow map and sample $450$ rays per pixel to save memory and computation. 
After training, we do importance sampling for each pixel in foreground and can afford a high resolution shadow map for background to generate more realistic effects. 
During inference time, we also have the option to use commercial renderer such as Blender~\cite{blender}, which we detail in the Appendix. 

\mysubsubsection{Datasets.}
We collected 724 outdoor HDR panoramas from online HDRI databases to train the sky encoder-decoder. 
We train the full model with nuScenes~\cite{nuscenes2019} and HoliCity~\cite{zhou2020holicity}.
For nuScenes, we use the official split containing 700 scenes for training and 150 scenes for evaluation. 
For HoliCity dataset, 
we apply 90\% v.s. 10\% data split for training and evaluation. 

\mysubsubsection{{Multi-view extension.}}
While we focus on monocular estimation, our model is extendable to multi-view input. It can consume multi-view images to predict more accurate lighting, as shown in Fig.~\ref{fig:qual_sv_shadow}. For the HDR sky prediction branch, we apply max pooling for ${\mathbf{f}}_\text{intensity}$, ${\mathbf{f}}_\text{latent}$, and average pooling to ${\mathbf{f}}_\text{dir}$ after rotating ${\mathbf{f}}_\text{dir}$ in different views to the canonical view. As for the volumetric lighting, since it is defined in the ``world'' coordinate space, we unproject and fuse multi-view images into a common lighting volume representation, akin to~\cite{philion2020lift}.

\begin{table}[t!]
\centering
\resizebox{0.7\linewidth}{!}{
\addtolength{\tabcolsep}{6pt}
\begin{tabular}{|l|c|}
\hline
Approach & \%  Ours (w/o adv. sup.) is preferred $\downarrow$\\
\hline 
Hold-Geoffroy \etal~\cite{hold2019deep} 	& $68.1 \pm 5.4$   \%  \\ 
Wang \etal~\cite{wang2021learning}      	& $94.2 \pm 2.0$   \%  \\
Ours                                 			& $\mathbf{40.6 \pm 10.2}$  \%  \\
\hline
\end{tabular}
} 
\vspace{0.5mm} 
\caption{\footnotesize \textbf{Quantitative results of user study}. Users compare baseline methods to an ablated version of our method (Ours w/o adv. supervision) in a pair-wise comparison. Each row reports the percentage of images that Ours w/o adv. supervision is preferred. Our method outperforms baselines, and adv.~supervision improves performance. 
} 
\label{table:userstudy} 
\end{table} 

\begin{figure*}[t!]
\centering
\begingroup
\setlength{\tabcolsep}{2pt}
\includegraphics[width=1\linewidth]{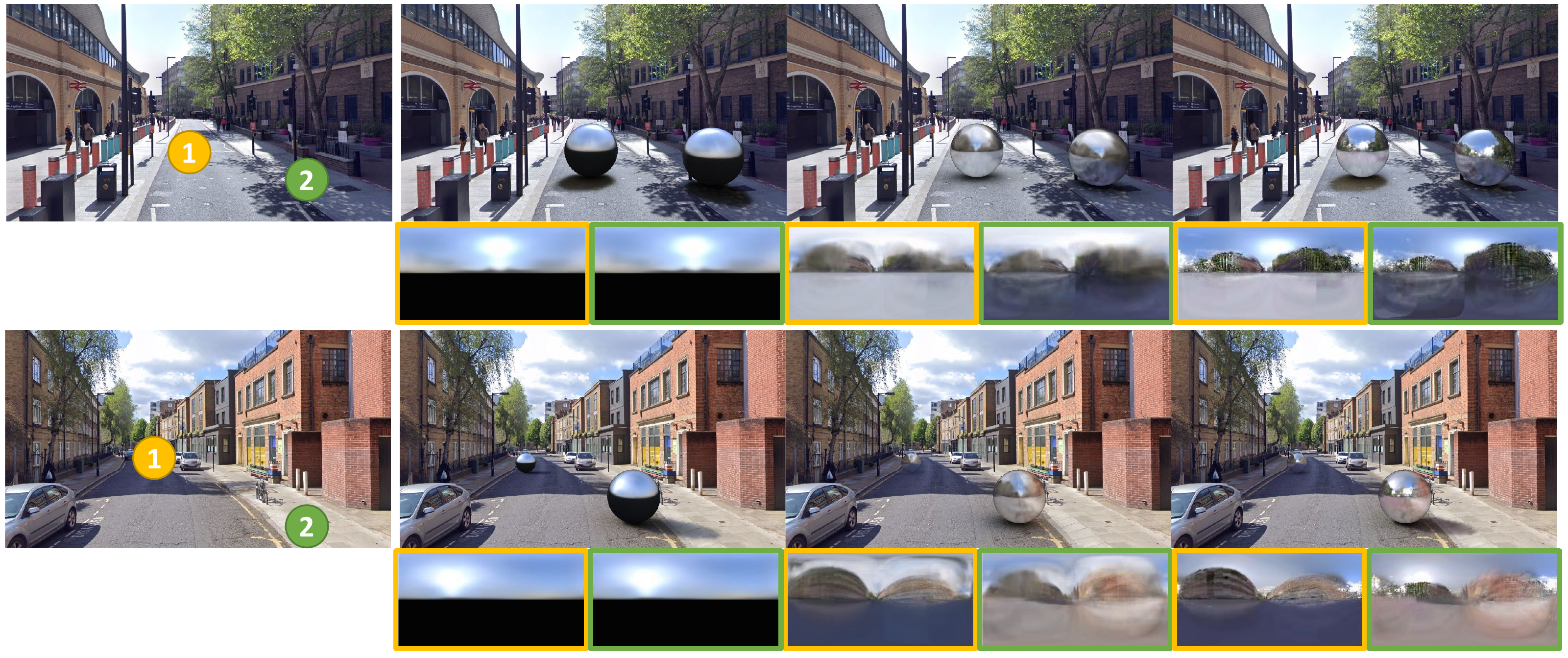}
\resizebox{0.99\textwidth}{!}{
\footnotesize
\begin{tabular}{C{0.3\textwidth}C{0.3\textwidth}C{0.3\textwidth}C{0.3\textwidth}}
Input image & Hold-Geoffroy \etal~\cite{hold2019deep} & Wang \etal~\cite{wang2021learning} & Ours \\
\end{tabular}
}
\endgroup
\caption{\textbf{Qualitative comparison of lighting estimation.} We insert a purely specular sphere into the image to visualize the lighting prediction, and display the environment maps on the bottom. Note the sun and environment map changes between locations. 
}
\label{fig:qual_sv_light} 
\end{figure*}

\subsection{Evaluation of Lighting Estimation} 
\label{sec:lightingcomp}

\mysubsubsection{Baselines.}
We compare with current state-of-the-art lighting estimation methods \cite{hold2019deep,wang2021learning}. 
Hold-Geoffroy \etal~\cite{hold2019deep} estimates the HDR sky environment map from a single image. Wang \etal~\cite{wang2021learning} predicts Volumetric Spherical Gaussian. 
We re-train or finetune these methods on the same data sources we used for our method to ensure a fair comparison.

\mysubsubsection{HDR evaluation of peak direction.}
We evaluate peak direction prediction on HoliCity dataset~\cite{zhou2020holicity}. We report the median angular error between the predicted direction and GT in Tbl.~\ref{table:quant_peak}. 
Wang \etal~\cite{wang2021learning} predicts HDR component in a self-supervised manner and cannot learn strong peaks. 
We also outperform Hold-Geoffroy \etal~\cite{hold2019deep}, which separately predicts a sky dome and its azimuth.

\mysubsubsection{LDR evaluation of novel view reconstruction.} 
Recall that any lighting representation, such as environment map and our hybrid lighting, aims to represent the complete or a subset of the light field, which can be rendered into images by querying the radiance function with specified camera rays. 
Prior work~\cite{srinivasan2020lighthouse,wang2021learning} proposed to use novel view radiance reconstruction PSNR as quantitative evaluation of the quality of lighting estimation, which we report on nuScenes dataset~\cite{nuscenes2019}. 

As the nuScenes-captured images may have different exposure values, we report both PSNR and scale invariant PSNR (si-PSNR) in Tbl.~\ref{table:quant_ldr}. For the latter, we multiply the predicted novel view with a scaling factor that minimizes L2 error.
Since Hold-Geoffroy \etal~\cite{hold2019deep} only predicts lighting on the upper hemisphere, we also constrain the evaluation on the upper hemisphere to make a fair comparison. 
Our method outperforms both baselines with a large margin, as~\cite{hold2019deep} ignores spatially-varying effects and usually predicts a sky dome with little high-frequency details. 
Our method also outperforms Wang \etal~\cite{wang2021learning} which cannot handle high HDR intensity of an outdoor scene.  

\begin{figure*}[t!]
\footnotesize
\centering
\begingroup
\setlength{\tabcolsep}{0.5pt}
\resizebox{0.99\linewidth}{!}{
\begin{tabular}{cccc}
\includegraphics[width=0.3\linewidth]{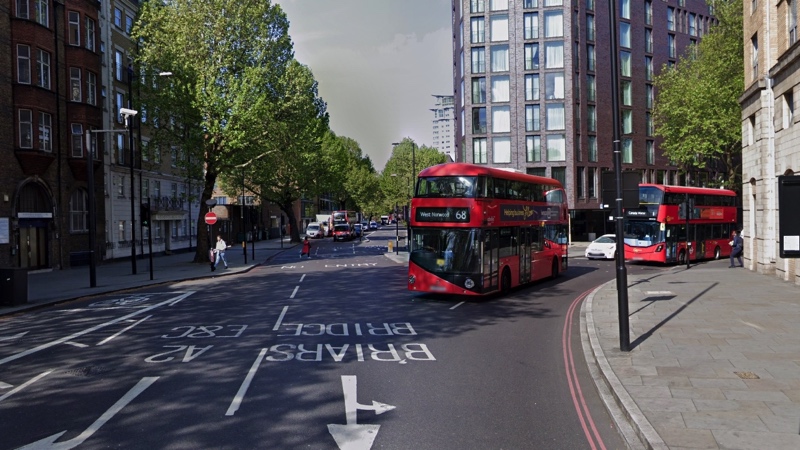} & 
\includegraphics[width=0.3\linewidth]{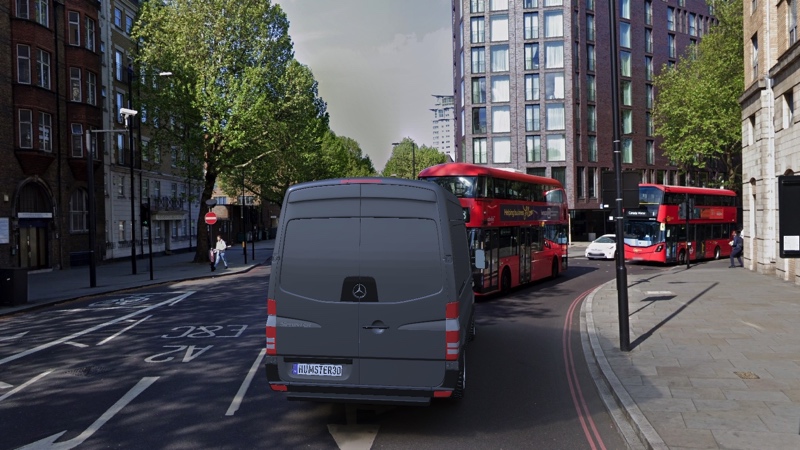} & 
\includegraphics[width=0.3\linewidth]{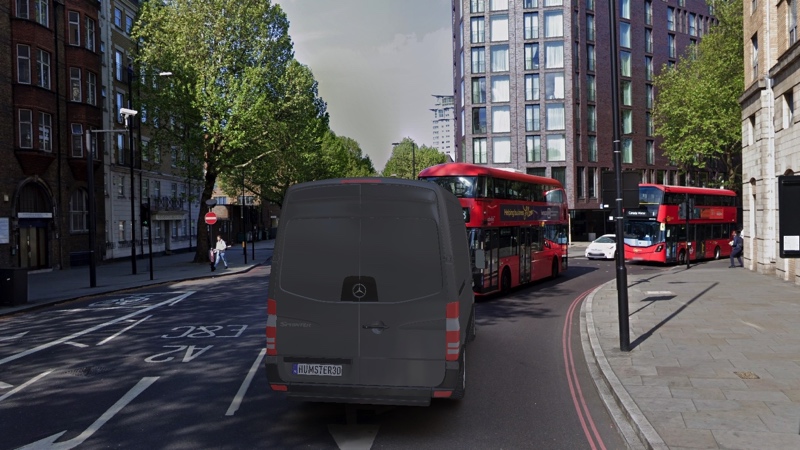} & 
\includegraphics[width=0.3\linewidth]{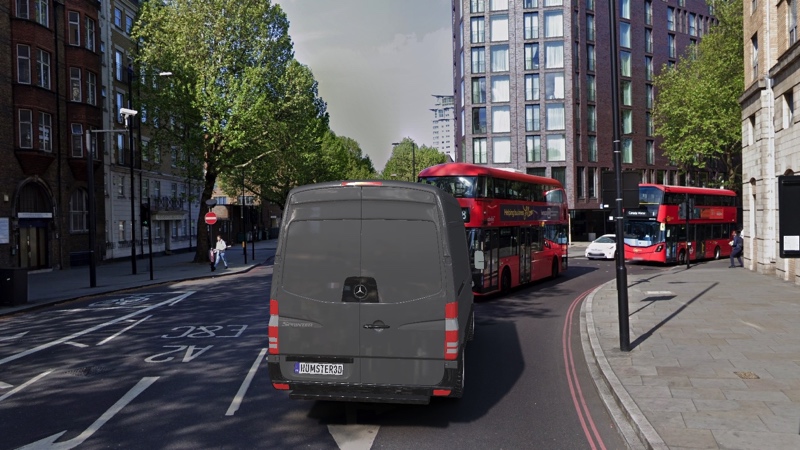} \\
\includegraphics[width=0.3\linewidth]{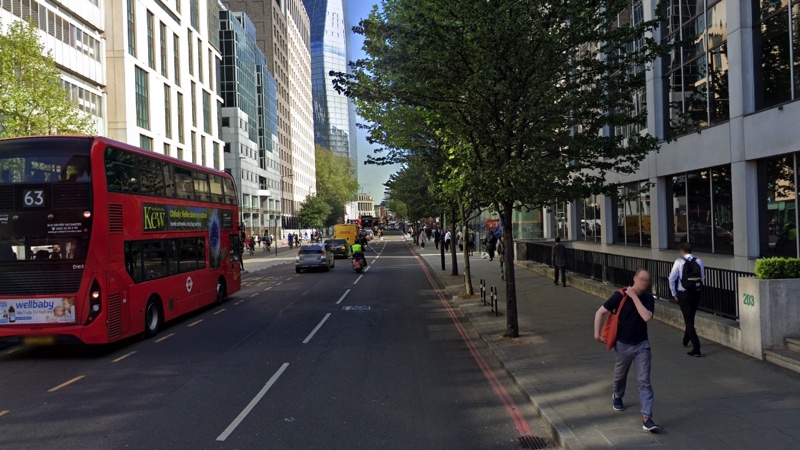} & 
\includegraphics[width=0.3\linewidth]{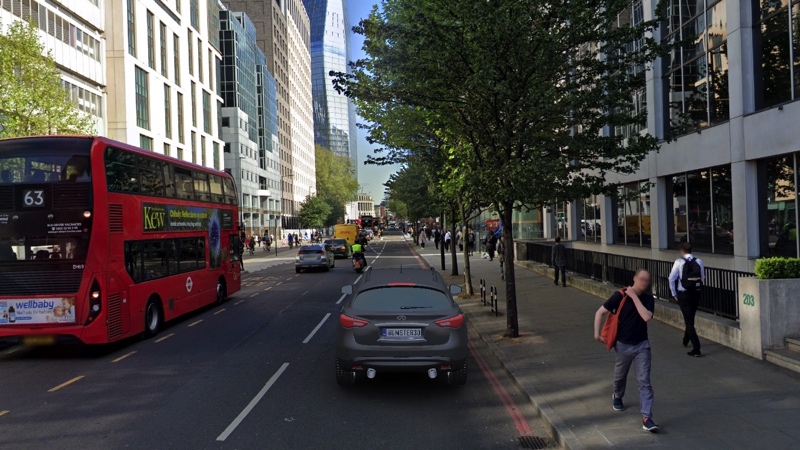} & 
\includegraphics[width=0.3\linewidth]{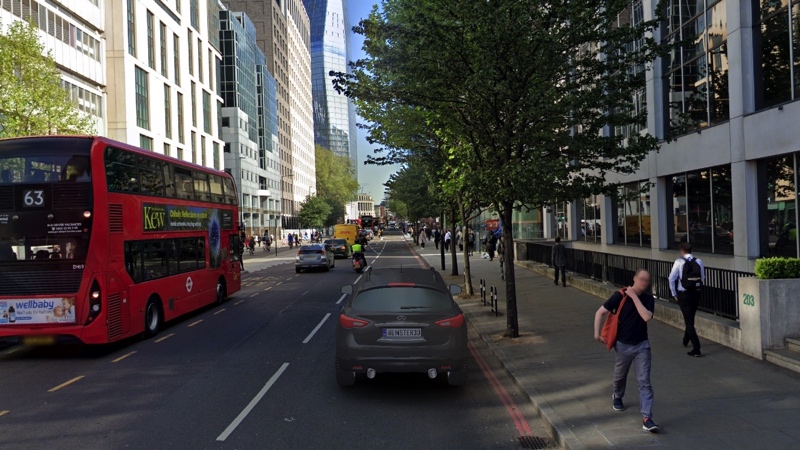} & 
\includegraphics[width=0.3\linewidth]{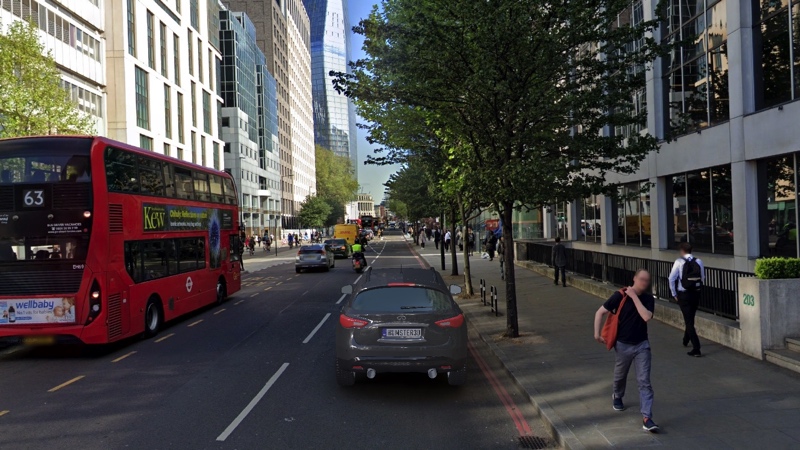} \\
Input image & Hold-Geoffroy \etal~\cite{hold2019deep} & Wang \etal~\cite{wang2021learning} & Ours \\
\end{tabular}
}
\endgroup
\caption{\footnotesize \textbf{Qualitative comparison of virtual object insertion.} 
Our method produces realistic cast shadows and high-frequency ``clear coat'' effects. } 
\label{fig:qual_insertion_comparison} 
\end{figure*}

\begin{figure*}[t!]
\small
\centering
\begingroup
\setlength{\tabcolsep}{0.5pt}
\resizebox{0.99\textwidth}{!}{
\begin{tabular}{cccc}
\includegraphics[width=0.3\linewidth]{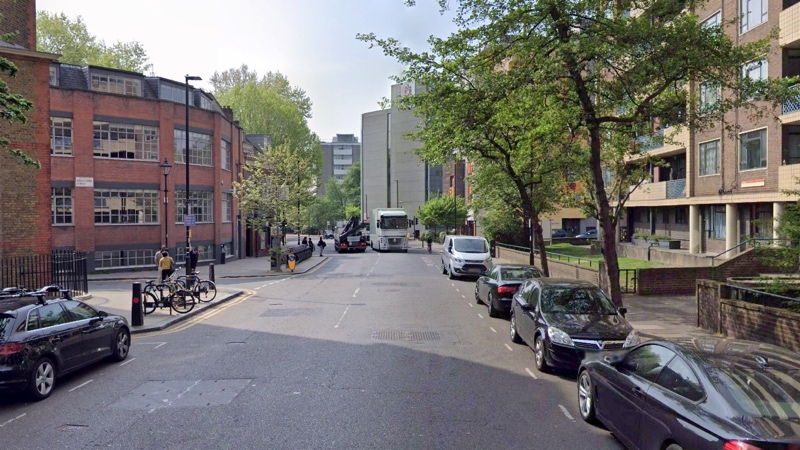} & 
\includegraphics[width=0.3\linewidth]{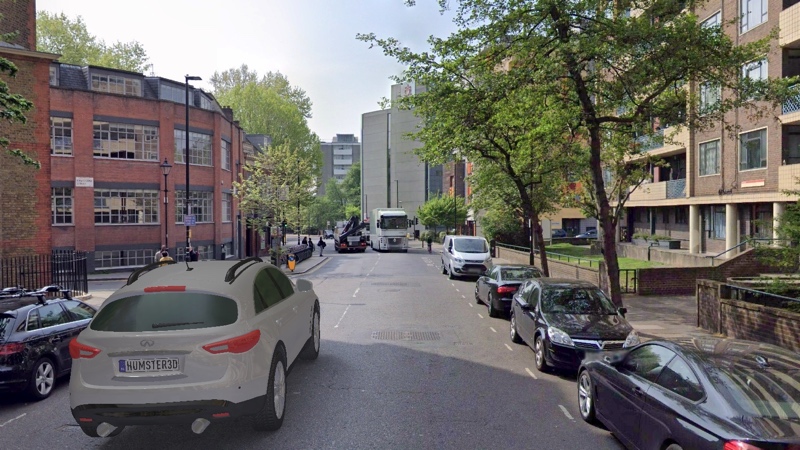} & 
\includegraphics[width=0.3\linewidth]{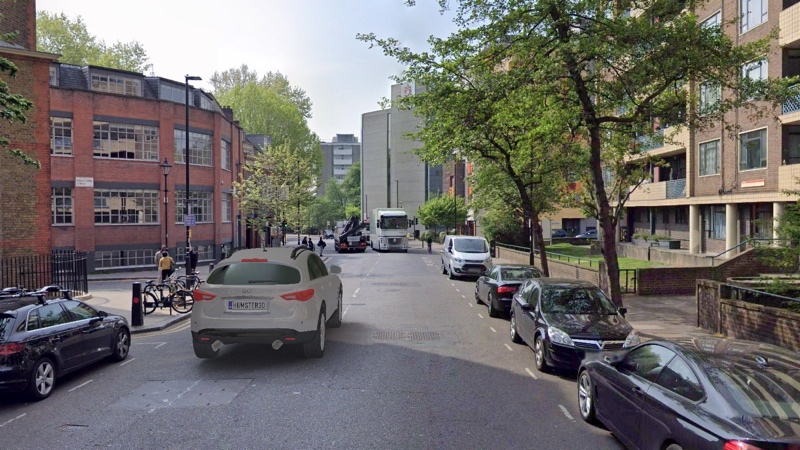} & 
\includegraphics[width=0.3\linewidth]{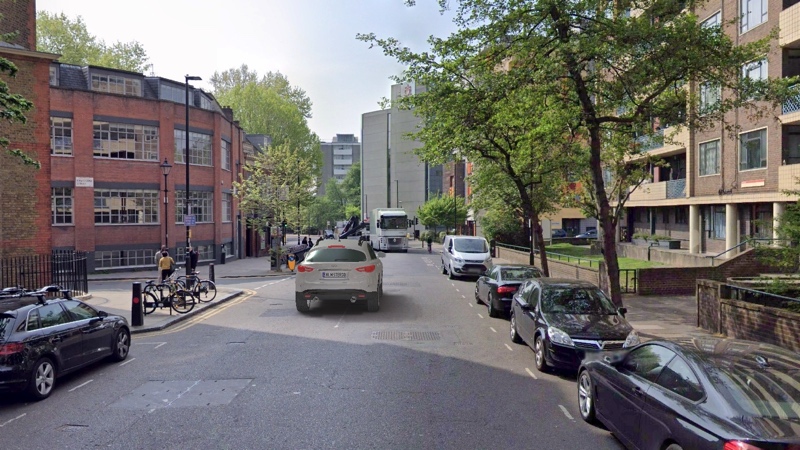} \\
Input image & Inside shadow & Shadow boundary & Outside shadow \\
\end{tabular}
}
\endgroup
\caption{\footnotesize \textbf{Qualitative results of spatially-varying shadow effects.} 
Our method can handle the spatial changes of shadow intensity around shadow boundary, not possible previously. (Results take six surrounding perspective views as input. )
}
\label{fig:qual_sv_shadow} 
\end{figure*}

\mysubsubsection{Human study.} 
To quantitatively evaluate the quality of object insertion, we perform a human study with Amazon Mechanical Turk, where we show two augmented images, randomly permuted, produced by our method and by the baseline. 
We then ask users to compare the realism of the inserted object, e.g. the cast shadows and the reflections, and select the more realistic image. 
For each comparison, we invite 15 users to judge 23 examples. We adopt majority vote for the preference of each example, and run three times to report mean and standard deviation in Tbl.~\ref{table:userstudy}. 
The ablated version of Ours (w/o adv. supervision) outperforms baselines, indicating the hybrid lighting representation improves upon prior works. 
Comparing Ours and Ours (w/o adv. supervision), the results indicate including adversarial supervision leads to more visually realistic editing.

\mysubsubsection{Qualitative comparison.} 
We first visualize the environment maps at different scene locations and insertion of a purely specular sphere in Fig.~\ref{fig:qual_sv_light}. 
Hold-Geoffroy \etal~\cite{hold2019deep} only predicts one environment lighting and  ignored spatially-varying effects. For inserted spheres around shadowed region, it still produces strong cast shadows. Also, the high-frequency details are not well preserved in the sky prediction. 
Wang \etal~\cite{wang2021learning} can generate high-frequency details but fails to handle the extreme HDR intensity of the outdoor scene, and thus cannot generate realistic cast shadows. 
Our method is the only one that handles extreme HDR, angular details and spatially-varying lighting. 
We show virtual object insertion results in Fig.~\ref{fig:qual_insertion_comparison}. Our lighting prediction preserves high-frequency details with HDR intensity, producing realistic highlights and clear coating effects, while prior method cannot generate such effects.


\mysubsubsection{Spatially-varying shadows.} 
Benefiting from the accurate HDR sky and surrounding scene estimation, our method can produce spatially-varying shadow effects. As shown in Fig.~\ref{fig:qual_sv_shadow}, the car casts an intense shadow when outside the shadow of the building, while the shadow caused by the car is much weaker when the car is inside the shadow region. 
Especially, it also shows reasonably different shadow intensity around the shadow boundary. 
This challenging effect requires accurate prediction of direction, intensity and geometry of HDR lighting, and will not occur with an incapable lighting representation. 

\begin{figure*}[t!]
\centering
\begingroup
\setlength{\tabcolsep}{0.5pt}
\resizebox{0.99\textwidth}{!}{
\begin{tabular}{C{0.3\textwidth}C{0.3\textwidth}C{0.3\textwidth}C{0.3\textwidth}}
Initial editing & After optimization & Initial editing & After optimization \\
\multicolumn{4}{c}{\includegraphics[width=1.2\linewidth]{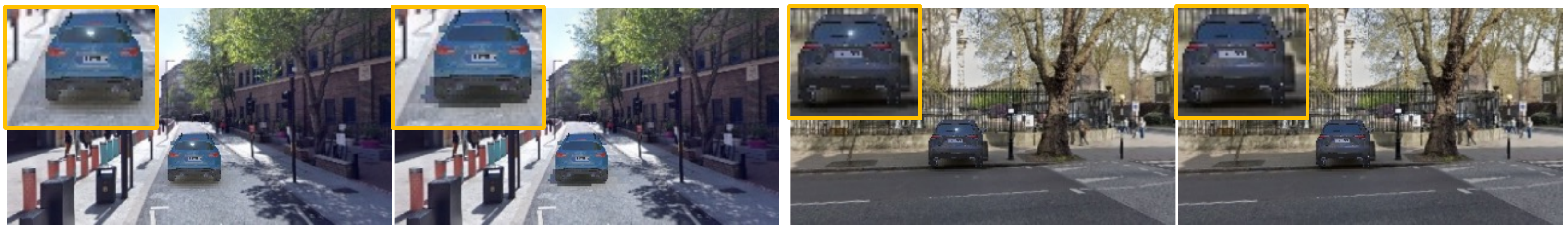}}
\end{tabular}
}
\endgroup
\vspace{-1mm}
\caption{\textbf{Discriminator test-time optimization.}
Note the discriminator corrects the shadow direction (left), and removes erroneous specular highlight (right). The pixelated effect is due to low-resolution rendering during end-to-end training. 
} 
\label{fig:qual_adv} 
\end{figure*} 

\subsection{Ablation Study} 
\label{sec:ablationstydt}

We verify the effectiveness of the sky decoder module and the adversarial supervision. 
In Tbl.~\ref{table:quant_peak}, compared to the ablated version that directly predict sky dome as an environment map (denoted as ``Ours w/o sky modeling''), the full model reduces the error by around 50\%, which demonstrates the sky modeling network is important for achieving accurate sun position prediction in outdoor scenes. 

Adversarial supervision improves the performance on quantitative evaluation in Tbl.~\ref{table:quant_peak},~\ref{table:quant_ldr}, especially for peak direction estimation, which indicates that discriminating on the final image editing result is complementary to existing supervision and benefits lighting prediction. 
In the user study (Tbl.~\ref{table:userstudy}), adversarial supervision improves perceptual realism and receives a higher user preference.
We also qualitatively visualize the behaviour of the discriminator in Fig.~\ref{fig:qual_adv}. To understand the ``photorealism'' implicitly perceived by the discriminator during the training process, we perform test-time optimization on the object insertion results to minimize the adversarial loss, and show the optimized results in Fig.~\ref{fig:qual_adv}. 
In the first example, the initial editing results fail to predict the correct sun location and produce wrong shadows. After test-time optimization, the shadow direction points to the bottom-left of the image and the shadow intensity also matches the visual cues from the rest of the scene. 
In the second example, the initial editing results contain an obviously erroneous highlight, and the discriminator detects the artifact and removes it. 
This agrees with the intuition that a neural discriminator have the capacity to catch lighting effects such as cast shadows and incorrect specular highlights. 
Further details are included in the Appendix. 

\begin{table}[t!]
\centering
\resizebox{0.7\linewidth}{!}{
\addtolength{\tabcolsep}{1pt}
\centering
\begin{tabular}{|l|c|c|c|c|c|c|}
\hline
\textbf{Method} &  {mAP} &  {car} &  {bus} & {trailer} &  {const.vehicle} &  {bicycle} \\
\hline
Real Data & 0.190  & 0.356 & 0.124 & 0.011 & 0.016 & 0.116 \\
+ Aug No Light. & 0.201 & 0.363 & 0.163 & 0.029 & \underline{0.021} & 0.120 \\
+ Aug Light. & \textbf{0.211} & \underline{0.369} & \underline{0.182} & \underline{0.036} & 0.020  & \underline{0.146} \\
\hline
\end{tabular}%
}
\vspace{0.5mm}
\caption{Performance of a SOTA 3D object detector~\cite{wang2021fcos3d} on nuScenes benchmark. 
mAP represents the mean for the 10 object categories. We report individual categories that saw a significant boost (full table in the Appendix). 
}
\label{tab:perception3d}
\end{table}

\subsection{Downstream Perception Task} 
\label{sec:downstream}
We investigate the benefits of our object insertion as data augmentation for a downstream 3D object detection task on nuScenes. The goal of this task is to place a 3D bounding box for $10$ different object categories. We first train a state-of-the-art monocular 3D  detector~\cite{wang2021fcos3d} on a 10\% subset of real data from the nuScenes training set. This subset was chosen randomly across all scenes but in a way that the number of objects per category resembles the original nuScenes training set.  We then augment the front-camera images of this subset with our method.
Specifically, we collect a set of 3D models with categories of \textit{car} and \textit{construction vehicles}, and randomly insert one object per image.
Our augmented dataset has approximately $15$K augmented (new) images. 
We use the same training strategy and model hyperparameters as~\cite{wang2021fcos3d} but do not supervise attributes or velocity as these are not present for the augmented data. 
Quantitatively, in Tbl.~\ref{tab:perception3d} we can observe  that the performance of the detector improves by $2\%$ when comparing to real data. Moreover, we can also see that while naively adding objects leads to a $1\%$ improvement, another $1\%$ is a result of having better light estimation.
Interestingly, we can also notice that the performance of the object detector also improves in different categories even though we do not directly augment those.

%% file: conc.tex
\section{Discussion}
\label{sec:conc}

In this paper, we proposed a hybrid representation of lighting and a novel differentiable object insertion module that allows end-to-end optimization of AR objectives. In a variety of comparisons, we demonstrate the effectiveness of our approach in lighting estimation. Furthermore, we showcase performance gains on a 3D object detection task when training the detector on our augmented dataset. 


While our method presents an effective way of rendering 3D assets into images, some limitations remain for future work. 
Currently, the inserted virtual object pixels do not pass the same capturing process as the background scene, and the shadow rendering assumes Lambertian surface. 
Modeling of camera ISP, weather, and non-Lambertian scene materials can be interesting directions for future work.